\documentclass[lang = american]{article} %

\usepackage{hyperref}

\usepackage{xcolor}
\hypersetup{
    colorlinks,
    linkcolor={blue!80!black},
    citecolor={blue!80!black},
    urlcolor={blue!80!black}
}

\usepackage{arxiv}
\usepackage{tikz}
\usepackage{tcolorbox}
\usepackage{amsmath}
\usepackage{scrextend}
\usepackage{amsthm}
\usepackage{amssymb}
\usepackage{booktabs}

\usepackage{epsf}

\usepackage{psfrag}
\usepackage{graphicx}
\usepackage{amsbsy}
\usepackage{epsfig}
\usepackage{ifthen}
\usepackage{eucal}
\usepackage{xcolor}
\usepackage{multirow}
\usepackage{tabularx}
\usepackage[margin=0pt]{subfig}
\usepackage{tikz}
\usetikzlibrary{shapes.geometric, arrows, shadows, bayesnet}
\usepackage{cleveref}
\crefname{figure}{Fig.}{Figs.}
\crefname{definition}{Defn.}{Defns.}
\crefname{corollary}{Corollary}{Corollaries}
\crefname{lemma}{Lemma}{Lemmas}
\crefname{proposition}{Prop.}{Props.}
\crefname{theorem}{Thm.}{Thms.}
\crefname{assumption}{Assumption}{Assumptions}
\crefname{remark}{Remark}{Remarks}
\crefname{example}{Ex.}{Exs.}
\crefname{principle}{Principle}{Principles}
\crefname{lemma}{Lemma}{Lemmas}
\crefname{table}{Tab.}{Tabs.}
\crefname{section}{\S}{\S\S}
\crefname{subsection}{\S}{\S\S}

\newtheorem{theorem}{Theorem}[section]
\newtheorem{principle}[theorem]{Principle}
\newtheorem{definition}[theorem]{Definition}
\newtheorem{example}[theorem]{Example}
\newtheorem{proposition}[theorem]{Proposition}

\newtheorem{assumption}[theorem]{Assumption}

\newcommand{\commentout}[1]{}

\renewcommand{\Pr}{p}

\newcommand{\jvk}[1]{{{\color{purple} ~ Julius: #1}}}

\newcommand{\RR}{\mathbb{R}}    %
\newcommand{\R}{\mathbb{R}}    %
\newcommand{\Hcal}{{\mathcal{H}}} %
\newcommand{\Xcal}{\mathcal{X}} %
\newcommand{\Ycal}{\mathcal{Y}} %

\newcommand{\xb}{\mathbf{x}}    %
\newcommand{\x}{\mathbf{x}}    %
\newcommand{\Eb}{\mathbf{E}}    %
\newcommand{\Fb}{\mathbf{F}}    %

\newcommand{\sgn}{\mathop{\mathrm{sgn}\,}}

\newcommand{\argmax}{\mathop{\mathrm{argmax}\,}}

\newcommand{\emp}{\mathrm{emp}}      %
\def\Xb{\mathbf{X}}

\newcommand{\eq}[1]{(\protect\ref{#1})}

\newcommand{\independent}{\perp\mkern-11mu\perp}

\newcommand{\pa}{\mathbf{pa}}
\newcommand{\PA}{\mathbf{PA}}

\numberwithin{equation}{section}

\begin{document}

\title{From Statistical to Causal Learning
}

\author{
Bernhard Sch\"olkopf \\
  Max Planck Institute for Intelligent Systems, T\"ubingen, Germany \\
  \texttt{bs@tuebingen.mpg.de} \\
   \And
Julius von K\"ugelgen \\
  Max Planck Institute for Intelligent Systems, T\"ubingen, Germany \\
  University of Cambridge, United Kingdom \\
  \texttt{jvk@tuebingen.mpg.de} \\
  }

\maketitle

\begin{abstract}
\renewcommand{\thefootnote}{\fnsymbol{footnote}}
We describe basic ideas underlying research to build and understand artificially intelligent systems: from symbolic approaches via statistical learning to interventional models relying on concepts of causality. Some of the hard open problems of machine learning and AI are intrinsically related to causality, and progress may require advances in our understanding of how to model and infer causality from data.\footnote{To appear in the \textit{Proceedings of the International Congress of Mathematicians 2022, EMS Press}. Both authors contributed equally to this work; names listed in alphabetical order.}
\renewcommand{\thefootnote}{\arabic{footnote}}
\setcounter{footnote}{0}
\end{abstract}

\bigskip

\leftline{\bfseries Mathematics Subject Classification 2020}
Primary 68T05; Secondary 68Q32, 68T01, 68T10, 68T30, 68T37

\bigskip

\leftline{\bfseries Keywords}
Causal inference, machine learning, causal representation learning

\section{Introduction}
In 1958, the New York Times reported on a new machine called the \textit{perceptron}. Frank Rosenblatt, its inventor, demonstrated that the perceptron was able to learn from experience. He predicted that later perceptrons would be able to recognize people, or instantly translate spoken language. Now a reality, this must have sounded like distant science fiction at the time. In hindsight, we may consider it the birth of machine learning, the field fueling most of the current advances in artificial intelligence (AI). 

Around the same time, another equally revolutionary development took place: scientists understood that computers could do more than compute numbers: they can process symbols. Although this insight was also motivated by artificial intelligence, in hindsight it was the birth of the field of computer science. 
\looseness-1 There was great optimism that the manipulation of symbols, in programs written by humans, implementing rules designed by humans, should be enough to generate intelligence. 
Below, we shall refer to this as the {\em symbol-rule hypothesis}.\footnote{The term should be taken with a grain of salt, since it suggests a separation between representations and computations which is hard to uphold in practice.} 

There was initially encouraging progress on seemingly hard problems such as automatic theorem proving and computer chess. One of the fathers of the field, Herb Simon, predicted in 1956 that ``machines will be capable, within twenty years, of doing any work a man can do.'' However, problems that appeared simple, such as most things animals could do, turned out to be hard. This came to be known as {\em Moravec‘s paradox}. When IBM's {\em Deep Blue} chess computer beat Garry Kasparov in 1997, Kasparov was physically facing a human during the match: while Deep Blue was capable of analyzing the game's search tree in unprecedented detail, it was unable to recognize and physically move chess pieces, so this task had to be relegated to a human, in an inversion of the famous {\em mechanical turk}.\footnote{\url{https://en.wikipedia.org/wiki/Mechanical_Turk}}
In the years to follow, the field of AI entered what came to be known as the {\em AI winter}. The community got disillusioned with the lack of progress and prospects, and interest greatly declined. However, largely independently of the field of classic AI, {\em machine learning} eventually started to boom. Like Rosenblatt's early work, it was built on the observation that all existing examples of truly intelligent systems---i.e., animals, including humans---were not built on the symbol-rule hypothesis: \looseness-1 both the representations and the rules implemented by natural intelligent systems are acquired from experience, through processes of evolution and learning.

Rather than exploring the well-known dichotomy between rule-based and learning-based approaches, we will explore the less known questions of causality and interventions. While the field of causality in computer science was initially strongly linked to classic AI, recent years have witnessed great interest in connecting it to machine learning \cite{PetJanSch17}. Below, we explore some of these connections, drawing from \cite{1911.10500,scholkopfetal21}. We will argue that the causal view is relevant when it comes to addressing crucial open problems of machine learning, related to notions of robustness and generalization beyond the training distribution. 

\paragraph{Overview}
In statistical learning, our starting point is a joint distribution $p(\Xb)$ generating the observable data. Here, $\Xb$ is a random vector, and we are usually given a dataset $\xb_1,\dots,\xb_m$ sampled i.i.d.\ from $p$. 
We are often interested in estimating properties of conditionals of some components of $\Xb$ given others, e.g., a classifier (which may be obtained by thresholding a conditional at~$0.5$). This is a nontrivial inverse problem, giving rise to statistical learning theory~(\cref{sec:slt}).

Causal learning is motivated by shortcomings of statistical learning (\cref{sec:from}). Its starting point is a structural causal model (SCM) \cite{Pearl2009} (\cref{sec:causal}). In an SCM, the components $X_1,\dots,X_n$ of $\Xb$ are identified with vertices of a directed graph whose arrows represent direct causal influences, and there is a random variable $U_i$ for each vertex, along with a function $f_i$ which computes $X_i$ from its graph parents $\PA_i$ and $U_i$, i.e.,
\begin{equation}
\label{eq:intro}
X_i := f_i(\PA_i, U_i).
\end{equation}
Given a distribution over the $U_i$, which are assumed independent, this also gives rise to a probabilistic model $p(\Xb)$. However, 
the model in~\eqref{eq:intro} is more structured: the graph connectivity and the functions $f_i$ create particular dependences between the observables. Moreover, it describes how the system behaves under intervention: by replacing functions by constants, we can compute the effect of setting some variables to specific values.

Causal learning builds on assumptions different from standard machine learning~(\cref{sec:icm}), and addresses a different level in the modeling hierarchy~(\cref{sec:levels}). It also comes with new problems, such as causal discovery, where we seek to infer properties of graph and functions from data~(\cref{ch:2-discovery}). 
In some cases, conditional independences among the $X_i$ contain information about the graph \cite{Spirtes2000}; but novel assumptions let us handle some cases that were previously unsolvable \cite{Janzingetal12}. Those assumptions have 
nontrivial implications for machine learning tasks such as semi-supervised learning, covariate shift adaptation and transfer learning~\cite{Schoelkopf2012}~(\cref{sec:implications}). %
Once provided with a causal model, causal reasoning~(\cref{sec:reasoning}) allows us to identify and estimate certain causal queries of interest from observational data.
We conclude with a list of some current and open problems~(\cref{sec:open}), with a particular emphasis on the topic of causal representation learning.

The presentation and notation will be somewhat informal in several respects.
We generally assume that all distributions possess densities (w.r.t.\ a suitable reference measure). We sometimes write $p(x)$ for the distribution (or density) of a random variable $X$. Accordingly, the same $p$ can denote another distribution 
$p(y)$, distinguished by the argument of $p(\cdot )$.  
\looseness-1 We also sometimes use summation for marginalization which supposes discrete variables; the corresponding expressions for continuous quantities would use integrals.

\section{Statistical Learning Theory}
\label{sec:slt}
\begin{figure}[t]
\begin{minipage}[c]{0.5\textwidth}
\caption[font=small,labelsep=none]{Given a small number of observations, how do we find a law underlying them? Leibniz argued that even if we generate a random set of points, we can always find a mathematical equation satisfied by these points.}
\label{fig1}
\end{minipage}
\hfill
\begin{minipage}[c]{0.5\textwidth}
\centering
\includegraphics[width=4cm]{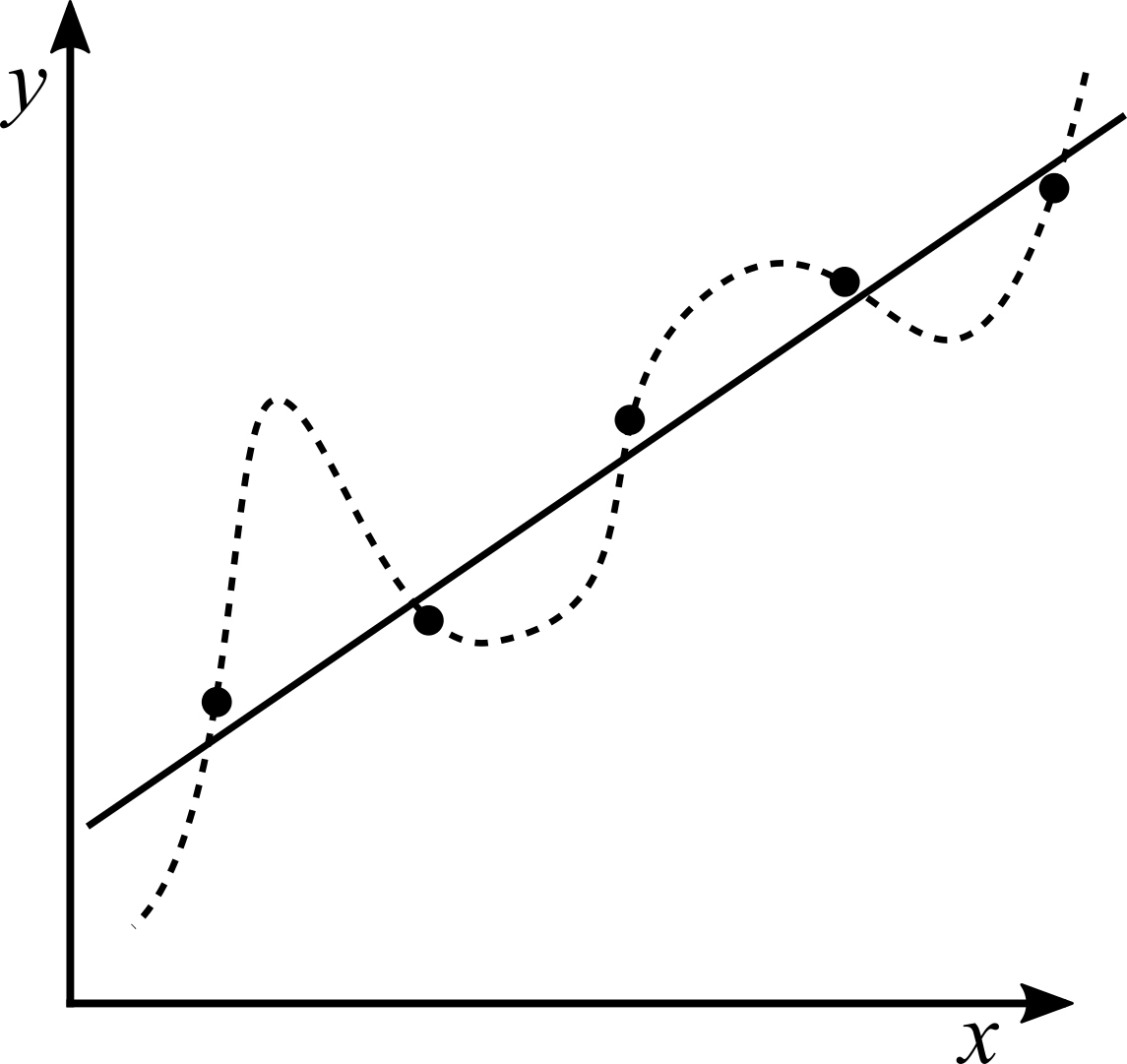}
\end{minipage}
\end{figure}

Suppose we have measured two statistically dependent observables and found the points to lie approximately on a straight line. An empirical scientist might be willing to hypothesize a corresponding law of nature (see~\cref{fig1}).
However, already Leibniz pointed out that if we scatter spots of ink randomly on a piece of paper by shaking a quill pen, we can also find a mathematical equation satisfied by these points \cite{LeibnizDiscours}. He argued that we would not call this a law of nature, because no matter how the points are distributed, there always exists such an equation; we would only call it a law of nature only if the equation is simple. This raises the question of what makes an equation simple. The physicist Rutherford took the pragmatic view that if there is a law, it should be directly evident from the data:
``if your experiment needs statistics, you ought to have done a better experiment.''\footnote{Cited after \url{http://www.warwick.ac.uk/statsdept/staff/JEHS/data/jehsquot.pdf}.} This view may have been a healthy one when faced with low-dimensional inference problems where regularities are immediately obvious; however, modern AI is facing inference problems that are harder: they are often high-dimensional and nonlinear, yet we may have little prior knowledge about the underlying regularity (e.g., for medical data, we usually do not have a mechanistic model). 

\textit{Statistical learning theory} studies the problem of how to still perform valid inference, provided that we have sufficiently large datasets and the computational means to process them. 
Let us look at some theoretical results for the simplest learning scenario, drawing from \cite{SchSmo02}; for details, see \cite{Vapnik98}.
Suppose we are given empirical observations,
\begin{equation}
  \label{eq:slt:data}
  (x_1,y_1),\dots,(x_m,y_m) \in \Xcal\times\Ycal,
\end{equation}
where $\Xcal$ is some nonempty set from which the {\em inputs} come, and $\Ycal=\{\pm 1\}$ is the {\em output} set, in our case consisting of just two {\em classes}. This situation is called \emph{pattern recognition}, and our goal is to use the \emph{training data} \eq{eq:slt:data} to infer a function $f:\Xcal\to \{\pm 1\}$ (from some function class chosen a priori) which will produce the correct output for a new input $x$ which we may not have seen before. To formalize what we mean by ``correct'', we make the assumption that all observations $(x_i,y_i)$ have been generated independently by performing a random experiment described by an unknown probability distribution $\Pr(x,y)$---a setting referred to as {\em i.i.d.\ (independent and identically distributed) data}. Our goal will be to minimize the expected error (or risk)
\begin{equation}
\label{eq:slt:r}
R[f]=\int_{\Xcal \times \Ycal} c(y, f(x)) \; d\Pr(x,y),
\end{equation}
where $c$ is a so-called loss function, e.g., the misclassification error
$c(y,f(x)) = \frac{1}{2}|f(x)-y|$ taking the value $0$ whenever $f(x)=y$ and $1$ otherwise.

The difficulty of the task stems from the fact that we are trying to
minimize a quantity that we cannot evaluate: since we do not
know $\Pr$, we cannot compute \eq{eq:slt:r}. We do know, however, the training data \eq{eq:slt:data}
sampled from $\Pr$.  We can thus try to infer a function $f$ from the
training sample whose risk is close to the minimum of \eq{eq:slt:r}. To this
end, we need what is called an \emph{induction
  principle}.

One way to proceed is to use the training sample to approximate \eq{eq:slt:r} by a finite sum, referred to as the \emph{empirical
risk}
\begin{equation}
\label{eq:slt:r_emp}
R_\emp[f]=\frac{1}{m}\sum_{i=1}^m c(x_i, y_i, f(x_i)).
\end{equation}
The \emph{empirical risk minimization (ERM) induction
principle} recommends that we choose (or ``learn'') an $f$ that minimizes \eq{eq:slt:r_emp}. 
We can then ask whether the ERM principle is statistically \emph{consistent}: in the limit of infinitely many data points, will ERM lead to a solution which will do as well as possible on future data generated by $\Pr$? 

\begin{figure}[t]
\centering
\includegraphics[width=17cm,trim=0em 23em 0em 18em, clip]{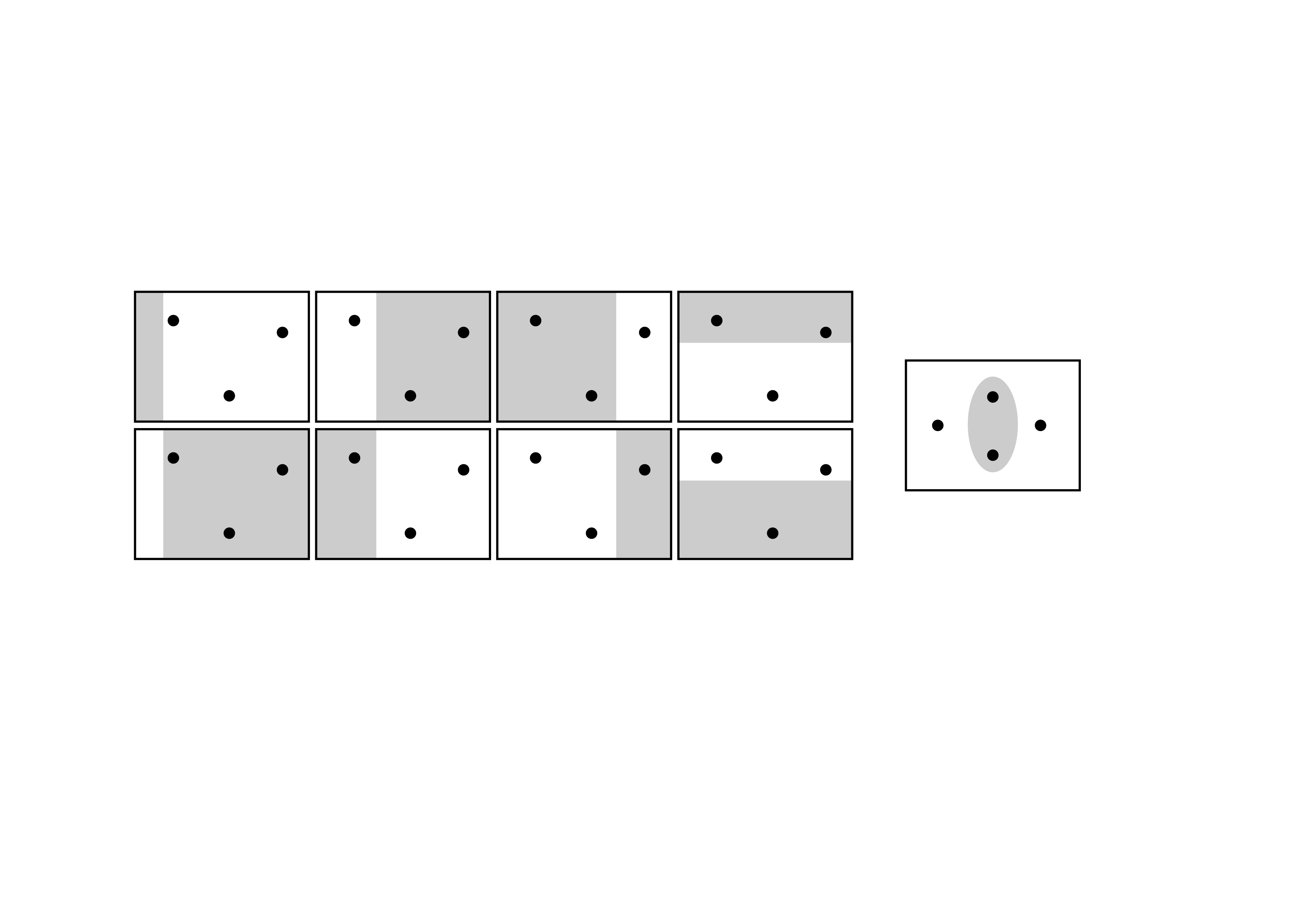}
\caption{Using straight lines, we can separate three points in all possible ways; we cannot do this for four points, no matter how they are placed. The class of linear separations is not ``falsifiable'' using three points, but it becomes falsifiable once we have four or more points.}
\label{fig-VC}
\end{figure}

It turns out that if the function class over which we minimize \eq{eq:slt:r_emp} is too large, then ERM is not consistent. Hence, we need to suitably restrict the class of possible functions. For instance, ERM is consistent for all probability distributions, provided that the \emph{VC dimension} of the function class is finite. The VC dimension is an example of a {\em capacity measure}. It is defined as the maximal number of points that can be separated (classified) in all possible ways using functions from the class. E.g., using linear classifiers (separating classes by straight lines) on $\RR^2$, we can realize all possible classifications for 3 suitably chosen points, but we can no longer do this once we have 4 points, no matter how they are placed (see~\cref{fig-VC}).
\looseness-1 This means that the VC dimension of this function class is 3. More generally, for linear separations in $\RR^d$, the VC dimension is~$d+1$. 

Whenever the VC dimension is finite, our class of functions (or explanations) becomes falsifiable in the sense that starting from a certain number of observations, no longer all possible labelings of the points can be explained (cf.~\cref{fig-VC}). If we can nevertheless explain a sufficiently large set of observed data, we thus have reason to believe that this is a meaningful finding.

Much of machine learning research is concerned with restrictions on classes of functions to make inference possible, be it by imposing prior distributions on function classes, through other constraints, or by designing self-regularizing learning procedures, e.g., gradient descent methods for neural networks \cite{LeCBenHin15}. While there is a solid theoretical understanding of supervised machine learning as described above (i.e., function learning from input-output examples), there are still details under investigation, such as the recently observed phenomenon of ``double descent'' \cite{belkin}.

A popular constraint, implemented in the \emph{Support Vector Machine (SVM)} \cite{Vapnik98,SchSmo02}, is to consider linear separations with large margin: it turns out that for large margin separations in high-dimensional (or infinite-dimensional) spaces, the capacity can be much smaller than the dimensionality, making learning possible in situations where it would otherwise fail.

For some learning algorithms, including SVMs and nearest neighbor classifiers, there are strong universal consistency results, guaranteeing convergence of the algorithm to the lowest achievable risk, for any problem to be learned \cite{DevGyoLug96,Vapnik98,SchSmo02,SteChr08}.
Note, however, that this convergence can be arbitrarily slow.

For a given sample size, it will depend on the problem being learned whether we achieve low expected error. 
In addition to asymptotic consistency statements, learning theory makes finite sample size statements: one can prove that with probability at least $1-\delta$ (for $\delta>0$), for all functions $f$ in a class of functions with VC dimension $h$,
\begin{equation}
  \label{slt-eq:unibound2}
R[f] \le R_\emp[f] + 
\sqrt{\frac{1}{m}\left( h \left(\log (2m/h)+1\right) + \log\frac{4}{\delta} \right)}.
\end{equation}%
This is an example of a class of results that relate the training error $R_\emp[f]$ and the test error $R[f]$ using a confidence interval (the square root term) depending on a capacity measure of a function class (here, its VC dimension $h$).
It says that with high probability, the expected error $R[f]$ on future observations generated by the unknown probability distribution is small, provided the two terms on the right hand side are small: the training error~$R_\emp[f]$ (i.e., the error on the examples we have already seen), and the square root term, which will be small whenever the capacity $h$ is small compared to the number of training observations~$m$. 
If, on the other hand, 
we try to learn something that may not make sense, such as the mapping from the name of people to their telephone number,
we would find that to explain all the training data (i.e., to obtain a small $R_\emp[f]$), we need a model whose capacity~$h$ is large, and the second term becomes large.
In any case, it is crucial for both consistency results and finite sample error bounds such as \eq{slt-eq:unibound2} that we have i.i.d.\ data.

\paragraph{Kernel Methods} 
A symmetric function $k:{\mathcal X}^2\to {\mathbb{R}}$, where ${\mathcal X}$ is a nonempty set, is called a positive definite (pd) \textit{kernel} if for arbitrary points
$x_1,\dots,x_m\in {\mathcal X}$ and
coefficients $a_1,\dots,a_m\in {\mathbb R}$:
$$\sum_{i,j}a_ia_jk(x_i,x_j)\ge 0.$$
The kernel is called strictly positive definite if for pairwise distinct points, the implication $\sum_{i,j}a_ia_j k(x_i,x_j)=0 \Longrightarrow \forall i: a_i=0$ is valid. 
Any positive definite kernel induces a mapping 
\begin{equation}\label{eq:kernelmap}
\Phi: x\mapsto k(x,.)
\end{equation}
into a \textit{reproducing kernel Hilbert space} (RKHS) $\Hcal$ satisfying
\begin{equation}\label{eq:kernel}
    \langle k(x,.),k(x',.)\rangle = k(x,x')
\end{equation}
for all $x,x'\in {\mathcal X}$.
Although $\Hcal$ may be infinite-dimensional, we can construct practical classification algorithms in $\Hcal$ provided that all computational steps are carried out in terms of scalar products, since those can be reduced to kernel evaluations \eq{eq:kernel}. 

In the SVM algorithm, the capacity of the function class is restricted by enforcing a large margin of class separation in $\Hcal$ via a suitable RKHS regularization term. The solution can be shown to take the form
\begin{equation}\label{eq:solution}
f(x) = \sgn \bigg( \sum_i \alpha_i k(x_i,x) + b \bigg),
\end{equation}
where the learned parameters $\alpha_i$ and $b$ are the solution of a convex quadratic optimization problem. A similar expansion of the solution in terms of kernel functions evaluated at training points holds true for a larger class of kernel algorithms beyond SVMs, regularized by an RKHS norm \cite{SchHerSmo01}.

\looseness-1 In kernel methods, the kernel plays three roles which are crucial for machine learning: it acts as a similarity measure for data points, induces a representation in a linear space\footnote{Note that the data domain $\Xcal$ need not have any structure other than being a nonempty set.} via \eq{eq:kernelmap}, and parametrizes the function class within which the solution is sought, cf.~\eq{eq:solution}.

\paragraph{Kernel Mean Embeddings}
Consider two sets of points 
$X:= \{ x_1,\dots,x_m\}\subset {\mathcal X}$ and  $Y:= \{ y_1,\dots,y_n\}\subset {\mathcal X}$. We define the mean map $\mu$ as \cite{SchSmo02}
\begin{equation}
\mu(X) = \frac{1}{m}\sum_{i=1}^m k(x_i,\cdot).
\end{equation}
For polynomial kernels $k(x,x')=(\langle x, x'\rangle+1)^d$, we have $\mu(X)=\mu(Y)$ if 
all empirical moments up to order $d$ coincide. For
strictly pd kernels, the means coincide only if $X=Y$, rendering $\mu$ injective \cite{SchSriGreFuk08}. 
The mean map has some other interesting properties \cite{SmoGreSonSch07}, e.g.,
$\mu(X)$ represents the operation of taking a mean of a
function on the sample $X$:
$$\langle \mu(X),f \rangle = \bigg\langle \frac{1}{m}\sum_{i=1}^m k(x_i,\cdot),f \bigg\rangle =
\frac{1}{m}\sum_{i=1}^m f(x_i)$$
Moreover, we have 
$$\| \mu(X)-\mu(Y)\| = \sup_{\|f\|\le 1} \left|\langle \mu(X)-\mu(Y),f\rangle \right|
= \sup_{\|f\|\le 1} \bigg| \frac{1}{m}\sum_{i=1}^m f(x_i)-\frac{1}{n}\sum_{i=1}^n f(y_i) \bigg|.$$

If $\mathbb{E}_{x,x'\sim p}[k(x,x')],~\mathbb{E}_{x,x'\sim q}[k(x,x')]<\infty$, then
the above statements, including the injectivity of $\mu$, generalize to Borel measures $p,q$, if we define the mean map as
$$\mu \colon p \mapsto \mathbb{E}_{x\sim p} [k(x,\cdot)],$$ 
and replace the notion of strictly pd kernels by that of
characteristic kernels \cite{FukGreSunSch08}.  
This means that we do not lose information when representing a probability distribution in the RKHS. This enables us to work with distributions using Hilbert space methods, and construct practical algorithms analyzing distributions using scalar product evaluations.

Note that the mean map $\mu$
can be viewed as a generalization of the \emph{moment generating
  function} $M_p$ of a random variable $x$ with distribution $p$,
\begin{equation*}
M_p (.) = \mathbb{E}_{x\sim p}\Big[ e^{\langle x,\;\cdot\;\rangle}\Big].
\end{equation*}
The map $\mu$ has applications in a number of tasks including computing functions of random variables \cite{SchMuaFukHarPet15}, testing
of homogeneity \cite{GreBorRasSchetal07}, and of independence \cite{GreFukTeoSonetal08}. 
The latter will be of particular interest to causal inference: we can develop a kernel-based independence test by computing the distance between sample-based embeddings of a joint distribution $\Pr(X,Y)$ and the product of its marginals $\Pr(X),\Pr(Y)$ \cite{Gretton2005,Gretton2005JMLR,GreFukTeoSonetal08,Zhang2011uai,PfiBuhSchPet18}, \looseness-1 and generalize it to conditional independence testing \cite{FukGreSunSch08,parmua20}, as required for certain causal discovery methods (see~\cref{ch:2-discovery}).

\section{From Statistical to Causal Models\label{sec:from}}
\paragraph{Methods Relying on i.i.d.\ Data}
In current successes of machine learning \cite{LeCBenHin15}, we generally
{\em (i) have large amounts of data}, often from simulations or large-scale human labeling, {\em (ii) use high capacity machine learning models} (e.g., neural networks with many adjustable parameters), and {\em (iii) employ high performance computing}.
Statistical learning theory offers a partial explanation for recent successes of 
learning:
huge datasets
enable training complex models and thus solving increasingly difficult tasks.

However, a crucial aspect that is
often ignored is that we {\em (iv) assume that the data are i.i.d.}
This assumption is crucial for good performance in practice, 
and it underlies theoretical statements such as~\eq{slt-eq:unibound2}.
When faced with problems that violate the i.i.d.\ assumption, all bets are off. Vision systems can be grossly misled if an object that is normally recognized with high accuracy is placed in a context that {\em in the training set} may be negatively correlated with the presence of the object. For instance, such a system may fail to recognize a cow standing on the beach. In order to successfully generalize in such settings, we would need to construct systems which do not merely rely on statistical dependences, but instead model 
mechanisms that are robust across certain violations of the i.i.d.\ assumption.
As we will argue, causality provides a natural framework for capturing such stable mechanisms and reasoning about different types of distribution shifts.

\paragraph{Correlation vs Causation}

It is a commonplace that \textit{correlation does not imply causation}.
Two popular and illustrative examples are the positive correlation between chocolate consumption and nobel prizes per capita \cite{messerli2012chocolate},
and that between the number of stork breeding pairs and human birth rates \cite{matthews2000storks}, neither of which admit a sensible interpretation in terms of direct causation. 
These examples naturally lead to the following questions: What exactly do we mean by ``causation''?
\looseness-1 What is its relationship to correlation? 
And, if correlation alone is not enough, what is needed to infer causation?

Here, we adopt a notion of causality based on manipulability~\cite{woodward2001causation} and intervention~\cite{Pearl2009} which has proven useful in fields such as agriculture \cite{wright1921correlation}, econometrics~\cite{haavelmo1944probability, hoover2001causality} and epidemiology~\cite{robins2000marginal}.
\begin{definition}[Causal effect]
\label{def:causal_effect}
We say that a random variable $X$ has a causal effect on a random variable $Y$ if there exist $x\neq x'$ s.t. the distribution of $Y$ after intervening on $X$ and setting it to $x$ differs from the distribution of $Y$ after setting $X$ to $x'$.
\end{definition}
Inherent to the notion of causation, there is a directionality and asymmetry which does not exist for correlation: 
\looseness-1 if $X$ is correlated with $Y$, then $Y$ is equally correlated with $X$;
but, if $X$ has a causal effect on $Y$, the converse (in the generic case)
does not hold.

\begin{figure}[t]
    \centering
    \subfloat{
    \includegraphics[width=.35\textwidth]{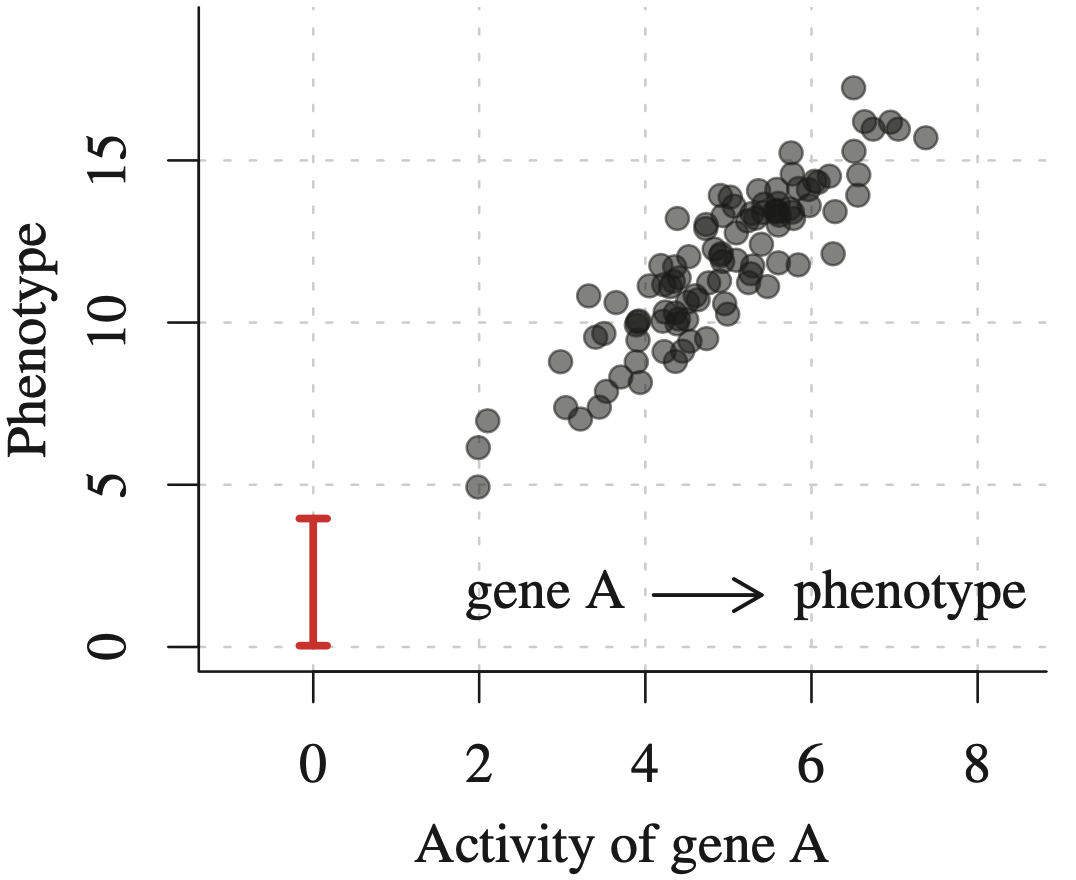}
    }
    \qquad     \qquad 
    \subfloat{
    \includegraphics[width=.35\textwidth]{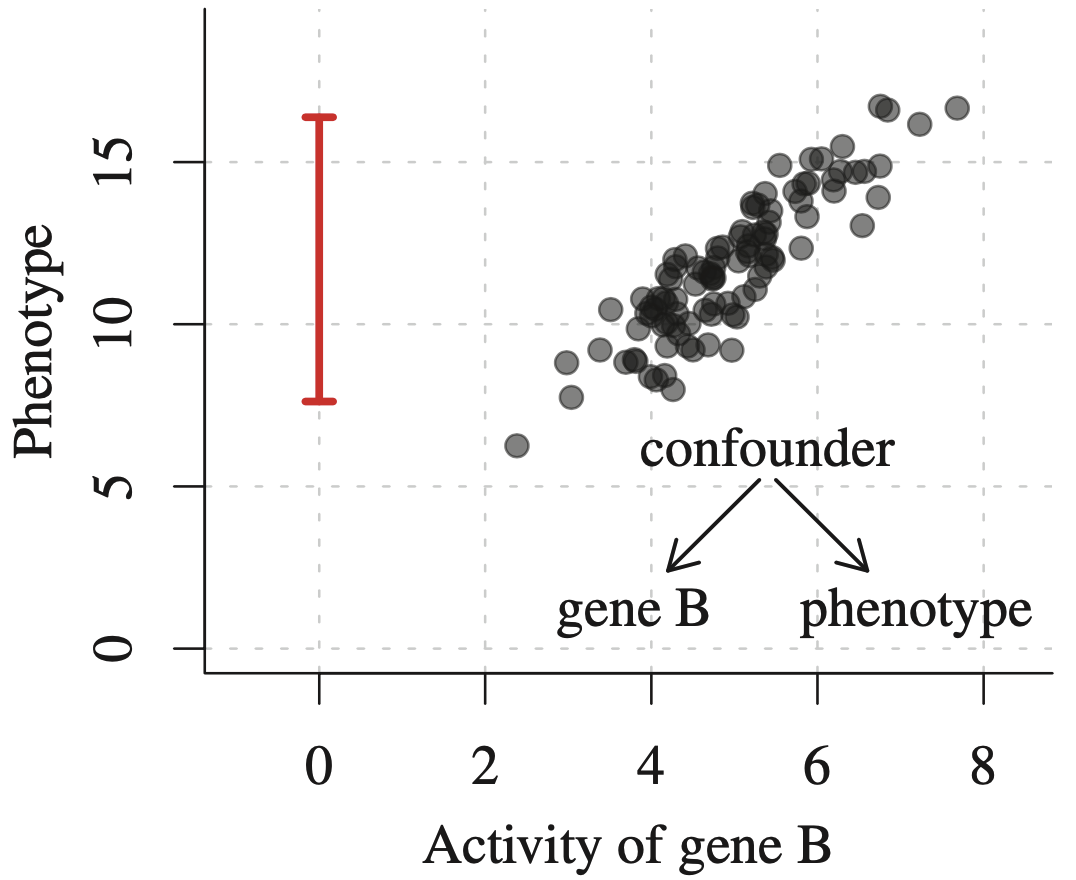}
    }
    \caption[Difference between correlation and causation]{Measurements of two genes ($x$-axis), gene A (left) and  gene B (right), show the same strong positive correlation with a phenotype ($y$-axis). However, this  statistical information alone is insufficient to predict the outcome of a knock-out experiment where the activity of a gene is set to zero (vertical lines at $x=0$).
    Answering such \textit{interventional} questions requires additional causal knowledge (inset causal graphs):
    knocking out gene A, which is a direct cause, would lead to a reduction in phenotype, whereas knocking out gene B, which shares a common cause, or confounder, with the phenotype but has no causal effect on it, would leave the phenotype unaffected.
    This shows that correlation alone is not enough to predict the outcome of perturbations to a system (toy data, figure from \cite{PetJanSch17}).}
    \label{fig:genes}
\end{figure}

We illustrate the intervention-based notion of causation and its difference from correlation (or, more generally, statistical dependence) in~\cref{fig:genes}.
Here, knocking out two genes $X_A$ and $X_B$ that are indistinguishable based on their correlation with a phenotype $Y$ would have very different effects.
Only intervening on $X_A$ would change the distribution of $Y$, whereas $X_B$ does not have a causal effect on $Y$---instead, their correlation arises from a different (confounded) causal structure.
Such causal relationships are most commonly represented in the form of \textit{causal graphs} where directed arrows indicate a direct causal effect.

The example in~\cref{fig:genes} shows that the same correlation can be explained by multiple causal graphs which lead to different experimental outcomes, i.e., \textit{correlation does not imply causation}.
However, there is a
connection between correlation and causation, expressed by Reichenbach~\cite{Reichenbach1956} as the Common Cause Principle, see~\cref{fig:reichenbach}.

\begin{principle}[Common Cause]
\label{princ:reichenbach}
If two random variables $X$ and $Y$ are statistically dependent ($X\not\independent Y$), then there exists a random variable $Z$ which causally influences both of them and which explains all their dependence in the sense of rendering them conditionally independent ($X\independent Y\mid Z$).
As a special case, $Z$ may coincide with $X$ or $Y$. 
\end{principle}
According to \Cref{princ:reichenbach}, statistical dependence always results from underlying causal relationships by which variables, including potentially unobserved ones, 
influence each other.
Correlation is thus an epiphenomenon, the by-product of a causal process. 

\newcommand{\xshift}{4em}
\newcommand{\yshift}{3em}
\newcommand{\nodescale}{1}
\begin{figure}[t]
\centering
\subfloat[]{
\centering
    \begin{tikzpicture}
    \centering
    \node (X) [obs, scale=\nodescale] {$X$};
    \node (Y) [obs, xshift=1.25*\xshift, scale=\nodescale] {$Y$};
    \edge{X}{Y};
    \end{tikzpicture}
}
\qquad \qquad 
\subfloat[]{
    \centering
    \begin{tikzpicture}
    \centering
    \node (Y) [obs, scale=\nodescale] {$Y$};
    \node (X) [obs, xshift=-1.25*\xshift, scale=\nodescale] {$X$};
    \edge{Y}{X};
    \end{tikzpicture}
}
\qquad \qquad 
\subfloat[]{
\centering
    \begin{tikzpicture}
    \centering
    \node (X) [obs, scale=\nodescale] {$X$};
    \node (Z) [latent, xshift=1.25*\xshift, scale=\nodescale] {$Z$};
    \node (Y) [obs, xshift=2.5*\xshift, scale=\nodescale] {$Y$};
    \edge{Z}{X,Y};
    \end{tikzpicture}
}%
\caption[Reichenbach's common cause principle]{ Reichenbach's common cause principle~\cite{Reichenbach1956} postulates that statistical dependence between two random variables $X$ and $Y$ has three elementary possible causal explanations shown as causal graphs in (a)--(c).
It thus states that association is always induced by an underlying causal process.
In (a) the common cause $Z$ coincides with $X$, and in (b) it coincides with $Y$.
Grey nodes indicate observed and white nodes unobserved variables.}
\label{fig:reichenbach}
\end{figure}
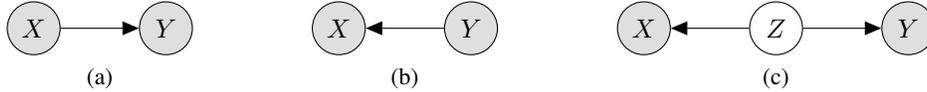

For the example of chocolate consumption ($X$) and Nobel laureates ($Y$), 
common sense suggests that neither of the two variables should have a causal effect on the other, i.e., neither chocolate consumption driving scientific success ($X\to Y$; \cref{fig:reichenbach}a) nor Nobel laureates increasing chocolate consumption ($Y\to X$; \cref{fig:reichenbach}c) seem plausible. 
\Cref{princ:reichenbach} then tells us that the observed correlation must be explained by a common cause $Z$ as in~\cref{fig:reichenbach}c. 
A plausible candidate for such a confounder could, for example, be economic factors driving both consumer spending
and investment in education and science.

Without such background knowledge or additional assumptions, however,  we cannot distinguish the three cases in~\cref{fig:reichenbach} through passive observation, i.e., in a purely data-driven way: the class of observational distributions over $X$ and $Y$ that can be realized by these models is the same in all three cases. 

To be clear, this does not mean that correlation cannot be useful, nor that causal insight is always required.
Both genes in~\cref{fig:genes} remain useful features for making predictions in a passive, or \textit{observational}, setting in which we measure the activities of certain genes and  are asked to predict the phenotype.
Similarly, chocolate consumption remains predictive of winning Nobel prizes.
However, if we want to answer interventional questions, such as the outcome of a gene-knockout experiment or the effect of a policy enforcing higher chocolate consumption, we need more than correlation: \textit{a causal model}.

\section{Causal Modeling Frameworks\label{sec:causal}}
Causal inference has a long history in a variety of disciplines, including statistics, econometrics, epidemiology, and AI.
As a result, different frameworks for causal modeling have emerged over the years and coexist today. 
The first framework described below (CGM) starts from the distribution of the observables, combining it with a directed graph to endow it with causal semantics. The second one (SCM) starts from a graph and a set of functional assignments, and generates the observed distribution as the push-forward of an unobserved noise distribution. Finally, we cover a non-graphical approach (PO) popular in statistics.

\paragraph{Causal Graphical Models (CGMs)}
The graphical models framework
\cite{lauritzen1996graphical, Koller2009}
provides a compact way of representing joint probability distributions by encoding the dependence structure between variables in graphical form.
Directed graphical models
are also known as \textit{Bayesian networks} \cite{pearl1985bayesian}.
While they do not offer a causal interpretation per se---indeed, different graphical models can be compatible with the same distribution (cf.\ \cref{princ:reichenbach})---when edges are endowed with the notion of direct causal effect~(\cref{def:causal_effect}), we refer to them as causal graphical models (CGM)~\cite{Spirtes2000}.

\begin{definition}[CGM]
\label{def:CBN}
A CGM $\mathcal{M}=(G,p)$ over 
$n$ random variables $X_1,\dots,X_n$ consists of: (i) a directed acyclic graph (DAG) $G$ in which
directed edges ($X_j\rightarrow X_i$) represent a direct causal effect of $X_j$ on $X_i$; and (ii) a joint distribution $p(X_1, \dots ,X_n)$ which is Markovian w.r.t.\
$G$:
\begin{equation}
\label{eq:cf}
p(X_1, \dots, X_n)=\prod_{i=1}^n p(X_i \mid \PA_i)
\end{equation}
where $\PA_i = \{X_j : (X_j\rightarrow X_i)\in G\}$ denotes the set of parents, or direct causes, of $X_i$ in $G$.
\end{definition}
We will refer to~\eqref{eq:cf} as the \textit{causal (or disentangled) factorization}.
While many other \textit{entangled factorizations} are possible, e.g.,
\begin{equation}
\label{eq:non-cf}
p(X_1,\dots,X_n) = \prod_{i=1}^n p(X_i \mid X_{i+1},\dots,X_n),
\end{equation}
only \eq{eq:cf} decomposes the joint distribution into causal conditionals, or \textit{causal mechanisms}, $p(X_i \mid \PA_i)$, which can have a meaningful physical interpretation, rather than being mere mathematical objects such as the factors on the RHS of~\eqref{eq:non-cf}.

It turns out that~\eqref{eq:cf} is equivalent to the following condition.
\begin{definition}[Causal Markov condition]
\label{def:causalMarkov}
A distribution $p$ satisfies the causal Markov condition w.r.t.\ a DAG $G$ if every variable is conditionally independent of its non-descendants in $G$ given its parents in $G$. 
\end{definition}
\Cref{def:causalMarkov} 
can equivalently be expressed in terms of \textit{d-separation}, a graphical criterion for directed graphs \cite{Pearl2009}, by saying that \textit{d-separation in $G$ implies (conditional) independence in $p$}.
\looseness-1 
The causal Markov condition thus provides a link between properties of $p$ and~$G$.
\begin{figure}
    \centering
    \subfloat[$G$]{
        \centering
        \begin{tikzpicture}
        \centering
        \node (X_1) [obs, scale=\nodescale] {$X_1$};
        \node (X_2) [obs, yshift=-\yshift, xshift=-0.5*\xshift, scale=\nodescale] {$X_2$};
        \node (X_3) [obs, yshift=-\yshift, xshift=0.5*\xshift, scale=\nodescale] {$X_3$};
        \edge {X_1} {X_2,X_3};
        \edge {X_2} {X_3};
        \end{tikzpicture}
    }
    \qquad     \qquad \qquad     \qquad
    \subfloat[$G'$]{
        \centering
        \begin{tikzpicture}
        \centering
        \node (X_1) [obs, scale=\nodescale] {$X_1$};
        \node (X_2) [det, yshift=-\yshift, xshift=-0.5*\xshift, scale=\nodescale] {$x_2$};
        \node (X_3) [obs, yshift=-\yshift, xshift=0.5*\xshift, scale=\nodescale] {$X_3$};
        \edge {X_1} {X_3};
        \edge {X_2} {X_3};
        \end{tikzpicture}
    }
    \qquad     \qquad \qquad     \qquad
    \subfloat[$G''$]{
        \centering
        \begin{tikzpicture}
        \centering
        \node (X_1) [obs, scale=\nodescale] {$X_1$};
        \node (X_2) [obs, yshift=-\yshift, xshift=-0.5*\xshift, scale=\nodescale] {$X_2$};
        \node (X_3) [det, yshift=-\yshift, xshift=0.5*\xshift, scale=\nodescale] {$x_3$};
        \edge {X_1} {X_2};
        \end{tikzpicture}
    }
    \caption[Predicting interventions with causal Bayesian networks]{(a) A directed acyclic graph (DAG) $G$ over three variables. A causal graphical model $(G,p)$ with causal graph $G$ and observational distribution $p$ can be used to answer interventional queries using the concept of \textit{graph surgery}: when a variable is intervened upon and set to a constant (white diamonds), this removes any influence from other variables, captured graphically by removing all incoming edges. (b) and (c) show post-intervention graphs $G'$ and $G''$ for $do(X_2:=x_2)$ and $do(X_3:=x_3)$, respectively. (An intervention on $X_1$ would leave the graph unaffected.)
    \commentout{
    Interventional distributions are computed as conditional distributions of $p$, factorised over the corresponding post-intervention graph: e.g., $p_G(X_3|do(X_2:=x_2))=p_{G'}(X_3|X_2=x_2)$, where the subscript indicates the graph w.r.t.\ which $p$ factorises.}
    }
    \label{fig:DAGs}
\end{figure}
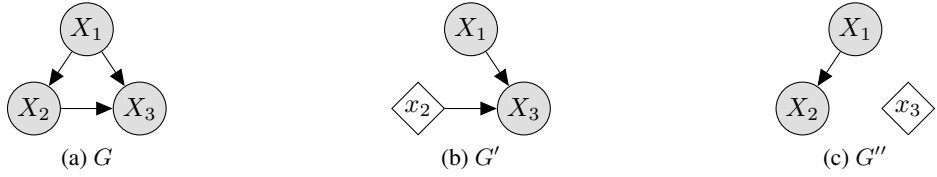

What makes CGMs causal is the interpretation of edges as cause-effect relationships which enables
reasoning about the outcome of interventions using the \textit{do-operator}~\cite{Pearl2009} and the concept of \textit{graph surgery}~\cite{Spirtes2000}. 
The central idea is that intervening on a variable, say by externally forcing it to take on a particular value, renders it independent of its causes and breaks their causal influence on it, see~\cref{fig:DAGs} for an illustration.
For example, if a gene is knocked out, it is no longer influenced by other genes that were previously regulating it; instead, its activity is now solely determined by the intervention.
This is fundamentally different from conditioning since passively observing the activity of a gene provides information about its driving factors (i.e., its direct causes).

To emphasize this difference between passive observation and active intervention, Pearl~\cite{Pearl2009} introduced the notation $do(X:=x)$ to denote an intervention by which variable $X$ is set to value $x$.
The term \textit{graph surgery} refers to the idea that the effect of such an intervention can be captured in the form of a modification to the original graph by removing all incoming edges to the intervened variable.
Interventional queries can then be answered by performing probabilistic inference in the modified post-intervention graph which typically implies additional (conditional) independences due to the removed edges.

\begin{example}
The interventional distribution $p(X_3\mid do(X_2=x_2))$ for the CGM in~\cref{fig:DAGs} is obtained via probabilistic
inference w.r.t.\ the post-intervention graph
$G'$ where~$X_1\independent X_2$:
\begin{align}
    p(X_3|do(X_2:=x_2))&=\sum_{x_1\in\mathcal{X}_1}p(x_1)p(X_3|x_1,x_2)
    \label{eq:intervention}\\
    &\neq \sum_{x_1\in\mathcal{X}_1}p(x_1\mid x_2)p(X_3\mid x_1,x_2)=p(X_3\mid x_2)
    \label{eq:conditioning}
\end{align}
It differs from the conditional $p(X_3\mid x_2)$ for which inference in done over $G$ where $X_1\not\independent X_2$.
Note the marginal $p(x_1)$ in~\eqref{eq:intervention}, in contrast to the conditional $p(x_1\mid x_2)$ in~\eqref{eq:conditioning}:
this is precisely the link which is broken by the intervention $do(X_2:=x_2)$, see~\cref{fig:DAGs}b.
The RHS of \eqref{eq:intervention} is an example of covariate adjustment: it controls for the confounder $X_1$ of the causal effect of $X_2$ on $X_3$, see~\cref{sec:reasoning} for more details on adjustment and computing interventions.
\end{example}

CGMs have been widely used in constraint- and score-based approaches to causal discovery \cite{Spirtes2000, heckerman1995learning} which we will discuss in~\cref{ch:2-discovery}.
Due to their conceptual simplicity, they are a useful and intuitive model for reasoning about interventions.
However, their capacity as a causal model is limited in that they do not support
\textit{counterfactual} reasoning, which is better addressed by the two causal modeling frameworks which we will discuss next.

\paragraph{Structural Causal Models (SCMs)}
Structural Causal Models, also referred to as functional causal models or non-parametric structural equation models, %
have ties to the graphical approach presented above, but rely on using directed functional parent-child relationships rather than causal conditionals. \looseness-1 While conceptually simple in hindsight, this constituted a major step in the understanding of causality, as later expressed by \cite{Pearl2009}~(p.\ 104):
\begin{quote}{``We played around with the possibility of replacing the parents-child relationship $p(X_i\mid \PA_i)$ with its functional counterpart $X_i=f_i(\PA_i,U_i)$ and, suddenly, everything began to fall into place: We finally had a mathematical object to which we could attribute familiar properties of physical mechanisms instead of those slippery epistemic probabilities $p(X_i \mid \PA_i)$ with which we had been working so long in the study of Bayesian networks.''}
\end{quote}

\begin{definition}[SCM]
\label{def:SCM}
An
SCM $\mathcal{M}=(\mathbf{F}, p_\mathbf{U})$ over a set $\mathbf{X}$ of $n$ random variables $X_1, \dots, X_n$ consists of (i) a set $\mathbf{F}$ of $n$ assignments (the structural equations), 
 \begin{equation}
 \label{eq:structural_eqs}
          \mathbf{F} = \{X_i := f_i(\PA_i,U_i)\}_{i=1}^n
 \end{equation}
where $f_i$ are deterministic functions computing each variable $X_i$ from its causal parents $\PA_i\subseteq \mathbf{X}\setminus \{X_i\}$ and an exogenous noise variable $U_i$; and (ii) a joint distribution $p_\mathbf{U}(U_1, \dots, U_n)$ over the exogenous noise variables.
\end{definition}
The paradigm of SCMs views the processes $f_i$ by which each observable $X_i$ is generated from others as a physical mechanism.
All randomness comes from the unobserved (also referred to as \emph{unexplained}) noise terms $U_i$ which capture both possible stochasticity of the process, as well as uncertainty due to unmeasured parts of the system.

Note also the assignment symbol ``$:=$'' which is used instead of an equality sign to indicate the asymmetry of the causal relationship: the LHS quantity is defined to take on the RHS value.
For example, we cannot simply rewrite a structural equation $X_2 :=f_2(X_1, U_2)$ as $X_1=g(X_2, U_2)$ for some $g$, as would be the case for a standard (invertible) equation.

In parametric, linear form (i.e., with linear $f_i$), SCMs are also known as structural equation models and have a long history in path analysis \cite{wright1921correlation} and economics \cite{haavelmo1944probability, hoover2001causality}.

Each SCM induces a corresponding causal graph via the input variables to the structural equations which is useful as a representation and provides a link to CGMs.

\begin{definition}[Induced causal graph]
\label{def:inducedgraph}
     The causal graph $G$ induced by an SCM $\mathcal{M}$ is the directed graph with vertex set $\mathbf{X}$ and a directed edge from each vertex in $\PA_i$ to $X_i$ for all $i$.
\end{definition}

\begin{example}%
\label{ex:scm}
Consider an SCM over $\mathbf{X}=\{X_1,X_2,X_3\}$ with some $p_\mathbf{U}(U_1,U_2,U_3)$ and
\begin{align*}
    X_1 := f_1(U_1), 
    \qquad \qquad 
    X_2 := f_2(X_1, U_2), 
    \qquad \qquad 
    X_3 := f_3(X_1, X_2, U_3).
\end{align*}
Following~\cref{def:inducedgraph}, the induced graph then corresponds to $G$ in~\cref{fig:DAGs}.
\end{example}

\Cref{def:SCM} allows for a rich class of causal models, including ones with cyclic causal relations and ones which do not obey the causal Markov condition~(\cref{def:causalMarkov}) due to complex covariance structures between the noise terms.
While work exists on such cyclic or confounded SCMs~\cite{bongers2021foundations}%
,
it is common to make the following two assumptions.

\begin{assumption}[Acyclicity]\label{ass:acyclicity}
The induced graph $G$
is a DAG: it does not contain cycles.
\end{assumption}

\begin{assumption}[Causal sufficiency/no hidden confounders]\label{ass:causalsufficiency}
The $U_i$ are jointly
independent, i.e., their  distribution factorises: $p_\mathbf{U}(U_1, \dots, U_n)=p_{U_1}(U_1)\times \dots \times p_{U_n}(U_n)$.
\end{assumption}

\Cref{ass:acyclicity} implies\footnote{Acyclicity is a sufficient, but not a necessary condition.} the existence of
a well-defined, unique (observational) distribution over $\mathbf{X}$ from which
we can draw via \textit{ancestral sampling}:\footnote{Since neither $\mathbf{F}$ nor $p$ are known a priori, ancestral sampling should be seen as a hypothetical sampling procedure; inference and learning are generally still necessary.}
first, we draw 
the noise variables from $p_\mathbf{U}$, and then we iteratively compute the corresponding $X_i$'s in topological order of the induced DAG (i.e., starting at the root node of the graph), substituting previously computed $X_i$ into the structural equations where necessary.
\commentout{E.g., for~\cref{ex:scm}, we would first draw $(U_1,U_2,U_3)\sim p_\mathbf{U}$, and then compute $X_1$, $X_2$, and $X_3$ (in this order).
Since the $f_i$ are deterministic, all $X_i$ are uniquely determined once the noise variables are fixed.}
Formally, the (observational) distribution $p(X_1,\dots,X_n)$ induced by an SCM under~\cref{ass:acyclicity} is defined as the push-forward of the noise distribution $p_U$ through the structural equations~$\mathbf{F}$.
Under~\cref{ass:causalsufficiency}, the causal conditionals are thus given by
\begin{equation}
\label{eq:induced distribution}
p(X_i\mid \PA_i=\pa_i):=p_{U_i}(f_{\pa_i}^{-1}(X_i)) \quad \text{for} \quad i=1,\dots,n,
\end{equation}
where $f_{\pa_i}^{-1}(X_i)$ denotes the pre-image of $X_i$ under $f_i$ for fixed $\PA_i=\pa_i$.

\Cref{ass:causalsufficiency} rules out the existence of hidden confounders because any unmeasured variables affecting more than one of the $X_i$ simultaneously would constitute 
a dependence between some of the noise terms (which account for any external, or exogenous, influences not explained by the observed $X_i$). 
In combination with~\cref{ass:acyclicity}, \cref{ass:causalsufficiency} (also known as \textit{causal sufficiency}), thus ensures that the distribution induced by an SCM factorises according to its induced causal graph as in~\eqref{eq:cf}.
\looseness-1 In  other words, it guarantees that the causal Markov condition is satisfied w.r.t.\ the induced causal graph \cite{Pearl2009}.
Below, unless explicitly stated otherwise, we will assume causal sufficiency.

Due to the conceptual similarity between interventions and the assignment character of structural equations, the computation of interventional distributions fits in naturally into the SCM framework.
To model an intervention, we simply replace the corresponding structural equation and consider the resulting entailed distribution.

\begin{definition}[Interventions in SCMs]
\label{def:intervention}
An intervention $do(X_i:=x_i)$ in an SCM $\mathcal{M}=(\mathbf{F},p_U)$ is modeled by replacing the  $i^{th}$ structural equation in $\mathbf{F}$ by $X_i:=x_i$, yielding the intervened SCM $\mathcal{M}^{do(X_i:=x_i)}=(\mathbf{F}',p_U)$.
The interventional distribution~$p(\mathbf{X}_{-i}\mid do(X_i=x_i))$, where $\mathbf{X}_{-i}=\mathbf{X}\setminus \{X_i\}$, and intervention graph $G'$ are those induced by $\mathcal{M}^{do(X_i=x_i)}$.
\end{definition}
This way of handling interventions coincides with that for CGMs:
e.g., after performing $do(X_2:=x_2)$ in~\cref{ex:scm}, $X_1$ no longer appears in the structural equation for $X_2$, and the edge $X_1\rightarrow X_2$ hence disappears in the intervened graph, as is the case for $G'$ in~\cref{fig:DAGs}.

In contrast to CGMs, 
SCMs also provide a framework for \textit{counterfactual reasoning}.
While (i) observations describe what is passively seen or measured and (ii) interventions describe active external manipulation or experimentation, (iii) counterfactuals are statements about what would or could have been, given that something else was in fact observed.
These three modes of reasoning are sometimes referred to as the three rungs of 
the ``ladder of causation''~\cite{pearl2018book}.
As an example, consider the following counterfactual query: 
\begin{quote}
Given that patient $X$ received treatment A and their health got worse, what would have happened  if they had been given treatment B instead, \textit{all else being equal}?
\end{quote}
The ``all else being equal'' part 
highlights
the difference between interventions and counterfactuals:
observing the factual outcome (i.e., what actually happened) provides information about the background state of the system (as captured by the noise terms in SCMs)
which can be used to reason about alternative, counterfactual, outcomes.
This differs from an intervention where such background information is not available.
For example, observing that treatment A did not work may tell us that the patient has a rare condition and that treatment B would have therefore worked.
However, given that treatment A has been prescribed, the patient's condition may have changed, and B may no longer work in a future intervention.

Note that counterfactuals cannot be observed empirically by their very definition and are therefore unfalsifiable.
Some therefore consider them unscientific \cite{popper1959logic} or at least problematic \cite{dawid2000causal}.
On the other hand, humans seem to perform counterfactual reasoning in practice, developing this ability in early childhood \cite{buchsbaum2012power}.
Counterfactuals are computed in SCMs through the following three-step procedure:
\begin{enumerate}
\item 
Update the noise distribution to  its posterior given 
the observed evidence (``abduction'').
\item
Manipulate the structural equations to capture the hypothetical intervention (``action'').
\item 
Use the modified SCM to infer the quantity of interest (``prediction'').
\end{enumerate}

\begin{definition}[Counterfactuals in SCMs]
Given evidence $\mathbf{X}=\mathbf{x}$ observed from an SCM $\mathcal{M}=(\mathbf{F},p_U)$, the counterfactual SCM  $\mathcal{M}^{\mathbf{X}=\mathbf{x}}$ is obtained by updating $p_U$ with its posterior: $\mathcal{M}^{\mathbf{X}=\mathbf{x}}=(\mathbf{F},p_{U\mid \mathbf{X}=\mathbf{x}})$.
Counterfactuals are then computed by performing interventions in the counterfactual SCM $\mathcal{M}^{\mathbf{X}=\mathbf{x}}$, see~\cref{def:intervention}.
\end{definition}

Note that while computing interventions only involved manipulating the structural equations, counterfactuals  also involve 
updating the noise distribution, highlighting 
the conceptual difference between the two.
Updating $p_U$ requires knowledge of the interaction between noise and observed variables, i.e., of the structural equations, which explains why additional assumptions are necessary. 
Note that the updated noise variables no longer need to be independent, even if the original system was causally sufficient~(\cref{ass:causalsufficiency}).

\begin{example}[Computing counterfactuals with SCMs]
Consider an SCM $\mathcal{M}$ defined by
\begin{align}
    X := U_X, \qquad
    Y := 3X + U_Y, \qquad 
    U_X,U_Y\sim\mathcal{N}(0,1). \label{eq:factual_Y}
\end{align}
Suppose we observe $X=2$ and $Y=6.5$ and want to answer the counterfactual ``what would $Y$ have been, had $X=1$?'', i.e., we are interested in $p(Y_{X=1}\mid X=2, Y=6.5)$.
Updating the noise using the observed evidence via~\eqref{eq:factual_Y},
we obtain
the counterfactual SCM $\mathcal{M}^{X=2,Y=6.5}$,
\begin{align}
    X := U_X, \qquad 
    Y := 3X + U_Y, \qquad 
    U_X\sim\delta(2), \qquad 
    U_Y\sim\delta(0.5), \label{eq:counterfactual_Y}
\end{align}
where $\delta(\cdot)$ denotes the Dirac delta measure.
\looseness-1
Performing the intervention  $do(X:=1)$ 
in \eqref{eq:counterfactual_Y} then gives the result $p(Y_{X=1}\mid X=2, Y=6.5)=\delta(3.5)$, i.e., ``$Y$ would have been $3.5$''.
This differs from the interventional distribution $p(Y\mid do(X=1))=\mathcal{N}(3,1)$, since
the factual observation helped determine the background state~($U_X=2, U_Y=0.5$).
\end{example}

The SCM viewpoint is intuitive and
lends itself well to
studying restrictions on function classes to enable induction (\cref{sec:slt}). 
For this reason, we will mostly focus on SCMs in the subsequent sections.

\paragraph{Potential Outcomes (PO)}
The potential outcomes framework was initially proposed by Neyman~\cite{neyman1923application} for randomized studies \cite{fisher1937design}, and later popularized and extended to observational settings by Rubin~\cite{rubin1974estimating} and others.
It is popular within statistics and epidemiology and perhaps best understood in the context of the latter. This 
is
also reflected in its terminology:
in the most common setting, we consider a binary treatment variable $T$, with $T=1$ and $T=0$ corresponding to treatment and control, respectively, whose causal effect on an outcome variable~$Y$ (often  a measure of health) is of interest.

One interpretation of the PO framework consistent with its roots in statistics is to view \textit{causal inference as a missing data problem}.
In the PO framework, for each individual (or unit) $i$ and treatment value $t$ there is a PO, or potential response, denoted $Y_i(t)$ capturing what would happen if individual $i$ received treatment $t$.
The POs are considered deterministic quantities in the sense that for a given individual $i$, $Y_i(1)$ and $Y_i(0)$ are fixed and all randomness in the realized outcome $Y_i$ stems from randomness in the treatment assignment:
\begin{equation}
\label{eq:factual_outcome}
Y_i=TY_i(1)+(1-T)Y_i(0).
\end{equation}
To decide whether patient $i$ should receive treatment, we need to reason about the \textit{individualized treatment effect} (ITE) $\tau_i$ as captured by the difference of the two POs.
\begin{definition}[ITE]
\label{def:ITE}
The ITE for individual $i$ under a binary treatment is defined as
\begin{equation}
\label{eq:unit-causal-effect}
\tau_i=Y_i(1)-Y_i(0).
\end{equation}
\end{definition}

The \textit{``fundamental problem of causal inference''} \cite{holland1986statistics} is that only one of the POs is ever observed for each $i$.
The other, unobserved PO becomes a counterfactual:%
\begin{equation}
\label{eq:counterfactual_outcome}
Y^{\textsc{cf}}_i=(1-T)Y_i(1)+TY_i(0).
\end{equation}
Consequently, $\tau_i$ can never be measured or computed from data, i.e., it is not identifiable (without further assumptions), as illustrated in~\cref{tab:PO}.

\begin{table}[]
    \begin{minipage}[c]{0.6\textwidth}
     \caption[Potential outcomes and the fundamental problem of causal inference]{
    Causal inference as a missing data problem: for each individual $i$ (rows),
    only the PO $Y_i(T_i)$ corresponding to the assigned treatment $T_i$ is observed; the other PO is a 
    counterfactual. Hence, the unit-level causal effect $\tau_i=Y_i(1)-Y_i(0)$ is non-identifiable. 
    }
    \label{tab:PO}
    \end{minipage}
    \hfill
    \begin{minipage}[c]{0.38\textwidth}
    \centering
    \begin{tabular}{cccccc}
        \toprule
        $i$ & $T_i$ & $Y_i(1)$ & $Y_i(0)$ & $\tau_i$\\
        \midrule
        $1$ & 1 & 7 & \textbf{?} & ? \\
        $2$ & 0 & \textbf{?} & 8 & ?\\
        $3$ & 1 & 3 & \textbf{?} & ?\\
        $4$ & 1 & 6 & \textbf{?} & ?\\
        $5$ & 0 & \textbf{?} & 4 & ?\\
        $6$ & 0 & \textbf{?} & 1 & ?\\
        \bottomrule
    \end{tabular}
    \end{minipage}
\end{table}
Implicit in 
the form of~\eqref{eq:factual_outcome} and~\eqref{eq:counterfactual_outcome} are the following two
assumptions.
\begin{assumption}[Stable unit treatment value; SUTVA]
\label{ass:sutva}
The observation on one unit should be unaffected by the particular assignment of treatments to
the other units
\cite{cox1958planning}.
\end{assumption}
\begin{assumption}[Consistency]
\label{ass:consistency}
If individual $i$ receives treatment $t$, then the observed outcome is $Y_i=Y_i(t)$, i.e., the potential outcome for $t$.
\end{assumption}
\Cref{ass:sutva}
is usually understood as (i) units do not interfere, and (ii) there is only one treatment level per group (treated or control) leading to well-defined POs \cite{imbens2015causal}.
It can be violated, e.g., through (i) population dynamics such as herd immunity from vaccination or (ii) technical errors or varying within-group dosage, respectively.
However, for many situations such as controlled studies it can be a reasonable assumption,
and we can then view different units as independent samples from a population.

So far, we have considered POs for a given unit as deterministic quantities.
However, most times it is impossible to fully characterize a unit, e.g., when dealing with complex subjects such as humans.
Such lack of complete information introduces uncertainty, so that POs are often instead treated as random variables.
This 
parallels the combination of deterministic structural equations with exogenous noise variables in SCMs.\footnote{
When all noise variables in an SCM are fixed, 
the other variables are uniquely determined;
without complete background knowledge, on the other hand, they are random.}
Indeed, there is a
equivalence between POs and SCMs \cite{Pearl2009}:
\begin{equation*}
Y_i(t) = Y\mid do(T=t) \qquad  \text{in an SCM with} \qquad  \mathbf{U}=\mathbf{u}_i,   
\end{equation*}
An individual in the PO framework thus corresponds to a particular instantiation of the $U_j$ in an SCM: the outcome is deterministic given $\mathbf{U}$, but since we do not observe $\mathbf{u}_i$ (nor can we characterize a given individual based on observed covariates), the counterfactual outcome is treated as a random variable.
In practice, all we observe is a featurised description $\mathbf{x}_i$ of an individual $i$ and have to reason about expected POs, $\mathbb{E}[Y(1), Y(0)\mid \mathbf{x}_i]$. 

Another common assumption 
is that of no hidden confounders
which we have
already encountered in form
of the causal Markov condition~(\cref{def:causalMarkov}) for CGMs and causal sufficiency~(\cref{ass:causalsufficiency}) for SCMs.
In the PO framework this becomes no hidden confounding between treatment and outcome and is referred to as (conditional) ignorability.

\begin{assumption}[Conditional ignorability]
\label{ass:ignorability}
Given a treatment $T\in\{0,1\}$, potential outcomes $Y(0),Y(1)$, and observed covariates $\mathbf{X}$, we have:
\begin{equation}
    Y(0) \independent T\mid\mathbf{X} \quad \text{and} \quad Y(1) \independent T\mid \mathbf{X}.
\end{equation}
\end{assumption}

\commentout{
POs and their popularization by Rubin and Rosenbaum in the 1970s revived causal inference as a field within statistics, and POs continue to be the dominant framework used in many disciplines.
}
The PO framework is tailored toward studying the (confounded) effect of a typically binary treatment variable on an outcome and
is mostly used for causal reasoning, i.e., estimating individual and population level causal effects~(\cref{sec:reasoning}).
In this context, it is sometimes seen as an advantage that an explicit graphical representation is not needed.
At the same time, the lack of a causal graph and the need for special treatment and outcome variables make POs rather unsuitable for causal discovery where other frameworks prevail.
\commentout{
Moreover, assumptions of the PO framework are usually phrased as a list of (conditional) independence statements  (involving POs), and can be tricky to understand without drawing a graph.

\jvk{Keep any of this?}
\paragraph{Single-world intervention graphs}
With the aim to reconcile the graphical and counterfactual approaches to causality, \cite{richardson2013single} have proposed a new framework termed \textit{single world intervention graphs} (SWIGs).
SWIGs can be seen as an extension of DAGs and are obtained by introducing a \textit{node-splitting} operation.
Node splitting is applied to variables which are intervened upon, and works by splitting them into two disconnected parts: one representing the original variable, $X$ say, and the other representing the value set by intervention, $x$.
Descendants of $X$ are newly labeled to receive $x$ as an argument.
The authors argue that SWIGs are particularly helpful for testing conditional independence statements involving PO (such as ignorability) using tools developed for DAGs, thereby combining the best of both worlds.
}

\section{Independent Causal Mechanisms\label{sec:icm}}
Let us consider the disentangled factorization \eq{eq:cf} of the joint distribution $p(X_1,\dots,X_n)$. This factorization according to the causal graph is possible whenever the $U_i$ are independent. We now consider an additional notion of independence, concerning how the factors in \eq{eq:cf} relate to one another. 

Consider a dataset that consists of altitude $A$ and average annual temperature $T$ of weather stations \cite{PetJanSch17}. Supposed we find that $A$ and $T$ are correlated, which we attribute due to the fact that the altitude has a causal effect on the temperature. Suppose we had two such datasets, one for Austria and one for Switzerland. The two joint distributions may be different, since the marginal distributions $p(A)$ over altitudes will likely differ. The conditionals $p(T\mid A)$, however, may be rather similar, since they reflect physical mechanisms generating temperature from altitude. 
The causal factorization $p(A)p(T\mid A)$ thus contains a component $p(T\mid A)$ that generalizes across countries, while the entangled factorization $p(T)p(A\mid T)$ does not.  
A similar reasoning applies when we consider interventions in a system. For a model to correctly predict the effect of interventions, it needs to have components that are robust when moving from an observational distribution to certain {\em interventional} distributions.

The above insights can be stated as follows \cite{Schoelkopf2012,PetJanSch17}:

\begin{principle}[Independent Causal Mechanisms (ICM)]
\label{princ:icm}
The causal generative process of a system's variables is composed of autonomous modules that do not inform or influence each other.
In the probabilistic case, this means that the conditional distribution of each variable given its causes (i.e., its mechanism) does not inform or influence the other mechanisms.
\end{principle}
This principle subsumes several notions important to causality, including separate intervenability of causal variables, modularity and autonomy of subsystems, and invariance~\cite{Pearl2009,PetJanSch17}.
In the two-variable case, it reduces to an independence between the cause distribution and the mechanism producing the effect from the cause.

Applied to the causal factorization \eq{eq:cf}, the principle tells us that the factors should be independent in two senses:
\begin{labeling}{(\textit{information~~~~~})}
\item[(\textit{influence})] changing (or intervening upon) one mechanism $p(X_i\mid\PA_i)$ does not change the other mechanisms $p(X_j\mid\PA_j)$ ($i\ne j$), and 
\item[(\textit{information})] knowing some mechanisms $p(X_i\mid\PA_i)$ ($i\ne j$) does not give us information about a mechanism $p(X_j\mid\PA_j)$. 
\end{labeling}
We view any real-world distribution as a product of causal mechanisms. A change in such a distribution (e.g., when moving from one setting/domain to a related one) will always be due to a change in at least one of those mechanisms. Consistent with \cref{princ:icm}, we hypothesize \cite{scholkopfetal21}:

\begin{principle}[Sparse Mechanism Shift (SMS)]
\label{pri:scsh}
Small distribution changes tend to manifest themselves in a sparse or local way in the causal/disentangled factorization~\eq{eq:cf}, i.e., they should usually not affect all factors simultaneously.
\end{principle}
In contrast, in a non-causal factorization, e.g., \eq{eq:non-cf}, many terms will be affected simultaneously if we change one of the physical mechanisms responsible for a system's statistical dependences. Such a factorization may thus be called \textit{entangled}. The notion of disentanglement has recently gained popularity in machine learning \cite{bengio2013representation,higgins2016beta,1811.12359,Suter.1811.00007}, sometimes loosely identified with statistical independence. The notion of invariant, autonomous, and independent mechanisms has appeared in various guises throughout the history of causality research, see \cite{Aldrich89,Pearl2009,Schoelkopf2012,PetJanSch17,scholkopfetal21}.

\commentout{The notion of invariant, autonomous, and independent mechanisms has appeared in various guises throughout the history of causality research. Early work on this was done by \cite{Haavelmo1944}, stating the assumption that changing one of the structural assignments leaves the other ones invariant. 
\cite{Hoover06} attributes to Herb Simon the {\em invariance criterion}: the true causal order is the one that is invariant under the right sort of intervention.
\cite{Aldrich89} provides an overview of the historical development of these ideas in economics. He argues that the ``most basic question one can ask about a relation should be: How autonomous is it?'' \cite{Frisch1948}. \cite{Pearl2009} discusses autonomy in detail, arguing that a causal mechanism remains invariant when other mechanisms are subjected to external influences. He points out that causal discovery methods may best work ``in longitudinal studies conducted under slightly varying conditions, where accidental independencies are destroyed and only structural independencies are preserved.'' 
Overviews are provided by \cite{Aldrich89,Hoover06,Pearl2009}, and \cite{PetJanSch17} (Sec.~2.2).}

\paragraph{Measures of Dependence of Mechanisms}
Note that the dependence of two mechanisms $p(X_i\mid\PA_i)$ and $p(X_j\mid \PA_j)$ does not coincide with the statistical dependence of the random variables $X_i$ and $X_j$. Indeed, in a causal graph, many of the random variables will be dependent even if all the mechanisms are independent.

\begin{figure}[t]
\begin{minipage}[c]{0.25\textwidth}
\caption{\label{fig:igci} If $p(x)$ and $f$ are chosen independently, then peaks of $p(y)$ tend to occur in regions where
$f$ has small slope.
As a result, $p(y)$ contains information about $f^{-1}$ (figure from \cite{PetJanSch17}).
}
\end{minipage}
\hfill
\begin{minipage}[c]{0.7\textwidth}
\centerline{
\begin{tikzpicture}[xscale=1., yscale=1.]
\small
  \draw (0.2,2.8) node(y) {$y$};
  \draw (3.8,0.2) node(x) {$x$};
  \draw (3.4,2.8) node(x) {$f(x)$};
  \draw[->] (-0.1,0) -- (4.2,0); 
  \draw[->] (0,-0.1) -- (0,3.2); 
  \foreach \j in {1,2,3,4}{
  	\draw (\j,-0.07) -- (\j,0.07); 
  }
  \foreach \j in {1,2,3}{
  	\draw (-0.07,\j) -- (0.07,\j); 
  }  
\draw [rounded corners=5pt] (0.2,0.2) -- (1,1.2) -- (2.8,1.3) -- (4,3);
\filldraw [fill=gray!20, rounded corners=3pt] 
(0.2,-0.78) -- (0.3,-0.62) -- (0.6,-0.3) -- (1.0,-0.52) -- (1.4,-0.28) -- (1.8,-0.22) -- (2.2,-0.38) -- (2.6,-0.26) -- (3.0,-0.5) -- (3.4,-0.21) -- (3.8,-0.68) -- (4.0,-0.78);
  \draw (1.8,-0.6) node(px) {$p(x)$};
\filldraw [fill=gray!20, rounded corners=3pt] 
(-1.50,0.2) -- (-1.46,0.3) -- (-1.01,0.6) -- (-0.97,0.8) -- (-1.09,1.1) -- (-0.05,1.25) --  (-0.99,1.4) -- (-1.1,1.8) -- (-0.88,2.2) -- (-1.25,2.6) -- (-1.48,2.9) -- (-1.5,3.00);
  \draw (-1.32,2.1) node(py) [rotate=-90] {$p(y)$};
\end{tikzpicture}
$\qquad$
}
\end{minipage}
\end{figure}
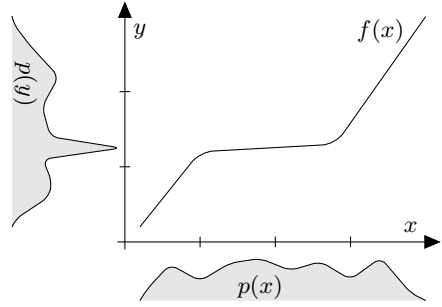

\commentout{Intuitively speaking, the independent noise terms $U_i$ provide and parametrize the uncertainty contained in the fact that a mechanism $p(X_i\mid \PA_i)$ is non-deterministic, and thus ensure that each mechanism adds an independent element of uncertainty. We thus like to think of the ICM~\cref{princ:icm} as containing the independence of the unexplained noise terms in an SCM \eqref{eq:structural_eqs} as a special case.\footnote{See also \cite{PetJanSch17}. Note that one can also implement the independent mechanisms principle by assigning independent priors for the causal mechanisms. We can view ICM as a meta-level independence, akin to assumptions of time-invariance of the laws of physics \cite{Bohm}.}
However, it goes beyond this, as the following example illustrates.} 
Consider two variables and structural assignments $X:= U$ and $Y:=f(X)$. I.e., the cause $X$ is a noise variable (with density $p(x)$),
and the effect $Y$ is a deterministic function of the cause. Let us moreover assume that the ranges of $X$ and $Y$ are both $[0,1]$, and $f$ is strictly monotonically increasing.
The ICM principle then reduces to an independence of $p(x)$ and $f$. Let us consider $p(x)$ and the derivative $f'$ as random variables on the probability space $[0,1]$ with Lebesgue measure, and use their correlation as a measure of dependence of mechanisms.
It can be shown that for $f\ne id$, independence of $p(x)$ and $f'$ implies dependence between $p(y)$ and $(f^{-1})'$ (see~\cref{fig:igci}). Other measures are possible and admit information-geometric interpretations. Intuitively, under the ICM assumption~(\cref{princ:icm}), the ``irregularity'' of the effect distribution becomes a \emph{sum} of (i) irregularity already present in the input distribution and (ii) irregularity introduced by the mechanism $f$, i.e., the irregularities of the two mechanisms add up rather than (partly) compensating each other. This would not be the case in the opposite (``anticausal'') direction (for details, see \cite{Janzingetal12}).
Other dependence measures have been proposed for high-dimensional linear settings and time series \cite{JanHoySch10,Shajarisales15,BesShaSchJan18,JanSch18b}.

\paragraph{Algorithmic Independence}
So far, we have discussed links between causal and statistical structures. The fundamental of the two is the causal structure, since it captures the physical mechanisms that generate statistical dependences in the first place. The statistical structure is an epiphenomenon that follows if we make the unexplained variables random.
It is awkward to talk about the (statistical) information contained in a mechanism, since deterministic functions in the generic case neither generate nor destroy information. This motivated us to devise an algorithmic model of causal structures in terms of Kolmogorov complexity \cite{JanSch10}. The Kolmogorov complexity (or algorithmic information) of a bit string is essentially the length of its shortest compression on a Turing machine, and thus a measure of its information content. Independence of mechanisms can be defined as vanishing mutual algorithmic information: two conditionals are considered independent if knowing (the shortest compression of) one does not help achieve a shorter compression of the other one.

Algorithmic information theory is an elegant framework for non-statistical graphical models. Just like statistical CGMs are obtained from SCMs by making the unexplained variables $U_i$ random, we obtain algorithmic CGMs by turning the $U_i$ into bit strings (jointly independent across nodes), and viewing the node $X_i$ as the output of a fixed Turing machine running the program $U_i$ with input $\PA_i$. Similar to the statistical case, one can define a local causal Markov condition, a global one in terms of d-separation, and a decomposition of the joint Kolmogorov complexity in analogy to \eq{eq:cf}, and prove that they are implied by the SCM~\cite{JanSch10}. 
\looseness-1 This approach shows that concepts of causality are not intrinsically tied to statistics: causality is about \emph{mechanisms} governing flow of information which may or may not be statistical.

The assumption of algorithmically independent mechanisms has interesting implications for physics: it implies the second law of thermodynamics (i.e., the arrow of time). Consider a process where an incoming ordered beam of photons (the cause) is scattered by an object (the mechanism). Then the outgoing beam (the effect) contains information about the object. 
\looseness-1 
Microscopically, the time evolution is reversible; however, the photons contain information about the object only \emph{after} the scattering. What underlies Loschmidt's paradox~\cite{loschmidt1876ueber}?

The asymmetry can be explained by applying the ICM~\cref{princ:icm} to initial state and system dynamics, postulating that the two be algorithmically independent, i.e., knowing one does not allow a shorter description of the other one. The Kolmogorov complexity of the system's state can then be shown to be non-decreasing under time evolution \cite{Janzing2016}. If we view Kolmogorov complexity as a measure of entropy, this means that the entropy of the state can only stay constant or increase, amounting to the second law of thermodynamics.

Note that the resulting state after time evolution is clearly \emph{not} independent of the system dynamic: it is precisely the state that, when fed to the inverse dynamics, would return us to the original (ordered) state. 
\section{Levels of Causal Modeling\label{sec:levels}}
Coupled differential equations are the canonical way of modeling physical phenomena. They allow us to predict the future behavior of a system, to reason about the effect of interventions, and---by suitable averaging procedures---to predict {\em statistical} dependences that are generated by a coupled time evolution.
They also allow us to gain insight into a system, explain its functioning, and, in particular, read off its causal structure.

Consider a coupled set of ordinary differential equations
\begin{equation}\label{eq:ode}
\frac{d\x}{dt} = f(\x), \; \x \in \RR^d,
\end{equation}
with initial value $\x(t_0)=\x_0$. 
We assume that they correctly describe the physical mechanisms of a system.\footnote{I.e., they do not merely phenomenologically describe its time evolution without capturing the underlying mechanisms (e.g., due to unobserved confounding, or a form of coarse-graining that does not preserve the causal structure \cite{Rubensteinetal17,scholkopfetal21}).}
The Picard–Lindelöf theorem states that, at least locally, if $f$ is Lipschitz, there exists a unique solution $\x(t)$. This implies, in particular, that the immediate future of $\x$ is implied by its past values.
In terms of infinitesimal differentials $dt$ and $d\x = \x(t+dt)-\x(t)$, \eqref{eq:ode} reads:
\begin{equation}
\x(t+dt) = \x(t) + dt\cdot f(\x(t)).
\end{equation}
From this, we can ascertain which entries of the vector $\x(t)$ cause the future of others $\x(t+dt)$, i.e., the causal structure. 

Compared to a differential equation, a statistical model derived from the joint distribution of a set of (time-independent) random variables is a rather superficial description of a system. It exploits that some of the variables allow the prediction of others as long as the experimental conditions do not change. 
If we drive a differential equation system with certain types of noise, or if we average over time, 
statistical dependences between components of $\x$ may emerge, which can be exploited by machine learning. In contrast to the differential equation model, such a model does not allow us to predict the effect of interventions; however, its strength is that it can often be learned from data.

\begin{table}[t]
\caption{\label{t:taxonomy}Simple model taxonomy. The most detailed model (top) is a mechanistic or physical one, often in terms of differential equations. At the other end of the spectrum (bottom), we have a purely statistical model; this can be learned from data and  is useful for predictions but often provides little insight beyond modeling associations between epiphenomena.
\looseness-1 Causal models can be seen as intermediate descriptions, abstracting away from physical realism while retaining the power to answer certain interventional or counterfactual questions.
}
\vspace{.5em}
{\resizebox{\textwidth}{!}{
\begin{tabular}{ c  c  c  c  c  c }
\toprule
\multirow{2}{*}{\textbf{Model}} & \textbf{Predict in i.i.d.} & \textbf{Predict under distr.} & \textbf{Answer counter-} & \textbf{Obtain} & \textbf{Learn from}  \\
 & \textbf{setting} & \textbf{shift/intervention} & \textbf{factual questions} & \textbf{physical insight} & \textbf{data} \\
 \midrule
Mechanistic/physical  & yes & yes & yes & yes & ? \\
Structural causal  & yes & yes & yes & ? & ?\\
Causal graphical & yes & yes & no & ? & ?\\
Statistical & yes & no & no & no & yes\\
\bottomrule
\end{tabular}}}
\end{table}

\looseness-1 Causal modeling lies in between these two extremes. It aims to provide understanding and predict the effect of interventions. Causal discovery and learning tries to arrive at such models in a data-driven way, using only weak assumptions (see Table~\ref{t:taxonomy}, from~\cite{PetJanSch17,scholkopfetal21}).

While we may naively think that causality is always about time, most existing causal models do not (and need not) consider time. For instance, returning to our example of altitude and temperature, there is an underlying dynamical physical process that results in higher places tending to be colder. On the level of microscopic equations of motion for the involved particles, there is a temporal causal structure. 
\looseness-1 However, when we talk about dependence or causality between altitude and temperature, we need not worry about the details of this temporal structure; we are given a dataset where time does not appear, and we can reason about how that dataset would look if we were to intervene on temperature or altitude.

Some work exists trying to build bridges between these different levels of description. One can derive SCMs that describe the interventional behavior of a coupled system that is in an equilibrium state and perturbed in an adiabatic way \cite{MooJanSch13}, with generalizations to oscillatory systems~\cite{RubBonMooSch18}. In this work, an SCM arises as a high-level abstraction of an underlying system of differential equations. It 
can only be derived if suitable high-level variables can be defined \cite{Rubensteinetal17}, which in practice may well be the exception rather than the rule.

\section{Causal Discovery}
\label{ch:2-discovery}
Sometimes, domain knowledge or the temporal ordering of events can help constrain the causal relationships between variables: e.g., we may know that certain attributes like age or sex are not caused by others; treatments influence health outcomes; and events do not causally influence their past.
When such domain knowledge is unavailable or incomplete, we need to perform \textit{causal discovery}:
infer which variables causally influence which others, i.e., learn the causal structure 
(e.g., a DAG) 
from data.
Since experiments are often difficult and expensive to perform while observational (i.e., passively collected) data is abundant, causal discovery from observational data is of particular interest.

As discussed in~\cref{sec:from} in the context of the Common Cause~\cref{princ:reichenbach}, the case where we have two variables is already difficult since the same dependence can be explained by multiple different causal structures. One might thus wonder if the case of more observables is completely hopeless. Surprisingly, this is not the case: the problem becomes easier (in a certain sense) because there are nontrivial conditional independence properties \cite{Spohn78,Dawid79,GeiPea90} implied by a causal structure.
\looseness-1 We first review two classical approaches to the multi-variate setting before returning to the two-variable case.

\paragraph{Constraint-Based Methods}
Constraint-based approaches to causal discovery test which (conditional) independences can be inferred from the data and then try to find a graph which implies them.
They are therefore also known as independence-based methods. 
Such a procedure requires a way of linking properties of the data distribution $p$ to properties of the underlying causal graph $G$.
This link is known as the faithfulness assumption.

\begin{assumption}[Faithfulness]
\label{ass:faithfulness}
The only (conditional) independences satisfied by $p$ are those implied by the causal Markov condition~(\cref{def:causalMarkov}).
\end{assumption}

Faithfulness
can be seen as the converse of the causal Markov condition.
Together, they
constitute a one-to-one correspondence between graphical separation in $G$ and conditional independence in $p$. 
While the causal Markov condition
is satisfied 
by construction, faithfulness is an assumption which may be violated.
A classical example for a violation of faithfulness is when causal effects along different paths cancel.

\begin{example}[Violation of faithfulness]
Consider the SCM from~\cref{ex:scm} and let
\begin{align*}
    X_1 := U_1, 
    \qquad    \qquad
    X_2 := \alpha X_1 + U_2,
    \qquad    \qquad
    X_3 := \beta X_1 + \gamma X_2 + U_3 
\end{align*}
with $U_1,U_2,U_3\overset{i.i.d.}{\sim}\mathcal{N}(0,1)$.
By substitution, we obtain
    $X_3=(\beta + \alpha \gamma)X_1 + \gamma U_2+U_3$.
Hence $X_3\independent X_1$ whenever $\beta + \alpha \gamma=0$, even though this independence is not implied by the causal Markov condition over the induced causal graph $G$, see~\cref{fig:DAGs}.
Here, faithfulness is violated if the direct effect of $X_1$ on $X_3$ ($\beta$) and the indirect effect via $X_2$ ($\alpha\gamma$) cancel.
\end{example}

\begin{figure}[t]
\centering
 \subfloat[Chains]{
 \centering
        \begin{tikzpicture}
        \centering
        \node (X) [obs, scale=\nodescale] {$X$};
        \node (Y) [obs, xshift=\xshift, scale=\nodescale] {$Y$};
        \node (Z) [obs, xshift=2*\xshift, scale=\nodescale] {$Z$};
        \edge {X} {Y};
        \edge {Y} {Z};
        \node (X2) [obs, yshift=-\yshift, scale=\nodescale] {$X$};
        \node (Y2) [obs, xshift=\xshift, yshift=-\yshift, scale=\nodescale] {$Y$};
        \node (Z2) [obs, xshift=2*\xshift, yshift=-\yshift, scale=\nodescale] {$Z$};
        \edge {Y2} {X2};
        \edge {Z2} {Y2};
        \end{tikzpicture}
}
\qquad \qquad 
\subfloat[Fork]{
\centering
        \begin{tikzpicture}
        \centering
        \node (Y) [obs, scale=\nodescale] {$Y$};
        \node (X) [obs, yshift=-\yshift, xshift=-\xshift, scale=\nodescale] {$X$};
        \node (Z) [obs, yshift=-\yshift, xshift=\xshift, scale=\nodescale] {$Z$};
        \edge {Y} {X,Z};
        \end{tikzpicture}
}
\qquad \qquad 
\subfloat[Collider]{
        \centering
        \begin{tikzpicture}
        \centering
        \node (X) [obs, scale=\nodescale] {$X$};
        \node (Y) [obs, yshift=-\yshift, xshift=\xshift, scale=\nodescale] {$Y$};
        \node (Z) [obs, xshift=2*\xshift, scale=\nodescale] {$Z$};
        \edge {X,Z} {Y};
        \end{tikzpicture}
}%
    \caption[Common graph motifs and Markov equivalence]{Illustration of Markov equivalence using common graph motifs. The chains in (a) and the fork in (b) all imply the relation $X\independent Z\mid Y$ (and no others). They thus form a Markov equivalence class, meaning they cannot be distinguished using conditional independence testing alone. The collider, or v-structure, in (c) implies $X\independent Z$ (but $X\not\independent Z\mid Y$) and forms its own Markov equivalence class, so it can be uniquely identified from observational data. 
    For this reason, v-structures are helpful for causal discovery. It can be shown that two graphs are Markov equivalent iff.\ they share the same skeleton and v-structures.
    }
    \label{fig:Markov-equivalence}
\end{figure}
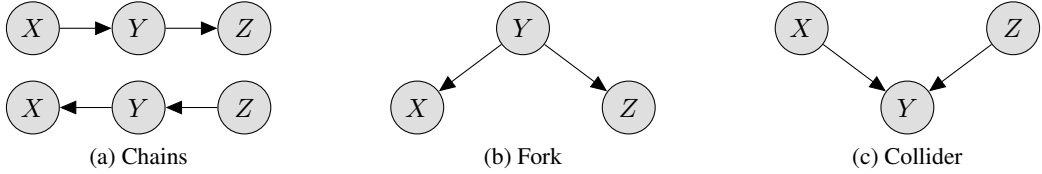

Apart from relying on faithfulness, a fundamental limitation to constraint-based methods is the fact that many different DAGs may encode the same d-separation / independence relations.
This is referred to as Markov equivalence and illustrated in~\cref{fig:Markov-equivalence}.

\begin{definition}[Markov equivalence]
Two DAGs are said to be Markov equivalent if they encode the same d-separation statements. The set of all DAGs encoding the same d-separations is called a Markov equivalence class.
\end{definition}

Constraint-based algorithms 
typically
first construct an undirected graph, or skeleton, which captures the (conditional) independences
found by testing, and then direct as many edges as possible using Meek's orientation rules~\cite{meek1995causal}.
The first step carries most of the computational weight and various algorithms have been devised to solve it efficiently.

The simplest procedure is implemented in the IC 
\cite{pearl1991theory} and SGS 
\cite{Spirtes2000} algorithms. For each pair of variables $(X, Y)$, these search through all subsets $\mathbf{W}$ of the remaining variables to check whether $X\independent Y \mid \mathbf{W}$.
If no such set $\mathbf{W}$ is found, then $X$ and $Y$ are connected with an edge.
Since this can be slow due to the large number of subsets,
the PC algorithm \cite{Spirtes2000} uses a much more efficient search procedure.
It starts from a complete graph and then sequentially test only subsets of the neighbors of $X$ or $Y$ of increasing size, removing an edge when a separating subset is found.
This neighbor search is no longer guaranteed to give the right result for causally insufficient systems, i.e., in the presence of hidden confounders.
The FCI (for Fast Causal Inference) algorithm \cite{Spirtes2000} addresses this setting, and produces a partially directed causal graph as output.

Apart from being limited to recovering a Markov equivalence class, constraint-based methods can suffer from statistical issues.
In practice, datasets are finite, and conditional independence testing is a notoriously difficult problem, especially if conditioning sets are continuous and multi-dimensional. So while in principle, the conditional independences implied by the causal Markov condition hold true irrespective of the complexity of the functions appearing in an SCM, for finite datasets, conditional independence testing is hard without additional assumptions \cite{shah2020hardness}.
Recent progress in (conditional) independence testing heavily relies on kernel function classes to represent probability distributions in reproducing kernel Hilbert spaces, see \cref{sec:slt}. 

\paragraph{Score-Based Methods}
\label{ch:2-score-based}
Score-based approaches to causal discovery assign a score to each graph $G$ from a set of candidate graphs (usually the set of all DAGs).
The score $S$ is supposed to reflect how well $G$ explains the observed data $\mathbf{D}=\{\mathbf{x}_1,\dots,\mathbf{x}_m\}$,
and we choose the graph $\hat{G}$ maximizing this score,
\begin{equation*}
    \hat{G}=\argmax_G S(G\mid \mathbf{D}).
\end{equation*}
Various score functions have been proposed, but most methods assume a parametric model which factorises according to $G$, parametrised by $\theta\in\Theta$.
Two common choices are multinomial models for discrete data \cite{cooper1992bayesian} and linear Gaussian models for continuous data~\cite{geiger1994learning}.
E.g., a penalized maximum likelihood approach using the BIC \cite{schwarz1978estimating} as a score yields
\begin{equation}
\label{eq:BIC-score}
    S_{\textsc{bic}}(G\mid \mathbf{D})=\log p(\mathbf{D}\mid G,\hat{\theta}^{\textsc{mle}}) - \frac{k}{2}\log m
\end{equation}
where $k$ is the number of parameters and $\hat{\theta}^{\textsc{mle}}$ is the maximum likelihood estimate for $\theta$ to $D$ in $G$.
Note that $k$ generally increases with the number of edges in $G$ so that the second term in~\eqref{eq:BIC-score} penalizes complex graphs which do not lead to substantial improvements.

Another choice 
of score function 
is the marginal likelihood, or evidence, in a Bayesian approach to causal discovery, which requires specifying prior distributions over graphs and parameters, $p(G,\theta)=p(G)p(\theta\mid G)$.
The score for $G$ is then given by
\begin{equation}
    \label{eq:Bayesian-score}
    S_{\textsc{bayes}}(G\mid \mathbf{D})=p(\mathbf{D}\mid G)=\int_{\Theta}p(\mathbf{D}\mid G,\theta)p(\theta\mid G)\mathrm{d}\,\theta.
\end{equation}
This integral is intractable in general, but can be computed exactly for some models such as a Dirichlet-multinomial under some mild additional assumptions
\cite{heckerman1995learning, heckerman2006bayesian}. 

A major drawback of score-based approaches is the combinatorial size of the search space.
The number of DAGs over $n$ random variables grows super-exponentially
and can be computed recursively (to account for acyclicity constraints)~\cite{robinson1973counting}.
E.g., the number of DAGs for $n=5$ and $n=10$ nodes is 29281 and 4175098976430598143, respectively.
Finding the best scoring DAG is 
NP-hard \cite{chickering1996learning}.
To overcome this problem, greedy search techniques can be applied, e.g.,
greedy equivalence search (GES)~\cite{chickering2002optimal} which optimizes for the BIC.
\commentout{Starting from an empty graph, GES first greedily adds, and then greedily removes individual edges until the objective can no longer be improved.
With further improvements and tweaks, GES-based methods have been successfully scaled up and applied to problems with up to a million variables~\cite{ramsey2017million}.}

In recent years, another class of methods has emerged that is based on assuming particular functional forms for the SCM assignments. Those arose from studying the cause-effect inference problem, as discussed below.

\paragraph{Cause-Effect Inference.}

\commentout{Let us return to the problem of causal discovery from observational data. Subject to suitable assumptions such as \emph{faithfulness} \cite{Spirtes2000}, one can sometimes recover aspects of the underlying graph from observations by performing conditional independence tests. However, there are several problems with this approach.
One is that in practice, our datasets are always finite, and conditional independence testing is a notoriously difficult problem, especially if conditioning sets are continuous and multi-dimensional. So while in principle, the conditional independences implied by the causal Markov condition hold true irrespective of the complexity of the functions appearing in an SCM, for finite datasets, conditional independence testing is hard without additional assumptions \cite{1804.07203}.
The other problem is that}
In the case of only two variables, the ternary concept of conditional independences collapses and the causal Markov condition~(\cref{def:causalMarkov}) thus has no nontrivial implications. 
However, we have seen in~\cref{sec:icm} that assuming an independence of mechanisms (\cref{princ:icm}) lets us find asymmetries between cause and effect, and thus address the cause-effect inference problem previously considered unsolvable~\cite{Janzingetal12}.
It turns out that this problem can be also addressed by making additional assumptions on function classes,
as not only the graph topology leaves a footprint in the observational distribution, but so do the functions $f_i$ in an SCM.
Such assumptions are typical for machine learning, where it is well-known that finite-sample generalization without assumptions on function classes is impossible, and where much attention is devoted to properties of function classes (e.g., priors or capacity measures), as discussed in~\cref{sec:slt}.

Let us provide an intuition as to why assumptions on the functions in an SCM should help learn about them from data. Consider a toy SCM with only two observables $X\to Y$. In this case, the structural equations~\eqref{eq:structural_eqs} turn into
\begin{align}
X := U, \qquad \qquad 
Y := f(X, V) \label{eq:SA2}
\end{align}
with noises $U\independent V$.
Now think of $V$ acting as a random selector variable choosing from among a set of functions ${\mathcal F} = \{ f_v (x) \equiv f(x,v) \; | \; v \in \mbox{supp}(V)\}$. If $f(x,v)$ depends on $v$ in a non-smooth way, it should be hard to glean information about the SCM from a finite dataset, given that $V$ is not observed and it randomly switches between arbitrarily different $f_v$.\footnote{Suppose $X$ and $Y$ are binary, and $U,V$ are uniform Bernoulli variables, the latter selecting from ${\mathcal F} = \{ id, not\}$ (i.e., identity and negation). In this case, the entailed distribution for $Y$ is uniform, {\em independent} of $X$, even though we have $X\to Y$. We would be unable to discern $X\to Y$ from data. (This would also constitute a violation of faithfulness~(\cref{ass:faithfulness})).
}
This motivates restricting the complexity with which $f$ depends on $V$. A natural restriction is to assume an \textit{ additive noise model}
\begin{align}
X := U,
\qquad \qquad 
Y := f(X) + V.
\end{align}
If $f$ in \eq{eq:SA2} depends smoothly on $V$, and if $V$ is relatively well concentrated, this can be motivated by a local Taylor expansion argument.
Such assumptions drastically reduce the effective size of the function class---without them, the latter could depend exponentially on the cardinality of the support of $V$. 

Restrictions of function classes can break the symmetry between cause and effect in the two-variable case: one can show that given a distribution over $X,Y$ generated by an additive noise model, one cannot fit an additive noise model in the opposite direction (i.e., with the roles of $X$ and $Y$ interchanged) \cite{Hoyer2008,Mooij2009,PetMooJanSch14,Kpotufe14,BauSchPet16}. This is subject to certain genericity assumptions, and notable exceptions include the case where $U,V$ are Gaussian and $f$ is linear. It generalizes results of \cite{Shimizu2006} for linear functions, and it can be generalized to include nonlinear rescaling \cite{Zhang2009}, cycles \cite{Mooij11}, confounders \cite{Janzing2009uai}, and multi-variable causal discovery \cite{Peters2011b}. There is now a range of methods that can detect causal direction better than chance~\cite{Mooijetal16}. 

\commentout{
Assumptions on function classes have thus helped address the cause-effect inference problem. They can also help address the other weakness of causal discovery methods based on conditional independence testing. Recent progress in (conditional) independence testing heavily relies on kernel function classes to represent probability distributions in reproducing kernel Hilbert spaces \cite{Gretton2005,Gretton2005JMLR,FukGreSunSch08,Zhang2011uai,PfiBuhSchPet18}.
}%

We have thus gathered some evidence that ideas from machine learning can help tackle causality problems that were previously considered hard. Equally intriguing, however, is the opposite direction: can causality help us improve machine learning?
\commentout{Present-day machine learning (and thus also much of modern AI) is based on statistical modeling, but as these methods become pervasive, their limitations are becoming apparent. }

\paragraph{Nonstationarity-Based Methods}
\label{ch:2-nonstationarity-based}
The last family of causal discovery approaches we mention is based on ideas of nonstationarity and invariance \cite{Schoelkopf2012}.
These approaches do not apply to purely observational data collected in an i.i.d.\ setting.
In contrast, they aim to leverage heterogeneity of data collected from different environments.
The main idea is the following: since causal systems are modular in the sense of the ICM~\cref{princ:icm}, changing one of the independent mechanisms should leave the other components, or causal conditionals, unaffected (SMS~\cref{pri:scsh}).
A correct factorization of the joint distribution according to the underlying causal structure should thus be able to explain heterogeneity by localized changes in one (or few) of the mechanisms while the others remain invariant.

One of the first works to use this idea~\cite{tian2001causal} analyzed which causal structures can be distinguished given data resulting from a set of mechanism changes.
Recent work~\cite{huang2019causal} additionally aims to learn a low-dimensional representation of the mechanism changes.
Other works \cite{peters2016causal,RojSchTurPet18} have proposed methods for finding the direct causes of a given target variable.
Using a recent result on identifiability of non-linear ICA~\cite{hyvarinen2018nonlinear} which also relies on non-stationarity, a method for learning general non-linear SCMs was proposed \cite{monti2019causal}.
Here the idea is to train a classifier to discriminate between the true value of some nonstationarity variable (such as a time-stamp or environment indicator) and a shuffled version thereof.

\section{Implications for Machine Learning\label{sec:implications}}

\paragraph{Semi-Supervised Learning}
Suppose our underlying causal graph is $X\to Y$, and
we wish to learn a mapping $X\to Y$. The causal factorization \eq{eq:cf} in this case is 
\begin{equation}
p(X,Y) = p(X) \; p(Y\mid X).
\end{equation}
The ICM~\cref{princ:icm} posits that the modules in a joint distribution's causal factorization do not inform or influence each other.
This means that, in particular, $p(X)$ should contain no information about $p(Y\mid X)$, which implies that semi-supervised learning \cite{ChaSchZie06} should be futile, as it is trying to use additional information about $p(X)$ (from unlabeled data) to improve our estimate of $p(Y\mid X=x)$.
How about the opposite direction? Is there hope that semi-supervised learning should be possible in that case? It turns out the answer is yes, due to work on cause-effect inference using the ICM~\cref{princ:icm}~\cite{Daniusisetal10}. It introduced a measure of dependence between the input and the conditional of output given input, and showed that if this dependence is zero in the causal direction, then it is strictly positive in the opposite direction. Independence of cause and mechanism in the causal direction thus implies that in the backward direction (i.e., for anticausal learning), the distribution of the input variable should contain information about the conditional of output given input, i.e., the quantity that machine learning is usually concerned with. This is exactly the kind of information that semi-supervised learning requires when trying to improve the estimate of output given input by using unlabeled inputs. 
This suggests that {\em semi-supervised learning should be impossible for causal learning problems, but feasible otherwise}, in particular for anticausal ones.
A meta-analysis of published semi-supervised learning benchmark studies corroborated this prediction \cite{Schoelkopf2012}, and similar results apply for natural language processing \cite{jinetal21}. 
These findings are intriguing since they provide insight into {\em physical} properties of learning problems, thus going beyond the methods and applications that machine learning studies usually provide. 

Subsequent developments include further theoretical analyses \cite{JanSch15,PetJanSch17} and a form of conditional semi-supervised learning \cite{KugMeyLooSch19}. 
The view of semi-supervised learning as exploiting dependences between a marginal $p(x)$ and a non-causal conditional $p(y\mid x)$ is consistent with the common assumptions employed to justify semi-supervised learning~\cite{ChaSchZie06,1911.10500}. 

\commentout{The \emph{cluster assumption} asserts that the labeling function (which is a property of $p(y\mid x)$) should not change within clusters of $p(x)$. The \emph{low-density separation assumption} posits that the area where $p(y\mid x)$ takes the value of $0.5$ should have small $p(x)$; and the \emph{semi-supervised smoothness assumption}, applicable also to continuous outputs, states that if two points in a high-density region are close, then so should be the corresponding output values. Note, moreover, that some of the theoretical results in the field use assumptions well-known from causal graphs (even if they do not mention causality): the {\em co-training theorem} \cite{BluMit98} makes a statement about learnability from unlabeled data, and relies on an assumption of predictors being conditionally independent given the label, which we would normally expect if the predictors are (only) caused by the label, i.e., an anticausal setting. This is nicely consistent with the above findings.}

\paragraph{Invariance and Robustness}
We have discussed the shortcomings of the i.i.d.\ assumption, which rarely holds true exactly in practice, and the fact that real-world intelligent agents need to be able to generalize not just within a single i.i.d.\ setting, but across related problems. This notion has been termed {\em out-of-distribution (o.o.d.) generalization}, attracting significant attention in recent years \cite{scholkopfetal21}. While most work so far has been empirical, statistical bounds would be desirable that generalize \eq{slt-eq:unibound2}, including additional quantities measuring the distance between training and test distribution, incorporating meaningful assumptions~\cite{shalit2017estimating}. Such assumptions are necessary \cite{ben2010impossibility}, and could be causal, or related to invariance properties.

The recent phenomenon of ``adversarial vulnerability'' \cite{1312.6199} shows that minuscule targeted violations of the i.i.d.\ assumption,
generated by adding suitably chosen noise to images (imperceptible to humans), can lead to dangerous errors such as confusion of traffic signs. These examples are compelling as they showcase non-robustnesses of artificial systems which are not shared by human perception. \looseness-1 Our own perception thus exhibits invariance or robustness properties that are not easily learned from a single training set.

Early causal work related to domain shift \cite{Schoelkopf2012} looked at the problem of learning from multiple cause-effect datasets that share a functional mechanism but differ in noise distributions. More generally, given (data from) multiple distributions, one can try to identify components which are robust, and find means to transfer them across problems~\cite{zhang_domain_2013,Bareinboim2014,zhang_multi-source_2015,GonZhaLiuTaoSch16,HuaZhaZhaSanGlySch17}.
\looseness-1   According to the ICM~\cref{princ:icm}, invariance of conditionals or functions (also referred to as covariate shift in simple settings) should only hold in the causal direction, a reversal of the impossibility described for SSL.

Building on the work of \cite{Schoelkopf2012,peters2016causal}, the idea of invariance for prediction has also been used for supervised learning  \cite{RojSchTurPet18,arjovsky2019invariant,2102.12353}.
In particular, ``invariant risk minimization'' (IRM) was proposed as an alternative to ERM, cf.~\eqref{eq:slt:r_emp}. 

{
\section{Causal Reasoning}
\label{sec:reasoning}
In contrast to causal discovery~(\cref{ch:2-discovery}), which aims to uncover the causal structure underlying a set of variables, \textit{causal reasoning} starts from a known (or postulated) causal graph and answers causal queries of interest.
While causal discovery often looks for {qualitative} relationships, causal reasoning usually aims to {quantify} them.
This requires two steps: (i) \textit{identifying} the query, i.e., deriving an estimand for it that only involves observed quantitites; and (ii) \textit{estimating} this using data.
Often, the quantities of interest can be described as treatment effects, i.e., contrasts between two interventions.

\begin{definition}[Treatment effects]
\label{def:treatment-effects}
The conditional average treatment effect (CATE), conditioned on (a subset of) features $\mathbf{x}$, is defined as
\begin{equation}
\label{eq:CATE-x}
    \tau(\mathbf{x}):=\mathbb{E}[Y\mid \mathbf{x}, do(T=1)]-\mathbb{E}[Y\mid \mathbf{x}, do(T=0)]=\mathbb{E}[Y(1)-Y(0)\mid \mathbf{x}].
\end{equation}
The average treatment effect (ATE) is defined as the population average of the CATE,
\begin{equation}
\label{eq:ATE-x}
    \tau:=\mathbb{E}[\tau(\Xb)]=\mathbb{E}[Y\mid do(T=1)]-\mathbb{E}[Y\mid do(T=0)]=\mathbb{E}[Y(1)-Y(0)].
\end{equation}
\end{definition}

While ITE~(\cref{def:ITE}) and CATE~\eqref{eq:CATE-x} are sometimes used interchangeably, there is a conceptual difference:
ITE refers to the difference of two POs and is thus bound to an individual, while CATE applies to subpopulations, e.g., the CATE for females in their 40s.
Since the ITE is fundamentally impossible to observe, it is often estimated by the CATE conditional on an individual's features $\mathbf{x}_i$ using suitable additional assumptions.

As is clear from~\cref{def:treatment-effects}, the treatment effects we want to estimate involve interventional expressions.
However, we usually only have access to observational data.
Causal reasoning can thus be cast as answering interventional queries using observational data and a causal model.
This involves dealing with confounders, both observed and unobserved.

Before discussing how to identify and estimate causal effects, we illustrate why causal assumptions are necessary using a well-known statistical phenomenon.

\paragraph{Simpson's Paradox and Covid-19}
\begin{figure}[t]
    \centering
    \begin{minipage}[c]{0.5\textwidth}
        \centering
        \includegraphics[width=\textwidth]{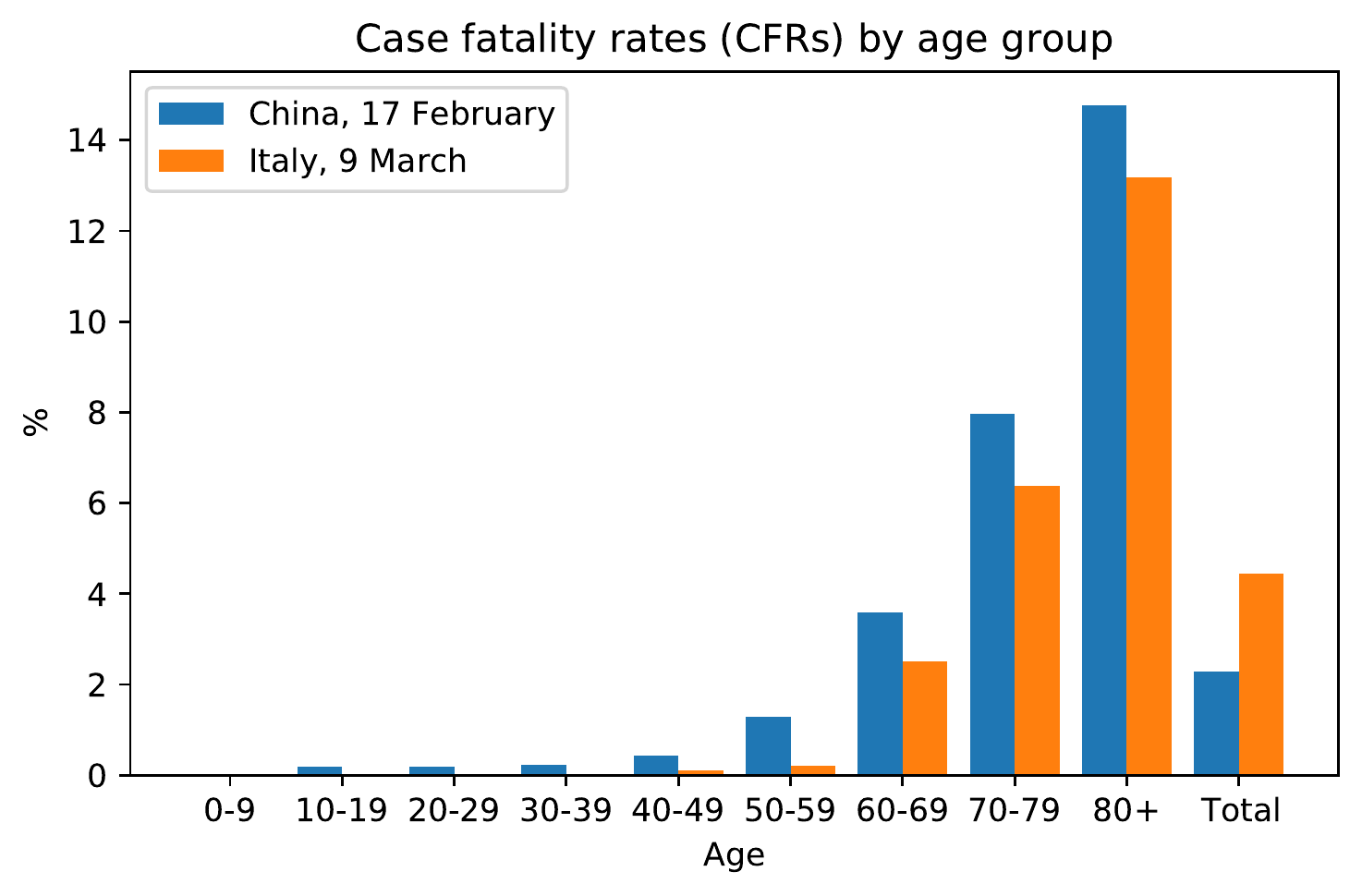}
    \end{minipage}%
    \begin{minipage}[c]{0.5\textwidth}
        \centering
        \includegraphics[width=\textwidth]{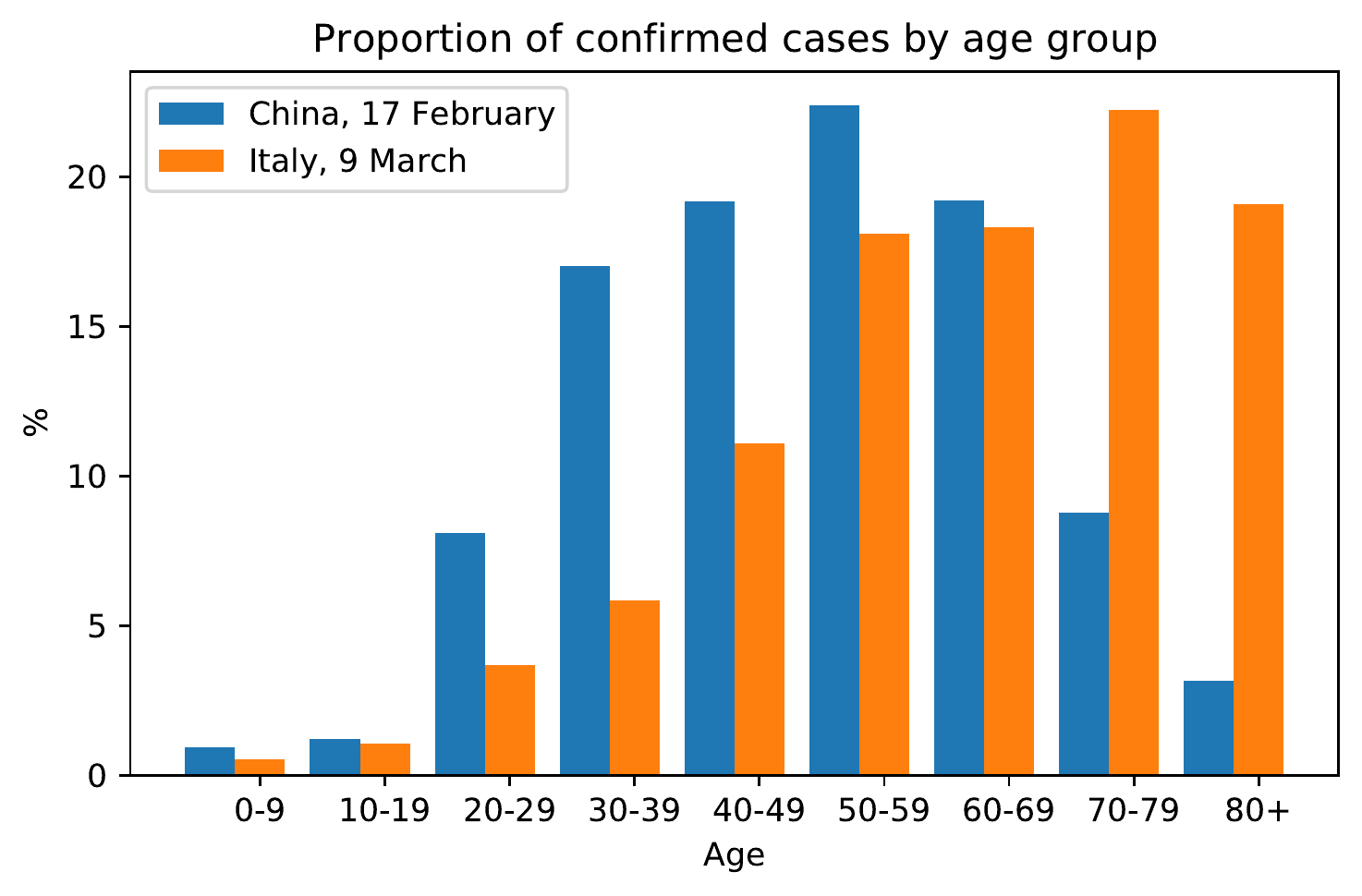}
    \end{minipage}
    \caption{Left: Covid-19 case fatality rates (CFRs) in Italy and China by age and in aggregate (``Total''),  including all confirmed cases and fatalities up to the time of reporting in early 2020 (see legend): 
    for all age groups, CFRs in Italy are lower than in China, but the total CFR in Italy is higher, an example of \textit{Simpson's paradox}.
    \looseness-1 Right: The case demographic differs between countries: in Italy, most cases occurred in the older population (figure from~\cite{von2020simpson}).%
    }
    \label{fig:simpson}
\end{figure}
\emph{Simpson's paradox} refers to the observation that aggregating data across subpopulations may yield opposite trends (and thus lead to reversed conclusions) from considering subpopulations separately~\cite{simpson1951interpretation}.
We observed a textbook example of this during the Covid-19 pandemic by comparing case fatality rates (CFRs), i.e., the proportion of confirmed Covid-19 cases which end in fatality, across different countries and age groups as illustrated in~\cref{fig:simpson}~\cite{von2020simpson}:
for \textit{all} age groups, CFRs in Italy are \textit{lower} than in China, but the \textit{total} CFR in Italy is \textit{higher}.

\text{How can such a pattern be explained?}
The case demographic (see~\cref{fig:simpson}, right) is rather different across the two countries, i.e., there is a statistical association between country and age.
In particular, Italy recorded a much larger proportion of cases in older patients who are generally at higher risk of dying from Covid-19~(see~\cref{fig:simpson}, left).
While this provides a consistent explanation in a \textit{statistical} sense, the phenomenon may still seem  puzzling as it defies our \textit{causal} intuition.
Humans appear to naturally extrapolate conditional probabilities to read them as causal effects, which can lead to inconsistent conclusions and may leave one wondering: \textit{how can the disease in Italy be less fatal for the young, less fatal for the old, but more fatal for the people overall?}
It is for this reason that the reversal of (conditional) probabilities in \cref{fig:simpson} is perceived as and referred to as a ``paradox''~\cite{pearl2014comment,hernan2011simpson}.

If we consider the country as treatment whose causal effect on fatality is of interest, then causal assumptions (e.g., in the form of a causal graph) are needed to decide how to handle covariates such as age that are statistically associated with the treatment, e.g., whether to stratify by (i.e., adjust for) age or not. 
This 
also explains why randomized controlled trials (RCTs)~\cite{fisher1937design} are the gold standard for causal reasoning:
randomizing the assignment breaks any potential links between the treatment variable and other covariates, thus eliminating potential problems of bias.
However, RCTs are costly and sometimes unethical to perform, so that causal reasoning often relies on observational data only.\footnote{For a treatment of more general types of data fusion and transportability of experimental findings across different populations we refer to \cite{pearl2014external,bareinboim2016causal}.}

We first consider the simplest setting \textit{without hidden confounders and with overlap}. We start with \textit{identification} of treatment effects on the population level, and then discuss different techniques for \textit{estimating} these from data.

\paragraph{Identification}
In absence of unmeasured variables (i.e., without hidden confounding), and provided we know the causal graph, it is straight-forward to compute causal effects by adjusting for covariates.
A principled approach to do so for any given graph was 
proposed by Robins \cite{robins1986new} and is known as the \textit{g-computation formula} (where the \textit{g} stands for general).
It is also known as \textit{truncated factorisation} \cite{Pearl2009} or \textit{manipulation theorem} \cite{Spirtes2000}.
It relies on the 
independence of causal mechanisms~(\cref{princ:icm}), i.e., the fact that intervening on a variable leaves the other causal conditionals in~\eqref{eq:cf} unaffected:
\begin{equation}
\label{eq:g-computation}
   p(X_1,\dots,X_n\mid do(X_i=x_i))=
\delta(X_i=x_i)
   \prod_{j\neq i}p(X_j\mid \PA_j)
\end{equation}
From~\eqref{eq:g-computation} the interventional distribution of interest can then be obtained by marginalization.
This is related to the idea of graph surgery (see~\cref{fig:DAGs}), and leads to a set of three inference rules for manipulating interventional distributions known as \textit{do-calculus}~\cite{Pearl2009} that have been shown to be complete for identifying causal effects~\cite{huang2006pearl,shpitser2006identification}.

Note that covariate adjustment
may be needed even if there are no clear confounders directly influencing both treatment and outcome, as shown by the example in~\cref{fig:g-formula}. 

\begin{figure}
\begin{minipage}[c]{0.625\textwidth}
\caption[Graphical criteria for covariate adjustment]{Treatment effect estimation with three observed covariates $X_1,X_2,X_3$: 
here, the valid adjustment sets for $T\to Y$ (see~\cref{prop:adjustment}) are $\{X_1\}$, $\{X_2\}$, and $\{X_1,X_2\}$. Including $X_3$ opens the \textit{non-directed path} $T\rightarrow X_3 \leftarrow X_2 \rightarrow Y$
and lies on the directed path $T\to X_3\to Y$, both of which can introduce bias.}
    \label{fig:g-formula}
\end{minipage}
\quad 
\begin{minipage}[c]{0.35\textwidth}
    \centering
    \begin{tikzpicture}
    \centering
    \node (X1) [obs, scale=\nodescale] {$X_1$};
    \node (X2) [obs, xshift=\xshift, scale=\nodescale] {$X_2$};
    \node (T) [obs, yshift=-1.25*\yshift, scale=\nodescale] {$T$};
    \node (Y) [obs, yshift=-1.25*\yshift, xshift=2*\xshift, scale=\nodescale] {$Y$};
    \node (X3) [obs, xshift=\xshift, yshift=-2.5*\yshift, scale=\nodescale] {$X_3$};
    \edge {X1} {T,X2};
    \edge{T}{Y, X3};
    \edge{X2}{Y,X3}
    \edge{X3}{Y}
    \end{tikzpicture}
    \end{minipage}
\end{figure}
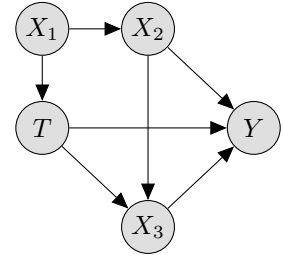

\begin{example}[]
Applying the g-computation formula~\eqref{eq:g-computation} to the setting of~\cref{fig:g-formula}, we obtain
\begin{align}
p(y\mid do(t))&=\sum_{x_1} p(x_1) \sum_{x_2} p(x_2\mid x_1) \sum_{x_3} p(x_3\mid t,x_2) p(y\mid t,x_2,x_3)\\
&=\sum_{x_1}p(x_1)\sum_{x_2}p(x_2\mid x_1)p(y\mid t,x_2)=\sum_{x_2}p(x_2)p(y\mid t,x_2) \label{eq:obvious-adjustment}\\
&\overset{(a)}{=}\sum_{x_1,x_2}p(x_1,x_2)p(y\mid t,x_1,x_2)\overset{(b)}{=}\sum_{x_1}p(x_1)p(y\mid t,x_1)\label{eq:less-obvious-adjustment}
\end{align}
where the last line follows by using the following conditional independences implied by the graph: (a) $Y\independent X_1\mid \{T,X_2\}$, and (b) $X_2\independent T\mid X_1$.
\end{example}

Note that both the RHS in~\eqref{eq:obvious-adjustment} and both sides in~\eqref{eq:less-obvious-adjustment} take the form
\begin{equation}
\label{eq:adjustment-set}
    p(y\mid do(t))=\sum_{\mathbf{z}}p(\mathbf{z})p(y|t,\mathbf{z}).
\end{equation}
In this case we call $\mathbf{Z}$ a \textit{valid adjustment set} for the effect of $T$ on $Y$.
Here, $\{X_1\}$, $\{X_2\}$, and $\{X_1,X_2\}$ are all valid adjustment sets, but it can be shown that, e.g., $\{X_1,X_3\}$ is not~(see~\cref{fig:g-formula}). 
\looseness-1 As computing the g-formula with many covariates can be cumbersome, graphical criteria for which subsets
constitute valid adjustment sets are useful in practice, even in the absence of unobserved confounders.

\begin{proposition}[\cite{shpitser2010validity}]
\label{prop:adjustment}
Under causal sufficiency, a set $\mathbf{Z}$ is a valid adjustment set for the causal effect of a singleton treatment $T$ on an outcome $Y$ (in the sense of~\eqref{eq:adjustment-set}) if and only if the following two conditions hold: (i) $\mathbf{Z}$ contains no descendant of any node on a directed path from $T$ to $Y$ (except for descendants of $T$ which are not on a directed path from $T$ to $Y$); and (ii) $\mathbf{Z}$ blocks
all non-directed paths from $T$ to~$Y$. 
\end{proposition}

Here, a path is called \textit{directed} if all directed edges on it point in the same direction, and \textit{non-directed} otherwise.
A path is \textit{blocked} (by a set of vertices $\mathbf{Z}$) if it contains a triple
of consecutive nodes connected in one of the following three ways: $A\to B\to C$ with $B\in \mathbf{Z}$, $A\leftarrow B\to C$ with $B\in \mathbf{Z}$,
or $A\to B\leftarrow C$, where neither
$B$ nor any descendant of $B$ is in $\mathbf{Z}$.

Two well-known types of adjustment set implied by~\cref{prop:adjustment} are \textit{parent adjustment}, where $\mathbf{Z}=\mathbf{Pa}_T$; and the \textit{backdoor criterion}, where $\mathbf{Z}$ is constrained to contain no descendants of $T$ and to block all ``back-door paths'' from $T$ to $Y$ ($T\leftarrow ...\, Y$).

Note that~\cref{prop:adjustment} only holds singleton treatments (i.e., interventions on a single variable). For treatments $\mathbf{T}$ involving multiple variables, a slightly more complicated version of~\cref{prop:adjustment} can be given in terms of \textit{proper} causal paths, and we refer to~\cite{Pearl1995,pearl2014confounding} for details.

Let us briefly return to our earlier example of Simpson's paradox and Covid-19. Considering a plausible causal graph for this setting~\cite{von2020simpson}, we find that age $A$ acts as a \textit{mediator} $C\to A \to F$ of the causal effect of country $C$ on fatality~$F$ (there is likely also a direct effect $C\to F$, potentially mediated by other, unobserved variables). If we are interested in the (total) causal effect of $C$ on $F$ (i.e., the overall influence of country on fatality), $A$ should not be included for adjustment according to~\cref{prop:adjustment}, and, subject to causal sufficiency, the total CFRs can be interpreted causally.\footnote{Mediation analysis~\cite{pearl2001direct} provides tools to tease apart and quantify the direct and indirect effects; the age-specific CFRs in~\cref{fig:simpson} then correspond to \textit{controlled direct effects}~\cite{von2020simpson}.}
For another classic example of Simpson's paradox in the context of kidney stone treatment~\cite{charig1986comparison}, on the other hand, the size of the stone acts as a \textit{confounder} and thus needs to be adjusted for to obtain sound causal conclusions. 

Valid covariate adjustment and the g-formula tell us how to compute interventions from the observational distribution when there are no hidden confounders.
To actually identify causal effects from data, however, we need to also be able to \textit{estimate} the involved quantities in~\eqref{eq:adjustment-set}.
This is a problem if a subgroup of the population never (or always) receives a certain treatment.
We thus need the additional assumption of a non-zero probability of receiving each possible treatment, referred to as \textit{overlap}, or common support.

\begin{assumption}[Overlap/common treatment support]
\label{ass:overlap}
For any treatment $t$ and any configuration of features $\mathbf{x}$, it holds that: $0<p(T=t \mid \mathbf{X}=\mathbf{x})<1$.
\end{assumption}

The combination of overlap and ignorability (i.e., no hidden confounders---see~\cref{ass:ignorability}) is also referred to as \textit{strong ignorability} and is a sufficient condition for identifying ATE and CATE: the absence of hidden confounders guarantees the existence of a valid adjustment set $\mathbf{Z}\subseteq \mathbf{X}$ for which $p(Y\mid do(T=t),\mathbf{Z})=p(Y\mid T=t,\mathbf{Z})$, and overlap guarantees that we can actually estimate the latter term for any $\mathbf{z}$ occurring with non-zero probability.\footnote{The overlap assumption can thus be relaxed to hold for at least one valid adjustment set.} 

\paragraph{Regression Adjustment}
Having identified a valid adjustment set  (using~\cref{prop:adjustment}),  \textit{regression adjustment} works by fitting a regression function $\hat{f}$ to $\mathbb{E}[Y\mid\mathbf{Z}=\mathbf{z}, T=t]=f(\mathbf{z},t)$ using an observational sample $\{(y_i,t_i,\mathbf{z}_i)\}_{i=1}^m$.
We can then use $\hat{f}$ to impute counterfactual outcomes as $\hat{y}_i^{\textsc{cf}}=\hat{f}(\mathbf{z}_i,1-t_i)$ in order to estimate the CATE.
The ATE is then given by the population average and can be estimated as
\begin{equation}
\label{eq:ATE-regression-adjustment}
\hat{\tau}_{\text{regression-adj.}}=\frac{1}{m_1}\sum_{i\,:\,t_i=1} \big(y_i-\hat{f}(\mathbf{z}_i,0)\big)
+
\frac{1}{m_0}\sum_{i\,:\,t_i=0} \big(\hat{f}(\mathbf{z}_i,1)-y_i\big),
\end{equation}
where $m_1$ and $m_0$ are the number of observations from the treatment and control groups, respectively. 
Note the difference to the RCT estimator where no adjustment is necessary,
\begin{equation}
\label{eq:ATE-RCT}
    \hat{\tau}_{\text{RCT}}=\frac{1}{m_1}\sum_{i\,:\,t_i=1} y_i-\frac{1}{m_0}\sum_{i\,:\,t_i=0}y_i.
\end{equation}

\paragraph{Matching and Weighting Approaches}
While regression adjustment indirectly estimates ATE via CATE, matching and weighting approaches aim to estimate ATE directly.
The general is idea to emulate the conditions of an RCT as well as possible.

\textit{Matching} approaches work by splitting the population into subgroups based on feature similarity.
This can be done on an individual level (so-called one-to-one or nearest neighbor matching) by matching each individual $i$ with the most similar one, $j(i)$, from the opposite treatment group (i.e., $t_i\neq t_{j(i)}$). The difference of their outcomes, $y_i-y_{j(i)}$, is then considered as a sample of the ATE, and their average taken as an estimate thereof,
\begin{equation}
    \label{eq:ATE-NN-matching}
    \hat{\tau}_{\text{NN-matching}}=\frac{1}{m_1}\sum_{i\,:\,t_i=1} (y_i-y_{j(i)})
    +
    \frac{1}{m_0}\sum_{i\,:\,t_i=0} (y_{j(i)}-y_i).
\end{equation}
Alternatively, the population can be split into larger subgroups with similar features (so-called strata). 
Each stratum is then treated as an independent RCT.
If there are $K$ strata containing $m_1, ..., m_K$ observations each, the stratified ATE estimator is
\begin{equation}
\label{eq:ATE-stratification}
    \hat{\tau}_{\text{stratified}} = \frac{\sum_{k=1}^K m_k \hat{\tau}_{\text{RCT}}^{(k)}}{\sum_{k=1}^K m_k}
\end{equation}
where $\hat{\tau}_{\text{RCT}}^{(k)}$ is the estimator from~\eqref{eq:ATE-RCT} applied to observation in the $k^{th}$ stratum.

\textit{Weighting} approaches, on the other hand, aim to counteract the confounding bias by reweighting each observation to make the population more representative of an RCT.
This means that underrepresented treatment groups are upweighted and overrepresented ones downweighted.
An example is the inverse probability weighting (IPW) estimator,
\begin{equation}
    \label{eq:ATE-IPW}
    \hat{\tau}_{\text{IPW}}=\frac{1}{m_1}\sum_{i\,:\,t_i=1} \frac{y_i}{p(T=1\mid\mathbf{Z}=\mathbf{z}_i)}-\frac{1}{m_0}\sum_{i\,:\,t_i=0}\frac{y_i}{p(T=0\mid\mathbf{Z}=\mathbf{z}_i)}.
\end{equation}
The treatment probability $p(T=1\mid\mathbf{Z})$ is also known as \textit{propensity score}.  While from a theoretical point of view $\mathbf{Z}$ should be a valid adjustment set, practitioners sometimes use all covariates to construct a propensity score.

\paragraph{Propensity Score-Methods}
To overcome the curse of dimensionality and gain statistical efficiency in high-dimensional, low-data regimes, propensity scores can be a useful tool, because covariates and treatment are rendered conditionally independent, $T\independent \mathbf{Z} \mid s(\mathbf{z})$, by the propensity score $s(\mathbf{z}):=p(T=1\mid\mathbf{Z}=\mathbf{z})$ \cite{rosenbaum1983central}.
Instead of adjusting for large feature sets or performing matching in high-dimensional spaces, the scalar propensity score can be used instead.
Applying this idea to the above methods gives rise to \textit{propensity score adjustment} and \textit{propensity score matching}.
For the latter, the difference in propensity scores is used as similarity between instances to find nearest neighbors or to define strata.

While simplifying in one respect, the propensity score needs to be estimated from data which is an additional source of error.
The standard approach for this is to estimate $s(\mathbf{z})$ by logistic regression, but more sophisticated methods are also possible. 
However, propensity score methods still rely on having identified a valid adjustment set~$\mathbf{Z}$ to give unbiased results.
Using all covariates to estimate $s$, without checking for validity as an adjustment set, can thus lead to wrong results.
Next, we consider the case of causal reasoning with \textit{unobserved confounders}.
While it is not possible to identify causal effects in the general case, we will discuss two particular situations in which ATE can still be estimated.
These are shown in~\cref{fig:hidden-confounders}a and~b.

\paragraph{Front-Door Adjustment}
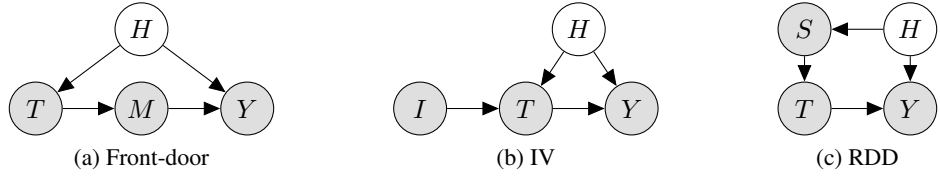
\begin{figure}[]
    \centering
    \subfloat[Front-door]{
        \centering
        \begin{tikzpicture}
        \centering
        \node (T) [obs, scale=\nodescale] {$T$};
        \node (M) [obs, xshift=\xshift, scale=\nodescale] {$M$};
        \node (Y) [obs, xshift=2*\xshift, scale=\nodescale] {$Y$};
        \node (H) [latent, xshift=\xshift, yshift=\yshift, scale=\nodescale] {$H$};
        \edge{H}{T,Y};
        \edge{T}{M};
        \edge{M}{Y};
        \end{tikzpicture}
    }
    \qquad     \qquad
    \subfloat[IV]{
        \centering
        \begin{tikzpicture}
        \centering
        \node (I) [obs, scale=\nodescale] {$I$};
        \node (T) [obs, xshift=\xshift, scale=\nodescale] {$T$};
        \node (Y) [obs, xshift=2*\xshift, scale=\nodescale] {$Y$};
        \node (H) [latent, xshift=1.5*\xshift, yshift=\yshift, scale=\nodescale] {$H$};
        \edge{T}{Y};
        \edge{I}{T};
        \edge{H}{T,Y};
        \end{tikzpicture}
    }
    \qquad    \qquad
    \subfloat[RDD]{
        \centering
        \begin{tikzpicture}
        \centering
        \node (T) [obs, scale=\nodescale] {$T$};
        \node (S) [obs, yshift=\yshift, scale=\nodescale] {$S$};
        \node (Y) [obs, xshift=\xshift, scale=\nodescale] {$Y$};
        \node (H) [latent, xshift=\xshift, yshift=\yshift, scale=\nodescale] {$H$};
        \edge{T}{Y};
        \edge{H}{S,Y};
        \edge{S}{T};
        \end{tikzpicture}
    }
    \caption[Methods for dealing with unobserved confounders]{ Overview of special settings which allow to estimate causal effects of treatment $T$ on outcome $Y$ when the strong ignorability assumption (no hidden confounding, and overlap) does not hold. In (a) the hidden confounder $H$ is dealt with by means of an observed mediator $M$, while (b) relies on an instrumental variable (IV) which is independent of $H$. (c) In a regression discontinuity design (RDD), treatment assignment is a threshold function of some observed decision score $S$ so that there is no overlap between treatment groups.}
    \label{fig:hidden-confounders}
\end{figure}
The first situation in which identification is possible even though a hidden variable $H$ confounds the effect between treatment and outcome is known as \textit{front-door adjustment}.
The corresponding causal graph is shown in~\cref{fig:hidden-confounders}a.
Front-door adjustment relies on the existence of an observed variable $M$ which blocks all directed paths from $T$ to $Y$, so that $T$ only causally influences $Y$ through $M$. 
For this reason $M$ is also called a \textit{mediator}. 
The other important assumption is that the hidden confounder does not 
influence the mediator other than through the treatment~$T$, i.e., $M\independent H \mid T$.
In this case, and provided $p(t,m)>0$ for all $t$ and $m$, the causal effect of $T$ on $Y$ is identifiable and is given by the following.
\begin{proposition}[Front-door adjustment]
For the causal graph in \cref{fig:hidden-confounders}a it holds that:
\begin{equation}
   \label{prop:front-door-adjustment}
    p(y\mid do(t))=\sum_{m}p(m\mid t)\sum_{t'}p(t')p(y\mid m,t').
\end{equation}
\end{proposition}

We give a sketch of the derivation, and refer to \cite{Pearl2009} for a proof using the rules of do-calculus.
Since $M$ mediates the causal effect of $T$ on $Y$, we have that
\begin{equation}
\label{eq:front-door-step1}
        p(y\mid do(t))=\sum_{m}p(m\mid do(t))p(y\mid do(m)).
\end{equation}
Since there are no backdoor paths from $T$ to $M$ we have
        $p(m\mid do(t))=p(m\mid t)$.

Moreover, $\{T\}$ is a valid adjustment set for the effect of $M$ on $Y$ by~\cref{prop:adjustment}, so
\begin{equation}
\label{eq:front-door-step3}
    p(y\mid do(m))=\sum_{t'}p(t')p(y\mid m,t').
\end{equation}
Substituting into \eqref{eq:front-door-step1} then yields expression~\eqref{prop:front-door-adjustment}.

We point out that the setting presented here is only the simplest form of front-door adjustment which is sufficient to convey the main idea.
It can be amended to include observed covariates $\mathbf{X}$ as well, as long as the conditions on the mediator remain satisfied.

\paragraph{Instrumental Variables (IV)}
The second setting for causal reasoning with hidden confounders is based on the idea of instrumental variables \cite{angrist1996identification, didelez2010assumptions, wright1928tariff}, see~\cref{fig:hidden-confounders}b. 
The IV approach relies on the existence of a special observed variable $I$ called instrument.

\begin{definition}[IV]
A variable $I$ is a valid instrument for estimating the effect of treatment $T$ on outcome $Y$ confounded by a hidden variable $H$ if all of the following three conditions hold: (i) $I\independent H$; (ii) $I\not\independent T$; and (iii) $I\independent Y \mid T$.
\end{definition} 

Condition (i) states that the instrument is independent of any hidden confounders $H$. 
Since this assumption cannot be tested, background knowledge is necessary to justify the use of a variable as IV in practice.
Conditions (ii) and (iii) state that the instrument is correlated with treatment, and only affects the outcome through $T$, and are referred to as relevance and exclusion restriction, respectively.

Given a valid IV, we apply a two-stage procedure: first obtain an estimate $\Hat{T}$ of the treatment variable~$T$ that is independent of $H$ by predicting $T$ from $I$.
Having thus created an unconfounded version of the treatment, a regression of $Y$ on $\Hat{T}$ then reveals the correct causal effect.
We demonstrate this idea for a simple linear model with continuous treatment variable where the causal effect can be obtained by two-stage least squares (2SLS). 

\begin{example}[Linear IV with 2SLS]
\label{example:IV}
Consider the linear SCM defined by
\begin{align*}
    T:=aI+bH+U_T, %
    \qquad \qquad 
    Y:=cH+dT+U_Y. %
\end{align*}
with $U_T, U_Y$ independent noise terms.
Then, since $I\independent H$, linear regression of $T$ on $I$ recovers the coefficient $a$ via $\Hat{T}=aI$.
Substituting for $T$ in the structural equation for $Y$ gives
\begin{equation*}
\label{eq:IV-combined}
    Y:=daI+(c+bd)H+U_Y+dU_T.
\end{equation*}
A second linear regression of $Y$ on $\Hat{T}=aI$ recovers the causal effect $d$ because $(I\independent H)\implies (\Hat{T}\independent H)$, whereas a naive regression of $Y$ on $T$ would give a different result, as $T\not\independent H$.
\end{example}

IVs have been studied extensively and more sophisticated versions than the simple example above exist, allowing for non-linear interactions and observed covariates.
Having discussed some special settings to deal with hidden confounding, we briefly present a technique to deal with violations of the overlap assumption.

\paragraph{Regression Discontinuity Design}
In a \textit{regression discontinuity design} (RDD) 
the treatment assignment mechanism behaves like a threshold function, i.e., the propensity score is discontinuous~\cite{imbens2008regression}. 
In the simplest setting, the assignment of treatment or control is determined by whether an \textit{observed score} $S$ is above a threshold~$s_0$, $T:=\mathbb{I}\{S\geq s_0\}$.
This score in turn depends on other covariates which may or may not be observed.
For example, patients may be assigned a risk score, and treatment is only prescribed if this score surpasses a given threshold.
Since the score may be assigned by another institution, not all relevant covariates  $H$ are usually observed.
However, it is assumed that the treatment decision only depends on the score, e.g., because doctors comply with the official rules.
The causal graph for such a simple RDD setting is shown in~\cref{fig:hidden-confounders}c.
While the score $S$ constitutes a valid adjustment set in principle, the problem with RDDs is the lack of overlap:
patients with low scores are always assigned $T=0$ and patients with high scores are always assigned $T=1$.
Because of this, covariate adjustment, matching, or weighting approaches do not apply.
The general idea of an RDD is to overcome this challenge by comparing observations with score in a small neighborhood of the decision cut-off value $s_0$, motivated by the consideration that patients with close scores but on opposite sides of $s_0$ differ only in whether they received the treatment or not.
For example, if the treatment cut-off value is 0.5 for a score in [0,1], then patients with scores of 0.49 and 0.51 are comparable and can be treated as samples from an RCT.
An RDD (in its simplest form) thus focuses on differences in the regression function $\mathbb{E}[Y\mid S=s, T=t(s)]=f(s)$ for $s\in[s_0-\epsilon, s_0+\epsilon]$, where $\epsilon>0$ is small.
\commentout{An important role is played by the type of regression used (linear, polynomial,...) and the estimated effect may depend on this choice.
RDDs also exist in fuzzy (rather than sharp) form where the treatment assignment is not fully deterministic.}

\paragraph{Half-Sibling Regression  and Exoplanet Detection}
We conclude this section with a real-world application 
performing causal reasoning in a confounded additive noise model. 
Launched in 2009, NASA's Kepler space telescope initially observed 150000 stars over four years, in search of exoplanet transits. These are events where a planet partially occludes its host star, causing a slight decrease in brightness, often orders of magnitude smaller than the influence of telescope errors. When looking at stellar light curves, we noticed that the noise structure was often shared across stars that were light years apart. Since that made direct interaction of the stars impossible, it was clear that the shared information was due to the telescope acting as a confounder.
We thus devised a method that (a) regresses a given star of interest on a large set of other stars chosen such that their measurements contain no information about the star's astrophysical signal, and (b) removes that regression in order to cancel the telescope's influence.\footnote{For events that are localized in time (such as exoplanet transits), we further argued that the same applies for suitably chosen past and future values of the star itself, which can thus also be used as predictors.} 
The method is called ``half-sibling'' regression since target and predictors share a parent, namely the telescope.
The method recovers the random variable representing the astrophysical signal almost surely (up to a constant offset), for an additive noise model (specifically, the observed light curve is a sum of the unknown astrophysical signal and an unknown function of the telescope noise), subject to the assumption that the telescope's effect on the star is in principle predictable from the other stars \cite{Scholkopfetal16}.

In 2013, the Kepler spacecraft suffered a technical failure, which left it with only two functioning reaction wheels, insufficient for the precise spatial orientation required by the original Kepler mission. NASA decided to use the remaining fuel to make further observations, however the systematic error was significantly larger than before---a godsend for our method designed to remove exactly these errors. 
We augmented it with models of exoplanet transits and an efficient way to search light curves, leading to the discovery of 36 planet candidates \cite{Foreman-Mackeyetal15}, of which 21 were
subsequently validated as bona fide exoplanets~\cite{Montet_2015}.
Four years later, astronomers found traces of water in the atmosphere of the exoplanet K2-18b---the first such discovery for an exoplanet in the habitable zone, i.e., allowing for liquid water \cite{1909.04642,Tsiaras}. The planet turned out to be one that had been first detected in our work \cite{Foreman-Mackeyetal15} (exoplanet candidate EPIC~201912552).}

\section{Current Research and Open Problems\label{sec:open}}

\paragraph{Conservation of Information}
{We have previously argued that the mechanization of information processing plays currently plays a similar role to the mechanization of energy processing in earlier industrial revolutions \cite{1911.10500}. Our present understanding of information is rather incomplete, as was the understanding of energy during the course of the first two industrial revolutions. The profound modern understanding of energy came with Emmy Noether and the insight that energy conservation is due to a symmetry (or covariance) of the fundamental laws of physics: they look the same no matter how we shift time. 
One might argue that information, suitably conceptualized, should also be a conserved quantity, 
and that this might also be a consequence of symmetries. 
The notions of invariance/independence discussed above may be able to play a role in this respect.

Mass seemingly played two fundamentally different roles (inertia and gravitation) until Einstein furnished a deeper connection in general relativity. It is noteworthy that causality introduces a layer of complexity underlying the symmetric notion of statistical mutual information. Discussing source coding and channel coding, Shannon \cite{Shannon59} remarked: {\em This duality can be pursued further and is related to a duality between past and future and the notions of control and knowledge. Thus we may have knowledge of the past but cannot control it; we may control the future but have no knowledge of it.
  }
}

\paragraph{What is an Object?}
Following the i.i.d.\ pattern recognition paradigm, machine learning learns objects by extracting patterns from many observations. An complementary view may consider objects 
as modules that can be separately manipulated or intervened upon \cite{tangemannetal21}. The idea that objects are defined by their behavior under transformation has been influential in fields ranging from psychology to mathematics \cite{klein1872vergleichende,maclane:71}.

\paragraph{Causal Representation Learning.}
\begin{figure}
\small
\noindent\begin{minipage}{0.45\textwidth}
\begin{tcolorbox}
{\bfseries Classic AI}: 

symbols provided a priori;

rules provided a priori.

\end{tcolorbox}
 \end{minipage}
\begin{minipage}{0.55\textwidth}
\begin{tcolorbox}
{\bfseries Machine Learning}: 
\\
representations (symbols) learned from data;
\\
only include {\em statistical} information.
\end{tcolorbox}
\end{minipage}

\vglue0.7mm

\begin{minipage}{0.45\textwidth}
\begin{tcolorbox}
{\bfseries Causal Modeling}: 
\\
structural causal models assume the causal variables (symbols) are given.
\end{tcolorbox}
\end{minipage}
\begin{minipage}{0.55\textwidth}
\begin{tcolorbox}[colback=red!5!white,colframe=red!75!black]
{\bfseries Causal Representation Learning}: 
\\
capture interventions, reasoning, planning---
{\em ``Thinking is acting in an imagined space''}
(Konrad Lorenz)
\end{tcolorbox}
 \end{minipage}
 \caption{\label{fig:causalrep}Causal representation learning aims to automatically learn representations that contain not just statistical information, but support interventions, reasoning, and planning. The long-term goal of this field is to learn causal world models supporting AI, or causal digital twins of complex systems.
 }
 \end{figure}
In hindsight, it appears somewhat naive that first attempts to build AI tried to realize intelligence by programs written by humans, since existing examples of intelligent systems appear much too complex for that. However, there is a second problem, which is just as significant: classic AI assumed that the symbols which were the basis of algorithms were provided a priori by humans. \looseness-1  When building a chess program, it is clear that the algorithms operate on chess board positions and chess pieces; however, if we want to solve a real-world problem in an unstructured environment (e.g., recognize spoken language), it is not clear what constitutes the basic symbols to be processed.

Traditional causal discovery and reasoning assumed that the elementary units are random variables connected by a causal graph. Real-world observations, however, are usually not structured into such units to begin with. For instance, objects in images that permit causal reasoning first need to be discovered \cite{LopNisChiSchBot17,von2020towards,tangemannetal21,locatello2020object}.
The emerging field of \textit{causal representation learning} strives to learn these variables from data, much like machine learning went beyond symbolic AI in not requiring that the symbols that algorithms manipulate be given a priori (see \cref{fig:causalrep}).

\begin{figure}[t]
    \centering
    \subfloat[]{
        \centering
        \begin{tikzpicture}
            \centering
            \node (X_1) [obs, yshift=\yshift, scale=\nodescale] {$X_1$};
            \node (X_2) [obs, xshift=-0.6*\xshift, scale=\nodescale] {$X_2$};
            \node (X_3) [obs, xshift=0.6*\xshift, scale=\nodescale] {$X_3$};
            \path[<->] (X_1) edge[dashed] node[xshift=-.5em, yshift=.5em] {\textbf{?}} (X_2);
            \path[<->] (X_1) edge[dashed] node[xshift=.5em, yshift=.5em] {\textbf{?}} (X_3);
            \path[<->] (X_2) edge[dashed] node[yshift=.5em] {\textbf{?}} (X_3);
        \end{tikzpicture}
    }
    \qquad 
    \subfloat[]{
        \centering
        \begin{tikzpicture}
            \centering
            \node (X_1) [obs, yshift=\yshift, scale=\nodescale] {$X_1$};
            \node (X_2) [det, xshift=-0.6*\xshift, scale=\nodescale] {$x_2$};
            \node (Q) [const, yshift=.5*\yshift, xshift=-0.9*\xshift, scale=\nodescale] {{\small$\mathbb{E}[X_3\mid do(x_2)]$\textbf{?}}};
            \node (X_3) [obs, xshift=0.6*\xshift, scale=\nodescale] {$X_3$};
            \edge{X_1, X_2}{X_3};
        \end{tikzpicture}
    }
    \qquad 
     \subfloat[]{
        \centering
        \begin{tikzpicture}
            \centering
            \node (X_1) [latent, dashed, yshift=\yshift, scale=\nodescale] {$S_1$};
            \node (X_2) [latent, dashed, xshift=-0.6*\xshift, scale=\nodescale] {$S_2$};
            \node (X_3) [latent, dashed, xshift=0.6*\xshift, scale=\nodescale] {$S_3$};
            \edge[dashed]{X_1, X_2}{X_3};
            \edge[dashed]{X_1}{X_2};
            \plate[inner sep=0.2em,
        yshift=0.1em, dashed] {plate}{(X_1) (X_2) (X_3)}{};
        \node (x) [right=of plate]{\includegraphics[width=0.2\textwidth]{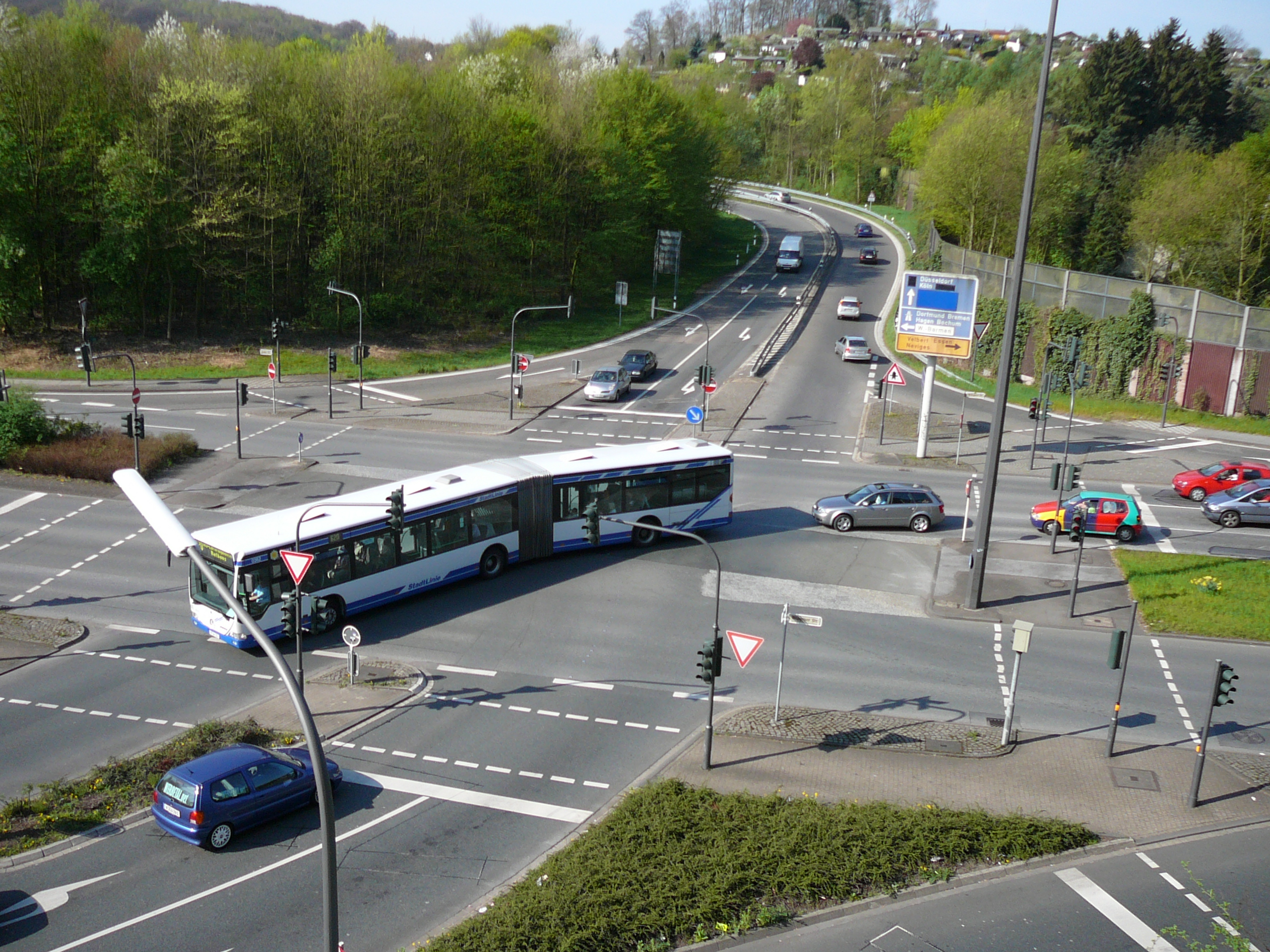}};
        \edge[dashed]{plate}{x};
        \node (f) [const, right=of plate, xshift=-0.5*\xshift, yshift=-0.5*\yshift] {$g$};
        \path[->] (x) edge[bend right=20] node[yshift=1em] {$g^{-1}$ \textbf{?}} (plate);
        \end{tikzpicture}
    }%
    \caption{Overview of different causal learning tasks: 
     (a) \textit{causal discovery}~(\cref{ch:2-discovery}) aims to learn the causal graph (or SCM) connecting a set of \textit{observed} variables%
    ;
    (b) \textit{causal reasoning}~(\cref{sec:reasoning}) aims to answer interventional or counterfactual queries based on a (partial) causal model over observed variables~$X_i$%
    ; (c) \textit{causal representation learning}~(\cref{sec:open}) aims to infer a causal model consisting of a small number of 
    high-level, abstract causal variables~$S_i$ and their relations from potentially high-dimensional, low-level observations~$\Xb=g(\mathbf{S})$.
    }
    \label{fig:overview}
\end{figure}
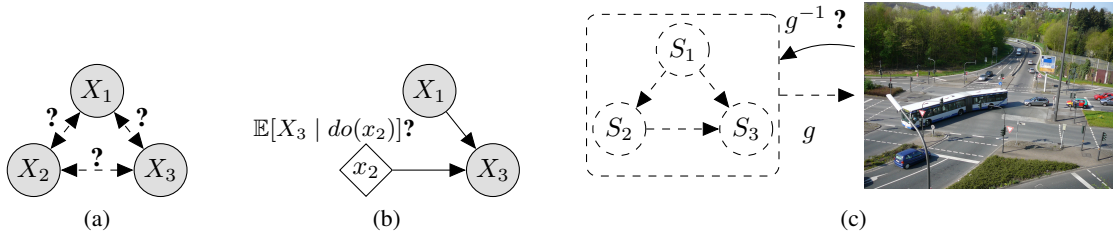

Defining objects or variables, and structural models connecting them, can sometimes be achieved by coarse-graining of microscopic models, including microscopic SCMs \cite{Rubensteinetal17}, ordinary differential equations \cite{RubBonMooSch18}, and temporally aggregated time series \cite{Gongetal17}. While most causal models in economics, medicine, or psychology use variables that are abstractions of more elementary concepts, it is challenging to state general conditions under which coarse-grained variables admit causal models with well-defined interventions \cite{1512.07942,Rubensteinetal17,chalupka2017causal}. 
 The task of identifying suitable units that admit causal models aligns with the general goal of modern machine learning to learn meaningful representations for data, where meaningful can mean \emph{robust}, \emph{transferable}, \emph{interpretable}, \emph{explainable}, or \emph{fair} \cite{NIPS2017_6995,Kilbertusetal17,ZhaBar18,karimi2020imperfect,karimi2021algorithmic,von2020fairness}.
\looseness-1  To combine structural causal modeling~(\cref{def:SCM}) and representation learning, we may try to devise machine learning models whose inputs may be high-dimensional and unstructured, but whose inner workings are (partly) governed by an SCM.
Suppose that our high-dimensional, low-level observations $\Xb=(X_1, ..., X_d)$ are explained by a small number of unobserved, or \textit{latent}, variables $\mathbf{S}=(S_1, ..., S_n)$ where $n\ll d$, in that $\Xb$ is generated by applying an injective map $g:\mathbb{R}^n \to \mathbb{R}^d$ to $\mathbf{S}$ (see~\cref{fig:overview}c):
\begin{equation}
\label{eq:decoder}
    \Xb=g(\mathbf{S}).
\end{equation}
A common assumption regarding~\eqref{eq:decoder} is that the latent $S_i$ are jointly
independent, e.g., for independent component analysis (ICA)~\cite{hyvarinen2000independent} (where $g$ is referred to as a \emph{mixing}) or disentangled representation learning~\cite{bengio2013representation} (where $g$ is called a \emph{decoder}).
Presently, however, we instead want think of the latent $S_i$ as \textit{causal variables} that support interventions and reasoning. 

The $S_i$ may thus well be dependent, and possess a causal factorization~\eqref{eq:cf},
\begin{equation}\label{eq:cf2}
p(S_1,\dots,S_n) = \prod_{i=1}^n  p(S_i \mid \PA_i),
\end{equation}
induced by an underlying  (acyclic) SCM $\mathcal{M}=(\Fb,p_\mathbf{U})$ with jointly
independent $U_i$ and 
\begin{equation}
\label{eq:latent_SCM}
    \Fb=\{S_i := f_i (\PA_i, U_i)\}_{i=1}^n.
\end{equation}
Our goal is to learn a latent causal model consisting of (i) the causal representation~$\mathbf{S}=g^{-1}(\Xb)$, along with (ii) the corresponding causal graph and (iii) the mechanisms $p(S_i \mid \PA_i)$ or~$f_i$.
This is a challenging task, since none of them are directly observed or known a priori; instead we typically only have access to observations of $\Xb$.
In fact, there is no hope in an i.i.d.\ setting since already the simpler case with independent $S_i$ (and $n=d$) is generally not identifiable (i.e., for arbitrary nonlinear $g$ in~\eqref{eq:decoder}): even independence does not sufficiently constrain the problem to uniquely recover, or identify, the true $S_i$'s up to any simple class of ambiguities such as permutations  and element-wise invertible transformations of the $S_i$~\cite{hyvarinen1999nonlinear}.

To link causal representation learning to the well-studied ICA setting with independent latents in~\eqref{eq:decoder}, we can consider the so-called \textit{reduced form} of an (acyclic) SCM: by recursive substitution of the structural assignments~\eqref{eq:latent_SCM} in topological order of the causal graph, we can write the latent causal variables $\mathbf{S}$ as function of the noise variables only
\begin{equation}
\label{eq:reduced_form}
    \mathbf{S}=f_\textsc{rf}(\mathbf{U}).
\end{equation}
Due to acyclicity, this mapping $f_\textsc{rf}:\mathbb{R}^n\to\mathbb{R}^n$ has a lower triangular Jacobian (possibly after re-ordering the $S_i$ w.l.o.g.).
However, \eqref{eq:reduced_form} is strictly less informative than~\eqref{eq:latent_SCM}: while they entail the same distribution~\eqref{eq:cf2}, the former no longer naturally supports interventions on the $S_i$ but only changes to the noise distribution $p_\mathbf{U}$ (an example of a so-called \textit{soft} intervention~\cite{eberhardt2007interventions}).
At the same time, the reduced form~\eqref{eq:reduced_form} allows us to rewrite~\eqref{eq:decoder} as
\begin{equation}
\label{eq:structured_decoder}
\Xb=g\circ f_\textsc{rf}(\mathbf{U})
\end{equation}
Through this lens, the task of learning the reduced form~\eqref{eq:reduced_form} could be seen as structured form of nonlinear ICA (i.e.,~\eqref{eq:decoder} with independent latents) where we additionally want to learn an intermediate representation through $f_\textsc{rf}$.
However, as discussed, we cannot even solve the problem with independent latents (i.e., identify $g\circ f_\textsc{rf}$ in~\eqref{eq:structured_decoder})~\cite{hyvarinen1999nonlinear}, let alone separate the SCM and mixing functions to recover the intermediate causal representation.

It is not surprising that is is not possible to solve the strictly harder causal representation learning problem in an i.i.d.\ setting and that additional causal learning signals are needed.
This gives rise to the following questions: 
{How can we devise causal training algorithms to learn the $S_i$? And, what types of additional data, assumptions, and constraints would these algorithms require beyond the i.i.d.\ setting?}
Two general ideas are to (i) build on the ICM~\cref{princ:icm} and enforce some form of (algorithmic) independence between the learned causal mechanisms $p(S_i \mid \PA_i)$ or $f_i$, and (ii) use heterogeneous (\textit{non-i.i.d.}) data, e.g., from multiple views or different environments, arising from interventions in the underlying latent SCM~\eqref{eq:latent_SCM}.
We briefly discuss some more concrete ideas based on recent work. 

\textit{Generative Approach: Causal Auto-Encoders.}
One approach is to try to learn the generative causal model~\eqref{eq:decoder} and ~\eqref{eq:latent_SCM}, or its reduced form~\eqref{eq:reduced_form}, using an \textit{auto-encoder} approach~\cite{kingma2013auto}.
An auto-encoder consists of an {\em encoder} function $q:\R^d \to \R^n$ which maps $\Xb$ to a latent ``bottleneck'' representation (e.g., comprising the unexplained noise variables~$\mathbf{U}$), and a \textit{decoder} function $\hat{g}:\R^n\to\R^d$ mapping back to the observations.
For example, the decoder may directly implement the composition $\hat{g}=g\circ f_\textsc{rf}$ from~\eqref{eq:reduced_form}. Alternatively, it could consist of multiple modules, implementing~\eqref{eq:decoder} and~\eqref{eq:latent_SCM} separately.
A standard procedure to train such an auto-encoder architecture is to minimise the reconstruction error, i.e., to satisfy $\hat{g}\circ q\approx id$ on a training set of observations of $\Xb$. 
As discussed, this alone is insufficient, so to make it causal we can impose additional constraints on the structure of the decoder~\cite{Leeb-SAE} and try to make the causal mechanisms independent by ensuring that they are invariant across problems and can be independently intervened upon.
For example, if we intervene on the causal variables $S_i$ or noise distribution $p_\mathbf{U}$ in our model of~\eqref{eq:latent_SCM} or~\eqref{eq:reduced_form}, respectively, this should still produce ``valid''
observations, as assessed, e.g., by the discriminator of a generative adversarial network \cite{GAN}.
While we ideally want to manipulate the causal variables, 
another way to intervene is to replace noise variables with the corresponding values computed from other input images, a procedure that has been referred to as hybridization \cite{besserve2020counterfactuals}.
Alternatively, if we have access to multiple environments, i.e., datasets collected under different conditions, we could rely on the Sparse Mechanism Shift~\cref{pri:scsh} by requiring that changes can be explained by shifts in only a few of the $p(S_i\mid \PA_i)$.

\commentout{
\jvk{keep any of this from previous version?}
In such an architecture, the encoder is an anticausal mapping that recognizes or reconstructs causal drivers in the world. These should be such that in terms of them, mechanisms can be formulated that are transferable (e.g., across tasks). 
The decoder establishes the connection between the low dimensional latent representation (of the noises driving the causal model) and the high dimensional world; this part constitutes a causal generative image model.
The ICM \cref{princ:icm} implies that if the latent representation reconstructs the (noises driving the) true causal variables, then interventions on those noises (and the mechanisms driven by them) are permissible and lead to valid generation of image data.
}

\textit{Discriminative Approach: Self-Supervised Causal Representation Learning.}
A different machine learning approach for unsupervised representation learning, that is not based on generative modeling but is discriminative in nature, is \textit{self-supervised learning with data augmentation}.
Here, the main idea is to apply some hand-crafted transformations to the observation to generate augmented views that are thought to share the main semantic characteristics with the original observation (e.g., random crops or blurs for images). 
One then directly learns a representation by maximizing the similarity across encodings of views related to each other by augmentations, while enforcing diversity across those of unrelated views. 
In recent work~\cite{von2021self}, we set out to better understand this approach theoretically, as well as to investigate its potential for learning causal representations.
Starting from~\eqref{eq:decoder}, we
postulate a latent causal model of the form $\mathbf{S}_c\to\mathbf{\mathbf{S}}_s$, where $\mathbf{S}_c$ is a (potentially multivariate) \textit{content} variable, defined as the high-level semantic part of the representation~$\mathbf{S}=(\mathbf{S}_c, \mathbf{S}_s)$ that is assumed invariant across views; and $\mathbf{S}_s$ is a (potentially multivariate) \textit{style} variable, defined as the remaining part of the representation that may change.
Within this setting, data augmentations have a natural interpretation as counterfactuals 
under a hypothetical intervention on the style variables, given the original view. 
It can be shown that in this case, subject to some technical assumptions, common contrastive self-supervised learning algorithms~\cite{chen2020simple,oord2018representation,gutmann2010noise} as well as appropriately constrained generative models isolate, or recover, the true content variables~$\mathbf{S}_c$ up to an invertible transformation.
By extending this approach to use multiple augmented views of the same observation, and linking these to different counterfactuals in the underlying latent SCM, it may be possible to recover a more-fine grained causal representation. 

\textit{Independent Mechanism Analysis.}
We also explored~\cite{IMA} to what extent the ICM~\cref{princ:icm} may be useful for unsupervised representation learning tasks such as~\eqref{eq:decoder}, particularly for imposing additional constraints on the mixing function $g$.
It turns out that independence between $p(\mathbf{S})$ and the mixing $g$---measured, e.g., as discussed in~\cref{sec:icm} in the context of~\cref{fig:igci} and~\cite{Janzingetal12}---does not impose nontrivial constraints when $\mathbf{S}$ is not observed, even when the $S_i$ are assumed independent as in ICA. 
However, by thinking of each $S_i$ as independently \textit{influencing} the observed distribution, we postulate another type of independence between the partial derivatives~$\frac{\partial g}{\partial S_i}$ of the mixing $g$ which has a geometric interpretation as an orthogonality condition on the columns of the Jacobian of~$g$.
The resulting \textit{independent mechanism analysis} (IMA) approach 
rules out some of the common examples of non-identifiability of nonlinear ICA~\cite{hyvarinen1999nonlinear,1811.12359}
mentioned above.
Since IMA does not require independent sources, it may also be a useful constraint for causal representation learning algorithms. 

\paragraph{Learning Transferable Mechanisms and Multi-Task Learning}
Machine learning excels in i.i.d.\ settings, and through the use of high capacity learning algorithms we can achieve outstanding performance on many problems, provided we have i.i.d.\ data for each individual problem (\cref{sec:slt}). However, natural intelligence excels at generalizing across tasks and settings. 
Suppose we want to build a system that can solve multiple tasks in multiple environments. \looseness-1 If we view learning as data compression, it would make sense for that system to utilize components that apply across tasks and environments, and thus need to be stored only once~\cite{1911.10500}.

Indeed, an artificial or natural agent in a complex world is faced with limited resources. This concerns training data, i.e., we only have limited data for each individual task/domain, and thus need to find ways of pooling/re-using data, in stark contrast to the current industry practice of large-scale labelling work done by humans. It also concerns computational resources: animals have constraints on the resources (e.g., space, energy) used by their brains, and evolutionary neuroscience knows examples where brain regions get re-purposed. Similar constraints apply as machine learning systems get embedded in physical devices that may be small and battery-powered.
Versatile AI models that robustly solve a range of problems in the real world will thus likely need to re-use components, which requires that the components are robust across tasks and environments \cite{SchJanLop16,scholkopfetal21}. This calls for a structure whose modules are maximally reusable. 
An elegant way to do this would be to employ a modular structure that mirrors modularity that exists in the world. In other words, if the are mechanisms at play in the world play similar roles across a range of environments, tasks, and settings, then it would be prudent for a model to employ corresponding computational modules \cite{RIMs}. For instance, if variations of natural lighting (the position of the sun, clouds, etc.) imply that the visual environment can appear in brightness conditions spanning several orders of magnitude, then visual processing algorithms in our nervous system should employ methods that can factor out these variations, rather than building separate sets of object recognizers for every lighting condition. If our brain were to model the lighting changes by a gain control mechanism, say, then this mechanism in itself need not have anything to do with the physical mechanisms bringing about brightness differences. It would, however, play a role in a modular structure that corresponds to the role the physical mechanisms play in the world's modular structure---in other words, it would \emph{represent} the physical mechanism. Searching for the most versatile yet compact models would then automatically produce a bias towards models that exhibit certain forms of structural isomorphy to a world that we cannot directly recognize.

A sensible inductive bias to learn such models is to look for independent causal mechanisms \cite{Locatello_Mixture}, and competitive training can play a role in this: for a pattern recognition task, learning causal models that contain independent mechanisms helps in transferring modules across substantially different domains \cite{ParKilRojSch18}.

\paragraph{Interventional World Models, Surrogate Models, Digital Twins, and Reasoning}
Modern representation learning excels at learning representations of data that preserve relevant statistical properties \cite{bengio2013representation,LeCBenHin15}. It does so, however, without taking into account causal properties of the variables, i.e., it does not care about the interventional properties of the variables it analyzes or reconstructs.
Going forward, causality will play a major role in taking representation learning to the next level, moving beyond the representation of statistical dependence structures towards models that support intervention, planning, and reasoning. This would realize Konrad Lorenz' notion of \emph{thinking} as \emph{acting in an imagined space}. It would also provide a means to learn causal \emph{digital twins} that go beyond reproducing statistical dependences captured by \emph{surrogate models} trained using machine learning. 

The idea of surrogate modeling is that we may have a complex phenomenon for which we have access to computationally expensive simulation data. If the mappings involved (e.g., from parameter settings to target quantities) can be fitted from data, we can employ machine learning, which will often speed them up by orders of magnitude. Such a speed-up can qualitatively change the usability of a model: for instance, we have recently built a system to map gravitational wave measurements to a probability distribution of physical parameters of a black hole merger event, including sky position \cite{daxetal21}. The fact that this model only requires seconds to evaluate makes it possible to immediately start electromagnetic follow-up observations using telescopes as soon as a gravitational wave event has been detected, enabling analysis of transient events.

Going forward, we anticipate that surrogate modeling will benefit from respecting the causal factorization \eq{eq:cf} decomposing the overall dependence structure into mechanisms (i.e., causal conditionals). We can then build an overall model of a system by modeling the mechanisms independently, each of them using the optimal method. Some of the conditionals we may know analytically, some we may be able to transfer from related problems, if they are invariant. For some, we may have access to real data to estimate them, and for others, we may need to resort to simulations, possibly fitted using surrogate models.

If the model is required to fully capture the effects of all possible interventions, then all components should be fitted as described in the causal directions (i.e., we fit the causal mechanisms). Such a model then allows to employ all the causal reasoning machinery described in~\cref{sec:causal} and~\cref{sec:reasoning} (e.g., computing interventional and, in the case of SCMs, counterfactual distributions). 
If, on the other hand, a model only needs to capture \textit{some} of the possible interventions, and is used in a purely predictive/observational mode for other variables, then we can get away with also using and fitting some non-causal modules, i.e., using a decomposition which lies in between \eq{eq:cf} and \eq{eq:non-cf}.

We believe that this overall framework will be a principled and powerful approach to build such (causal) digital twins or causal surrogate models by combining a range of methods and bringing them to bear according to their strengths.

\paragraph{Concluding Remarks.}
Most of the discussed fields are still in their infancy, and the above account is biased by personal taste and knowledge. With the current hype around machine learning, there is much to say in favor of some humility towards what machine learning can do, and thus towards the current state of AI---the hard problems have not been solved yet, making basic research in this field all the more exciting.

\paragraph{Ackowledgements}
Many thanks to all past and present members of the T\"ubingen causality team, and to Cian Eastwood and Elias Bareinboim for feedback on the manuscript.

\bibliographystyle{emss}  
{\small
\bibliography{ICM-references,references_SAB2022,references_julius}

\begin{thebibliography}{100}
\providecommand{\url}[1]{\texttt{#1}}
\providecommand{\urlprefix}{URL }
\providecommand{\eprint}[2][]{\url{#2}}

\bibitem{Aldrich89}
J.~Aldrich, Autonomy. \emph{Oxford Economic Papers} \textbf{41} (1989), 15--34

\bibitem{angrist1996identification}
J.~D. Angrist, G.~W. Imbens, and D.~B. Rubin, Identification of causal effects
  using instrumental variables. \emph{Journal of the American statistical
  Association} \textbf{91} (1996), no. 434, 444--455

\bibitem{arjovsky2019invariant}
M.~Arjovsky, L.~Bottou, I.~Gulrajani, and D.~Lopez-Paz, Invariant risk
  minimization. \emph{arXiv preprint 1907.02893}  (2019)

\bibitem{Bareinboim2014}
E.~Bareinboim and J.~Pearl, Transportability from multiple environments with
  limited experiments: Completeness results. In \emph{{A}dvances in {N}eural
  {I}nformation {P}rocessing {S}ystems 27}, pp. 280--288, 2014

\bibitem{bareinboim2016causal}
E.~Bareinboim and J.~Pearl, Causal inference and the data-fusion problem.
  \emph{Proceedings of the National Academy of Sciences} \textbf{113} (2016),
  no.~27, 7345--7352

\bibitem{BauSchPet16}
S.~Bauer, B.~Sch{\"o}lkopf, and J.~Peters, The arrow of time in multivariate
  time series. In \emph{Proceedings of the 33nd international conference on
  machine learning}, pp. 2043--2051, 48, 2016

\bibitem{belkin}
M.~Belkin, D.~Hsu, S.~Ma, and S.~Mandal, Reconciling modern machine learning
  practice and the bias-variance trade-off. 2018, \eprint{arXiv:1812.11118}

\bibitem{ben2010impossibility}
S.~Ben-David, T.~Lu, T.~Luu, and D.~P{\'a}l, Impossibility theorems for domain
  adaptation. In \emph{Proceedings of the international conference on
  artificial intelligence and statistics 13 ({AISTATS})}, pp. 129--136, 2010

\bibitem{bengio2013representation}
Y.~Bengio, A.~Courville, and P.~Vincent, Representation learning: A review and
  new perspectives. \emph{IEEE Transactions on Pattern Analysis and Machine
  Intelligence} \textbf{35} (2013), no.~8, 1798--1828

\bibitem{1909.04642}
B.~Benneke, I.~Wong, C.~Piaulet, H.~A. Knutson, I.~J.~M. Crossfield,
  J.~Lothringer, C.~V. Morley, P.~Gao, T.~P. Greene, C.~Dressing, D.~Dragomir,
  A.~W. Howard, P.~R. McCullough, E.~M. R. K. J.~J. Fortney, and J.~Fraine,
  Water vapor on the habitable-zone exoplanet {K2}-18b. \emph{arXiv preprint
  1909.04642}  (2019)

\bibitem{besserve2020counterfactuals}
M.~Besserve, A.~Mehrjou, R.~Sun, and B.~Sch{\"o}lkopf, Counterfactuals uncover
  the modular structure of deep generative models. In \emph{International
  conference on learning representations}, 2020

\bibitem{BesShaSchJan18}
M.~Besserve, N.~Shajarisales, B.~Sch{\"o}lkopf, and D.~Janzing, Group
  invariance principles for causal generative models. In \emph{Proceedings of
  the 21st international conference on artificial intelligence and statistics
  (aistats)}, pp. 557--565, 2018

\bibitem{bongers2021foundations}
S.~Bongers, P.~Forr{\'e}, J.~Peters, and J.~M. Mooij, Foundations of structural
  causal models with cycles and latent variables. \emph{The Annals of
  Statistics} \textbf{49} (2021), no.~5, 2885--2915

\bibitem{buchsbaum2012power}
D.~Buchsbaum, S.~Bridgers, D.~Skolnick~Weisberg, and A.~Gopnik, The power of
  possibility: Causal learning, counterfactual reasoning, and pretend play.
  \emph{Philosophical Transactions of the Royal Society B: Biological Sciences}
  \textbf{367} (2012), no. 1599, 2202--2212

\bibitem{1512.07942}
K.~Chalupka, F.~Eberhardt, and P.~Perona, Multi-level cause-effect systems. In
  \emph{Artificial intelligence and statistics}, pp. 361--369, PMLR, 2016

\bibitem{chalupka2017causal}
K.~Chalupka, F.~Eberhardt, and P.~Perona, Causal feature learning: an overview.
  \emph{Behaviormetrika} \textbf{44} (2017), no.~1, 137--164

\bibitem{ChaSchZie06}
O.~Chapelle, B.~Sch{\"o}lkopf, and A.~Zien (eds.), \emph{Semi-supervised
  learning}. MIT Press, Cambridge, MA, USA, 2006

\bibitem{charig1986comparison}
C.~R. Charig, D.~R. Webb, S.~R. Payne, and J.~E. Wickham, Comparison of
  treatment of renal calculi by open surgery, percutaneous nephrolithotomy, and
  extracorporeal shockwave lithotripsy. \emph{Br Med J (Clin Res Ed)}
  \textbf{292} (1986), no. 6524, 879--882

\bibitem{chen2020simple}
T.~Chen, S.~Kornblith, M.~Norouzi, and G.~Hinton, A simple framework for
  contrastive learning of visual representations. \emph{arXiv preprint
  2002.05709}  (2020)

\bibitem{chickering1996learning}
D.~M. Chickering, Learning bayesian networks is np-complete. In \emph{Learning
  from data}, pp. 121--130, Springer, 1996

\bibitem{chickering2002optimal}
D.~M. Chickering, Optimal structure identification with greedy search.
  \emph{Journal of machine learning research} \textbf{3} (2002), no. Nov,
  507--554

\bibitem{cooper1992bayesian}
G.~F. Cooper and E.~Herskovits, A bayesian method for the induction of
  probabilistic networks from data. \emph{Machine learning} \textbf{9} (1992),
  no.~4, 309--347

\bibitem{cox1958planning}
D.~R. Cox, Planning of experiments  (1958)

\bibitem{Daniusisetal10}
P.~Daniu\v{s}is, D.~Janzing, J.~M. Mooij, J.~Zscheischler, B.~Steudel,
  K.~Zhang, and B.~Sch{\"o}lkopf, Inferring deterministic causal relations. In
  \emph{Proceedings of the 26th annual conference on {U}ncertainty in
  {A}rtificial {I}ntelligence ({UAI})}, pp. 143--150, 2010

\bibitem{Dawid79}
A.~P. Dawid, Conditional independence in statistical theory. \emph{Journal of
  the Royal Statistical Society B} \textbf{41} (1979), no.~1, 1--31

\bibitem{dawid2000causal}
A.~P. Dawid, Causal inference without counterfactuals. \emph{Journal of the
  American Statistical Association} \textbf{95} (2000), no. 450, 407--424

\bibitem{daxetal21}
M.~Dax, S.~R. Green, J.~Gair, J.~H. Macke, A.~Buonanno, and B.~Sch{\"o}lkopf,
  Real-time gravitational-wave science with neural posterior estimation.
  \emph{Physical Review Letters}  (2021)

\bibitem{DevGyoLug96}
L.~Devroye, L.~Gy{\"o}rfi, and G.~Lugosi, \emph{A probabilistic theory of
  pattern recognition}. Applications of Mathematics 31, Springer, New York, NY,
  1996

\bibitem{didelez2010assumptions}
V.~Didelez, S.~Meng, and N.~A. Sheehan, Assumptions of {IV} methods for
  observational epidemiology. \emph{Statistical Science} \textbf{25} (2010),
  22--40

\bibitem{eberhardt2007interventions}
F.~Eberhardt and R.~Scheines, Interventions and causal inference.
  \emph{Philosophy of Science} \textbf{74} (2007), no.~5, 981--995

\bibitem{fisher1937design}
R.~A. Fisher, The design of experiments. \emph{Oliver \& Boyd, Edinburgh \&
  London.}  (1937), no.~2

\bibitem{Foreman-Mackeyetal15}
D.~Foreman-Mackey, B.~T. Montet, D.~W. Hogg, T.~D. Morton, D.~Wang, and
  B.~Sch{\"o}lkopf, A systematic search for transiting planets in the {K2}
  data. \emph{The Astrophysical Journal} \textbf{806} (2015), no.~2

\bibitem{FukGreSunSch08}
K.~Fukumizu, A.~Gretton, X.~Sun, and B.~Sch{\"o}lkopf, Kernel measures of
  conditional dependence. In \emph{Advances in neural information processing
  systems}, pp. 489--496, 2008

\bibitem{geiger1994learning}
D.~Geiger and D.~Heckerman, Learning gaussian networks. In \emph{Proceedings of
  the tenth international conference on uncertainty in artificial
  intelligence}, pp. 235--243, 1994

\bibitem{GeiPea90}
D.~Geiger and J.~Pearl, Logical and algorithmic properties of independence and
  their application to {Bayesian} networks. \emph{Annals of Mathematics and
  Artificial Intelligence} \textbf{2} (1990), 165–178

\bibitem{GonZhaLiuTaoSch16}
M.~Gong, K.~Zhang, T.~Liu, D.~Tao, C.~Glymour, and B.~Sch{\"o}lkopf, Domain
  adaptation with conditional transferable components. In \emph{Proceedings of
  the 33nd international conference on machine learning}, pp. 2839--2848, 2016

\bibitem{Gongetal17}
M.~Gong, K.~Zhang, B.~Sch{\"o}lkopf, C.~Glymour, and D.~Tao, Causal discovery
  from temporally aggregated time series. In \emph{Proceedings of the
  thirty-third conference on uncertainty in artificial intelligence}, 2017

\bibitem{GAN}
I.~Goodfellow, J.~Pouget-Abadie, M.~Mirza, B.~Xu, D.~Warde-Farley, S.~Ozair,
  A.~Courville, and Y.~Bengio, Generative adversarial nets. In \emph{Advances
  in neural information processing systems 27}, pp. 2672--2680, 2014

\bibitem{RIMs}
A.~Goyal, A.~Lamb, J.~Hoffmann, S.~Sodhani, S.~Levine, Y.~Bengio, and
  B.~Sch{\"o}lkopf, Recurrent independent mechanisms. In \emph{International
  conference on learning representations}, 2020

\bibitem{IMA}
L.~Gresele, J.~von K{\"{u}}gelgen, V.~Stimper, B.~Sch{\"{o}}lkopf, and
  M.~Besserve, Independent mechanism analysis, a new concept? In \emph{Advances
  in neural information processing systems 34}, 2021

\bibitem{GreBorRasSchetal07}
A.~Gretton, K.~M. Borgwardt, M.~Rasch, B.~Sch{\"o}lkopf, and A.~J. Smola, A
  kernel method for the two-sample-problem. In \emph{Advances in neural
  information processing systems 19}, pp. 513--520, 2007

\bibitem{Gretton2005}
A.~Gretton, O.~Bousquet, A.~Smola, and B.~Sch\"{o}lkopf, Measuring statistical
  dependence with {H}ilbert-{S}chmidt norms. In \emph{Algorithmic learning
  theory}, pp. 63--78, Springer-Verlag, 2005

\bibitem{GreFukTeoSonetal08}
A.~Gretton, K.~Fukumizu, C.~H. Teo, L.~Song, B.~Sch{\"o}lkopf, and A.~J. Smola,
  A kernel statistical test of independence. In \emph{Advances in neural
  information processing systems 20}, pp. 585--592, 2008

\bibitem{Gretton2005JMLR}
A.~Gretton, R.~Herbrich, A.~Smola, O.~Bousquet, and B.~Sch\"olkopf, Kernel
  methods for measuring independence. \emph{Journal of Machine Learning
  Research} \textbf{6} (2005), 2075--2129

\bibitem{gutmann2010noise}
U.~M. Gutmann and A.~Hyv{\"a}rinen, Noise-contrastive estimation: A new
  estimation principle for unnormalized statistical models. In
  \emph{International conference on artificial intelligence and statistics},
  pp. 297--304, 2010

\bibitem{haavelmo1944probability}
T.~Haavelmo, The probability approach in econometrics. \emph{Econometrica:
  Journal of the Econometric Society}  (1944), iii--115

\bibitem{heckerman1995learning}
D.~Heckerman, D.~Geiger, and D.~M. Chickering, Learning bayesian networks: The
  combination of knowledge and statistical data. \emph{Machine learning}
  \textbf{20} (1995), no.~3, 197--243

\bibitem{heckerman2006bayesian}
D.~Heckerman, C.~Meek, and G.~Cooper, A bayesian approach to causal discovery.
  In \emph{Innovations in machine learning}, pp. 1--28, Springer, 2006

\bibitem{hernan2011simpson}
M.~A. Hern{\'a}n, D.~Clayton, and N.~Keiding, The {S}impson's paradox
  unraveled. \emph{International journal of epidemiology} \textbf{40} (2011),
  no.~3, 780--785

\bibitem{higgins2016beta}
I.~Higgins, L.~Matthey, A.~Pal, C.~Burgess, X.~Glorot, M.~Botvinick,
  S.~Mohamed, and A.~Lerchner, Beta-{VAE}: Learning basic visual concepts with
  a constrained variational framework. In \emph{International conference on
  learning representations}, 2016

\bibitem{holland1986statistics}
P.~W. Holland, Statistics and causal inference. \emph{Journal of the American
  statistical Association} \textbf{81} (1986), no. 396, 945--960

\bibitem{hoover2001causality}
K.~D. Hoover, \emph{Causality in macroeconomics}. Cambridge University Press,
  2001

\bibitem{Hoyer2008}
P.~O. Hoyer, D.~Janzing, J.~M. Mooij, J.~Peters, and B.~Sch\"olkopf, Nonlinear
  causal discovery with additive noise models. In \emph{{A}dvances in {N}eural
  {I}nformation {P}rocessing {S}ystems 21 ({NIPS})}, pp. 689--696, 2009

\bibitem{huang2019causal}
B.~Huang, K.~Zhang, J.~Zhang, J.~D. Ramsey, R.~Sanchez-Romero, C.~Glymour, and
  B.~Sch{\"o}lkopf, Causal discovery from heterogeneous/nonstationary data.
  \emph{J. Mach. Learn. Res.} \textbf{21} (2020), no.~89, 1--53

\bibitem{HuaZhaZhaSanGlySch17}
B.~Huang, K.~Zhang, J.~Zhang, R.~Sanchez-Romero, C.~Glymour, and
  B.~Sch{\"o}lkopf, Behind distribution shift: Mining driving forces of changes
  and causal arrows. In \emph{{IEEE} 17th international conference on data
  mining (icdm 2017)}, pp. 913--918, 2017

\bibitem{huang2006pearl}
Y.~Huang and M.~Valtorta, Pearl's calculus of intervention is complete. In
  \emph{Proceedings of the twenty-second conference on uncertainty in
  artificial intelligence}, pp. 217--224, 2006

\bibitem{hyvarinen2000independent}
A.~Hyv{\"a}rinen and E.~Oja, Independent component analysis: algorithms and
  applications. \emph{Neural networks} \textbf{13} (2000), no. 4-5, 411--430

\bibitem{hyvarinen1999nonlinear}
A.~Hyv{\"a}rinen and P.~Pajunen, Nonlinear independent component analysis:
  Existence and uniqueness results. \emph{Neural networks} \textbf{12} (1999),
  no.~3, 429--439

\bibitem{hyvarinen2018nonlinear}
A.~Hyvarinen, H.~Sasaki, and R.~Turner, Nonlinear ica using auxiliary variables
  and generalized contrastive learning. In \emph{The 22nd international
  conference on artificial intelligence and statistics}, pp. 859--868, 2019

\bibitem{imbens2008regression}
G.~W. Imbens and T.~Lemieux, Regression discontinuity designs: A guide to
  practice. \emph{Journal of econometrics} \textbf{142} (2008), no.~2, 615--635

\bibitem{imbens2015causal}
G.~W. Imbens and D.~B. Rubin, \emph{Causal inference in statistics, social, and
  biomedical sciences}. Cambridge University Press, 2015

\bibitem{Janzing2016}
D.~Janzing, R.~Chaves, and B.~Sch\"olkopf, Algorithmic independence of initial
  condition and dynamical law in thermodynamics and causal inference. \emph{New
  Journal of Physics} \textbf{18} (2016), no. 093052, 1--13

\bibitem{JanHoySch10}
D.~Janzing, P.~Hoyer, and B.~Sch{\"o}lkopf, Telling cause from effect based on
  high-dimensional observations. In \emph{Proceedings of the 27th international
  conference on machine learning}, edited by J.~F{\"u}rnkranz and T.~Joachims,
  pp. 479--486, 2010

\bibitem{Janzingetal12}
D.~Janzing, J.~M. Mooij, K.~Zhang, J.~Lemeire, J.~Zscheischler,
  P.~Daniu\v{s}is, B.~Steudel, and B.~Sch\"olkopf, Information-geometric
  approach to inferring causal directions. \emph{Artificial Intelligence}
  \textbf{182--183} (2012), 1--31

\bibitem{Janzing2009uai}
D.~Janzing, J.~Peters, J.~M. Mooij, and B.~Sch\"{o}lkopf, Identifying
  confounders using additive noise models. In \emph{Proceedings of the 25th
  annual conference on {U}ncertainty in {A}rtificial {I}ntelligence ({UAI})},
  pp. 249--257, 2009

\bibitem{JanSch10}
D.~Janzing and B.~Sch{\"o}lkopf, Causal inference using the algorithmic
  {M}arkov condition. \emph{IEEE Transactions on Information Theory}
  \textbf{56} (2010), no.~10, 5168--5194

\bibitem{JanSch15}
D.~Janzing and B.~Sch{{\"o}}lkopf, Semi-supervised interpolation in an
  anticausal learning scenario. \emph{Journal of Machine Learning Research}
  \textbf{16} (2015), 1923--1948

\bibitem{JanSch18b}
D.~Janzing and B.~Sch{\"o}lkopf, Detecting non-causal artifacts in multivariate
  linear regression models. In \emph{Proceedings of the 35th international
  conference on machine learning ({ICML})}, pp. 2250--2258, 2018

\bibitem{jinetal21}
Z.~Jin, J.~von K{\"u}gelgen, J.~Ni, T.~Vaidhya, A.~Kaushal, M.~Sachan, and
  B.~Sch{\"o}lkopf, Causal direction of data collection matters: Implications
  of causal and anticausal learning for nlp. In \emph{Proceedings of the 2021
  conference on empirical methods in natural language processing (emnlp)}, 2021

\bibitem{karimi2021algorithmic}
A.-H. Karimi, B.~Sch{\"o}lkopf, and I.~Valera, Algorithmic recourse: from
  counterfactual explanations to interventions. In \emph{Conference on
  fairness, accountability, and transparency}, pp. 353--362, 2021

\bibitem{karimi2020imperfect}
A.-H. Karimi, J.~von K{\"{u}}gelgen, B.~Sch{\"{o}}lkopf, and I.~Valera,
  Algorithmic recourse under imperfect causal knowledge: a probabilistic
  approach. In \emph{Advances in neural information processing systems 33},
  2020

\bibitem{Kilbertusetal17}
N.~Kilbertus, M.~Rojas~Carulla, G.~Parascandolo, M.~Hardt, D.~Janzing, and
  B.~Sch\"{o}lkopf, Avoiding discrimination through causal reasoning. In
  \emph{Advances in neural information processing systems 30}, pp. 656--666,
  2017

\bibitem{kingma2013auto}
D.~P. Kingma and M.~Welling, Auto-encoding variational {B}ayes. \emph{arXiv
  preprint arXiv:1312.6114}  (2013)

\bibitem{klein1872vergleichende}
F.~Klein, \emph{Vergleichende {B}etrachtungen {\"u}ber neuere geometrische
  {F}orschungen}. Verlag von Andreas Deichert, Erlangen, 1872

\bibitem{Koller2009}
D.~Koller and N.~Friedman, \emph{Probabilistic graphical models: principles and
  techniques}. MIT press, 2009

\bibitem{Kpotufe14}
S.~Kpotufe, E.~Sgouritsa, D.~Janzing, and B.~Sch{\"{o}}lkopf, Consistency of
  causal inference under the additive noise model. In \emph{Proceedings of the
  31th international conference on machine learning}, pp. 478--486, 2014

\bibitem{NIPS2017_6995}
M.~J. Kusner, J.~Loftus, C.~Russell, and R.~Silva, Counterfactual fairness. In
  \emph{Advances in neural information processing systems 30}, pp. 4066--4076,
  Curran Associates, Inc., 2017

\bibitem{lauritzen1996graphical}
S.~L. Lauritzen, \emph{Graphical models}. 17, Clarendon Press, 1996

\bibitem{LeCBenHin15}
Y.~LeCun, Y.~Bengio, and G.~Hinton, Deep learning. \emph{{N}ature} \textbf{521}
  (2015), no. 7553, 436--444

\bibitem{Leeb-SAE}
F.~Leeb, Y.~Annadani, S.~Bauer, and B.~Schölkopf, Structural autoencoders
  improve representations for generation and transfer. \emph{arXiv preprint
  2006.07796}  (2020)

\bibitem{LeibnizDiscours}
G.~W. Leibniz, Discours de m{\'e}taphysique. 1686, (cited after Chaitin, 2010)

\bibitem{1811.12359}
F.~Locatello, S.~Bauer, M.~Lucic, G.~Rätsch, S.~Gelly, B.~Schölkopf, and
  O.~Bachem, Challenging common assumptions in the unsupervised learning of
  disentangled representations. \emph{Proceedings of the 36th International
  Conference on Machine Learning}  (2019)

\bibitem{Locatello_Mixture}
F.~Locatello, D.~Vincent, I.~Tolstikhin, G.~Rätsch, S.~Gelly, and
  B.~Schölkopf, Competitive training of mixtures of independent deep
  generative models. \emph{arXiv preprint 1804.11130}  (2018)

\bibitem{locatello2020object}
F.~Locatello, D.~Weissenborn, T.~Unterthiner, A.~Mahendran, G.~Heigold,
  J.~Uszkoreit, A.~Dosovitskiy, and T.~Kipf, Object-centric learning with slot
  attention. In \emph{Advances in neural information processing systems}, 2020

\bibitem{LopNisChiSchBot17}
D.~Lopez-Paz, R.~Nishihara, S.~Chintala, B.~Sch{\"o}lkopf, and L.~Bottou,
  Discovering causal signals in images. In \emph{Ieee conference on computer
  vision and pattern recognition (cvpr)}, pp. 58--66, 2017

\bibitem{loschmidt1876ueber}
J.~Loschmidt, {\"Uber} den {Z}ustand des {W\"a}rmegleichgewichtes eines
  {S}ystems von {K\"o}rpern mit {R\"ucksicht} auf die {Schwerkraft}.
  \emph{Akademie der Wissenschaften, Wien. Mathematisch-Naturwissenschaftliche
  Klasse, Sitzungsberichte} \textbf{73} (1876), 128--142

\bibitem{2102.12353}
C.~Lu, Y.~Wu, J.~M. Hernández-Lobato, and B.~Schölkopf, Nonlinear invariant
  risk minimization: A causal approach. 2021, arXiv:2102.12353

\bibitem{maclane:71}
S.~MacLane, \emph{Categories for the working mathematician}. Springer-Verlag,
  New York, 1971

\bibitem{matthews2000storks}
R.~Matthews, Storks deliver babies (p= 0.008). \emph{Teaching Statistics}
  \textbf{22} (2000), no.~2, 36--38

\bibitem{meek1995causal}
C.~Meek, Causal inference and causal explanation with background knowledge. In
  \emph{Proceedings of the eleventh conference on uncertainty in artificial
  intelligence}, pp. 403--410, Morgan Kaufmann Publishers Inc., 1995

\bibitem{messerli2012chocolate}
F.~H. Messerli, Chocolate consumption, cognitive function, and nobel laureates.
  \emph{The New England Journal of Medicine} \textbf{367} (2012), no.~16,
  1562--1564

\bibitem{Montet_2015}
B.~T. Montet, T.~D. Morton, D.~Foreman-Mackey, J.~A. Johnson, D.~W. Hogg, B.~P.
  Bowler, D.~W. Latham, A.~Bieryla, and A.~W. Mann, Stellar and planetary
  properties of {K2} campaign 1 candidates and validation of 17 planets,
  including a planet receiving earth-like insolation. \emph{The Astrophysical
  Journal} \textbf{809} (2015), no.~1, 25

\bibitem{monti2019causal}
R.~P. Monti, K.~Zhang, and A.~Hyv{\"a}rinen, Causal discovery with general
  non-linear relationships using non-linear ica. In \emph{Uncertainty in
  artificial intelligence}, pp. 186--195, PMLR, 2020

\bibitem{Mooij11}
J.~M. Mooij, D.~Janzing, T.~Heskes, and B.~Sch{\"o}lkopf, On causal discovery
  with cyclic additive noise models. In \emph{{A}dvances in {N}eural
  {I}nformation {P}rocessing {S}ystems 24 ({NIPS})}, 2011

\bibitem{Mooij2009}
J.~M. Mooij, D.~Janzing, J.~Peters, and B.~Sch\"{o}lkopf, Regression by
  dependence minimization and its application to causal inference. In
  \emph{Proceedings of the 26th international conference on machine learning
  ({ICML})}, pp. 745--752, 2009

\bibitem{MooJanSch13}
J.~M. Mooij, D.~Janzing, and B.~Sch{\"o}lkopf, From ordinary differential
  equations to structural causal models: The deterministic case. In
  \emph{Proceedings of the 29th annual conference on uncertainty in artificial
  intelligence ({UAI})}, pp. 440--448, 2013

\bibitem{Mooijetal16}
J.~M. Mooij, J.~Peters, D.~Janzing, J.~Zscheischler, and B.~Sch\"olkopf,
  Distinguishing cause from effect using observational data: methods and
  benchmarks. \emph{Journal of Machine Learning Research} \textbf{17} (2016),
  no.~32, 1--102

\bibitem{neyman1923application}
J.~S. Neyman, On the application of probability theory to agricultural
  experiments. essay on principles. section 9.(tlanslated and edited by dm
  dabrowska and tp speed, statistical science (1990), 5, 465-480). \emph{Annals
  of Agricultural Sciences} \textbf{10} (1923), 1--51

\bibitem{oord2018representation}
A.~v.~d. Oord, Y.~Li, and O.~Vinyals, Representation learning with contrastive
  predictive coding. \emph{arXiv preprint 1807.03748}  (2018)

\bibitem{ParKilRojSch18}
G.~Parascandolo, N.~Kilbertus, M.~Rojas-Carulla, and B.~Sch{\"o}lkopf, Learning
  independent causal mechanisms. In \emph{Proceedings of the 35th international
  conference on machine learning, pmlr 80:4036-4044}, 2018

\bibitem{parmua20}
J.~Park and K.~Muandet, A measure-theoretic approach to kernel conditional mean
  embeddings. In \emph{Advances in neural information processing systems 33
  (neurips 2020)}, pp. 21247--21259, Curran Associates, Inc., 2020

\bibitem{pearl1985bayesian}
J.~Pearl, Bayesian networks: A model of self-activated memory for evidential
  reasoning. In \emph{Proceedings of the 7th conference of the cognitive
  science society, 1985}, pp. 329--334, 1985

\bibitem{Pearl1995}
J.~Pearl, Causal diagrams for empirical research. \emph{Biometrika} \textbf{82}
  (1995), no.~4, 669--688

\bibitem{pearl2001direct}
J.~Pearl, Direct and indirect effects. In \emph{Proceedings of the seventeenth
  conference on uncertainty in artificial intelligence}, pp. 411--420, 2001

\bibitem{Pearl2009}
J.~Pearl, \emph{Causality: Models, reasoning, and inference}. 2nd edn.,
  Cambridge University Press, New York, NY, 2009

\bibitem{pearl2014comment}
J.~Pearl, Comment: understanding simpson’s paradox. \emph{The American
  Statistician} \textbf{68} (2014), no.~1, 8--13

\bibitem{pearl2014external}
J.~Pearl and E.~Bareinboim, External validity: From do-calculus to
  transportability across populations. \emph{Statistical Science} \textbf{29}
  (2014), no.~4, 579--595

\bibitem{pearl2018book}
J.~Pearl and D.~Mackenzie, \emph{The book of why: the new science of cause and
  effect}. Basic Books, 2018

\bibitem{pearl2014confounding}
J.~Pearl and A.~Paz, Confounding equivalence in causal inference. \emph{Journal
  of Causal Inference} \textbf{2} (2014), no.~1, 75--93

\bibitem{pearl1991theory}
J.~Pearl and T.~Verma, A theory of inferred causation. In \emph{Principles of
  knowledge representation and reasoning: Proceedings of the second
  international conference}, p. 441, 2, 1991

\bibitem{peters2016causal}
J.~Peters, P.~B{\"u}hlmann, and N.~Meinshausen, Causal inference by using
  invariant prediction: identification and confidence intervals. \emph{Journal
  of the Royal Statistical Society: Series B (Statistical Methodology)}
  \textbf{78} (2016), no.~5, 947--1012

\bibitem{PetJanSch17}
J.~Peters, D.~Janzing, and B.~Sch{\"o}lkopf, \emph{Elements of causal inference
  - foundations and learning algorithms}. MIT Press, Cambridge, MA, USA, 2017

\bibitem{Peters2011b}
J.~Peters, J.~M. Mooij, D.~Janzing, and B.~Sch\"{o}lkopf, Identifiability of
  causal graphs using functional models. In \emph{Proceedings of the 27th
  annual conference on {U}ncertainty in {A}rtificial {I}ntelligence ({UAI})},
  pp. 589--598, 2011

\bibitem{PetMooJanSch14}
J.~Peters, J.~M. Mooij, D.~Janzing, and B.~Sch{\"o}lkopf, Causal discovery with
  continuous additive noise models. \emph{Journal of Machine Learning Research}
  \textbf{15} (2014), 2009--2053

\bibitem{PfiBuhSchPet18}
N.~Pfister, P.~B{\"u}hlmann, B.~Sch{\"o}lkopf, and J.~Peters, Kernel-based
  tests for joint independence. \emph{Journal of the Royal Statistical Society:
  Series B (Statistical Methodology)} \textbf{80} (2018), no.~1, 5--31

\bibitem{popper1959logic}
K.~Popper, The logic of scientific discovery  (1959)

\bibitem{Reichenbach1956}
H.~Reichenbach, \emph{The direction of time}. University of California Press,
  Berkeley, CA, 1956

\bibitem{robins1986new}
J.~Robins, A new approach to causal inference in mortality studies with a
  sustained exposure period—application to control of the healthy worker
  survivor effect. \emph{Mathematical modelling} \textbf{7} (1986), no. 9-12,
  1393--1512

\bibitem{robins2000marginal}
J.~M. Robins, M.~A. Hernan, and B.~Brumback, Marginal structural models and
  causal inference in epidemiology. \emph{Epidemiology} \textbf{11} (2000),
  no.~5, 550--560

\bibitem{robinson1973counting}
R.~W. Robinson, Counting labeled acyclic digraphs, new directions in the theory
  of graphs (proc. third ann arbor conf., univ. michigan, ann arbor, mich.,
  1971). 1973

\bibitem{RojSchTurPet18}
M.~Rojas-Carulla, B.~Sch{\"o}lkopf, R.~Turner, and J.~Peters, Invariant models
  for causal transfer learning. \emph{Journal of Machine Learning Research}
  \textbf{19} (2018), no.~36, 1--34

\bibitem{rosenbaum1983central}
P.~R. Rosenbaum and D.~B. Rubin, The central role of the propensity score in
  observational studies for causal effects. \emph{Biometrika} \textbf{70}
  (1983), no.~1, 41--55

\bibitem{RubBonMooSch18}
P.~K. Rubenstein, S.~Bongers, B.~Sch{\"o}lkopf, and J.~M. Mooij, From
  deterministic {ODEs} to dynamic structural causal models. In
  \emph{Proceedings of the 34th conference on uncertainty in artificial
  intelligence (uai)}, 2018

\bibitem{Rubensteinetal17}
P.~K. Rubenstein, S.~Weichwald, S.~Bongers, J.~M. Mooij, D.~Janzing,
  M.~Grosse-Wentrup, and B.~Sch{\"o}lkopf, Causal consistency of structural
  equation models. In \emph{Proceedings of the thirty-third conference on
  uncertainty in artificial intelligence}, 2017

\bibitem{rubin1974estimating}
D.~B. Rubin, Estimating causal effects of treatments in randomized and
  nonrandomized studies. \emph{Journal of educational Psychology} \textbf{66}
  (1974), no.~5, 688

\bibitem{1911.10500}
B.~Sch\"olkopf, Causality for machine learning. \emph{arXiv preprint
  1911.10500, to appear in: {\em R. Dechter, J. Halpern, and H. Geffner.
  Probabilistic and Causal Inference: The Works of Judea Pearl. ACM books}}
  (2019)

\bibitem{SchHerSmo01}
B.~Sch{\"o}lkopf, R.~Herbrich, and A.~J. Smola, A generalized representer
  theorem. In \emph{Annual conference on computational learning theory}, edited
  by D.~Helmbold and R.~Williamson, pp. 416--426, no. 2111 in Lecture Notes in
  Computer Science, Springer, Berlin, 2001

\bibitem{Scholkopfetal16}
B.~Sch{\"o}lkopf, D.~Hogg, D.~Wang, D.~Foreman-Mackey, D.~Janzing, C.-J.
  Simon-Gabriel, and J.~Peters, Modeling confounding by half-sibling
  regression. \emph{Proceedings of the National Academy of Science (PNAS)}
  \textbf{113} (2016), no.~27, 7391--7398

\bibitem{SchJanLop16}
B.~Sch{\"o}lkopf, D.~Janzing, and D.~Lopez-Paz, Causal and statistical
  learning. In \emph{Oberwolfach reports}, pp. 1896--1899, 13(3), 2016

\bibitem{Schoelkopf2012}
B.~Sch{\"o}lkopf, D.~Janzing, J.~Peters, E.~Sgouritsa, K.~Zhang, and J.~M.
  Mooij, On causal and anticausal learning. In \emph{Proceedings of the 29th
  international conference on machine learning ({ICML})}, pp. 1255--1262, 2012

\bibitem{scholkopfetal21}
B.~Sch{\"o}lkopf, F.~Locatello, S.~Bauer, N.~R. Ke, N.~Kalchbrenner, A.~Goyal,
  and Y.~Bengio, Toward causal representation learning. \emph{Proceedings of
  the IEEE} \textbf{109} (2021), no.~5, 612--634

\bibitem{SchMuaFukHarPet15}
B.~Sch{\"o}lkopf, K.~Muandet, K.~Fukumizu, S.~Harmeling, and J.~Peters,
  Computing functions of random variables via reproducing kernel {H}ilbert
  space representations. \emph{Statistics and Computing} \textbf{25} (2015),
  no.~4, 755--766

\bibitem{SchSmo02}
B.~Sch{\"o}lkopf and A.~J. Smola, \emph{Learning with kernels}. MIT Press,
  Cambridge, MA, 2002

\bibitem{SchSriGreFuk08}
B.~Sch{\"o}lkopf, B.~K. Sriperumbudur, A.~Gretton, and K.~Fukumizu, {RKHS}
  representation of measures applied to homogeneity, independence, and
  {F}ourier optics. In \emph{Oberwolfach reports}, edited by K.~Jetter,
  S.~Smale, and D.-X. Zhou, pp. 42--44, 30, 2008

\bibitem{schwarz1978estimating}
G.~Schwarz et~al., Estimating the dimension of a model. \emph{The annals of
  statistics} \textbf{6} (1978), no.~2, 461--464

\bibitem{shah2020hardness}
R.~D. Shah and J.~Peters, The hardness of conditional independence testing and
  the generalised covariance measure. \emph{The Annals of Statistics}
  \textbf{48} (2020), no.~3, 1514--1538

\bibitem{Shajarisales15}
N.~Shajarisales, D.~Janzing, B.~Sch{\"o}lkopf, and M.~Besserve, Telling cause
  from effect in deterministic linear dynamical systems. In \emph{Proceedings
  of the 32nd international conference on machine learning ({ICML})}, pp.
  285--294, 2015

\bibitem{shalit2017estimating}
U.~Shalit, F.~D. Johansson, and D.~Sontag, Estimating individual treatment
  effect: generalization bounds and algorithms. In \emph{International
  conference on machine learning}, pp. 3076--3085, 2017

\bibitem{Shannon59}
C.~E. Shannon, Coding theorems for a discrete source with a fidelity criterion.
  In \emph{Ire international convention records}, pp. 142--163, 7, 1959

\bibitem{Shimizu2006}
S.~Shimizu, P.~O. Hoyer, A.~Hyv\"{a}rinen, and A.~J. Kerminen, A linear
  non-{G}aussian acyclic model for causal discovery. \emph{Journal of Machine
  Learning Research} \textbf{7} (2006), 2003--2030

\bibitem{shpitser2006identification}
I.~Shpitser and J.~Pearl, Identification of joint interventional distributions
  in recursive semi-markovian causal models. In \emph{Proceedings of the 21st
  national conference on artificial intelligence}, pp. 1219--1226, 2006

\bibitem{shpitser2010validity}
I.~Shpitser, T.~VanderWeele, and J.~M. Robins, On the validity of covariate
  adjustment for estimating causal effects. In \emph{Proceedings of the
  twenty-sixth conference on uncertainty in artificial intelligence}, pp.
  527--536, AUAI Press, 2010

\bibitem{simpson1951interpretation}
E.~H. Simpson, The interpretation of interaction in contingency tables.
  \emph{Journal of the Royal Statistical Society: Series B (Methodological)}
  \textbf{13} (1951), no.~2, 238--241

\bibitem{SmoGreSonSch07}
A.~J. Smola, A.~Gretton, L.~Song, and B.~Sch{\"o}lkopf, A {H}ilbert space
  embedding for distributions. In \emph{Algorithmic learning theory: 18th
  international conference}, pp. 13--31, 2007

\bibitem{Spirtes2000}
P.~Spirtes, C.~Glymour, and R.~Scheines, \emph{Causation, prediction, and
  search}. 2nd edn., MIT Press, Cambridge, MA, 2000

\bibitem{Spohn78}
W.~Spohn, \emph{Grundlagen der {E}ntscheidungstheorie}. Scriptor-Verlag, 1978

\bibitem{SteChr08}
I.~Steinwart and A.~Christmann, \emph{Support vector machines}. Springer, New
  York, NY, 2008

\bibitem{Suter.1811.00007}
R.~Suter, D.~Miladinovic, B.~Sch{\"o}lkopf, and S.~Bauer, Robustly disentangled
  causal mechanisms: Validating deep representations for interventional
  robustness. In \emph{International conference on machine learning}, pp.
  6056--6065, PMLR, 2019

\bibitem{1312.6199}
C.~Szegedy, W.~Zaremba, I.~Sutskever, J.~Bruna, D.~Erhan, I.~Goodfellow, and
  R.~Fergus, Intriguing properties of neural networks. \emph{arXiv preprint
  1312.6199}  (2013)

\bibitem{tangemannetal21}
M.~Tangemann, S.~Schneider, J.~von K{\"u}gelgen, F.~Locatello, P.~Gehler,
  T.~Brox, M.~K{\"u}mmerer, M.~Bethge, and B.~Sch{\"o}lkopf, Unsupervised
  object learning via common fate. \emph{arXiv preprint arXiv:2110.06562}
  (2021)

\bibitem{tian2001causal}
J.~Tian and J.~Pearl, Causal discovery from changes. In \emph{Proceedings of
  the seventeenth conference on uncertainty in artificial intelligence}, pp.
  512--521, Morgan Kaufmann Publishers Inc., 2001

\bibitem{Tsiaras}
A.~Tsiaras, I.~Waldmann, G.~Tinetti, J.~Tennyson, and S.~Yurchenko, Water
  vapour in the atmosphere of the habitable-zone eight-earth-mass planet
  {K2}-18b. \emph{{Nature Astronomy}}  (2019)

\bibitem{Vapnik98}
V.~N. Vapnik, \emph{Statistical learning theory}. Wiley, New York, NY, 1998

\bibitem{von2020simpson}
J.~von K{\"u}gelgen, L.~Gresele, and B.~Sch{\"o}lkopf, Simpson's paradox in
  {C}ovid-19 case fatality rates: a mediation analysis of age-related causal
  effects. \emph{{IEEE} Transactions on Artificial Intelligence} \textbf{2}
  (2021), no.~1, 18--27

\bibitem{von2020fairness}
J.~von K{\"u}gelgen, A.-H. Karimi, U.~Bhatt, I.~Valera, A.~Weller, and
  B.~Sch{\"o}lkopf, On the fairness of causal algorithmic recourse. In
  \emph{36th aaai conference on artificial intelligence}, 2022

\bibitem{von2021self}
J.~von K{\"u}gelgen, Y.~Sharma, L.~Gresele, W.~Brendel, B.~Sch{\"o}lkopf,
  M.~Besserve, and F.~Locatello, Self-supervised learning with data
  augmentations provably isolates content from style. In \emph{Advances in
  neural information processing systems 34}, 2021

\bibitem{von2020towards}
J.~von K{\"u}gelgen, I.~Ustyuzhaninov, P.~Gehler, M.~Bethge, and
  B.~Sch{\"o}lkopf, Towards causal generative scene models via competition of
  experts. In \emph{{ICLR} 2020 workshop on causal learning for decision
  making}, 2020

\bibitem{KugMeyLooSch19}
J.~von Kügelgen, A.~Mey, M.~Loog, and B.~Sch{\"o}lkopf, Semi-supervised
  learning, causality and the conditional cluster assumption. \emph{Conference
  on Uncertainty in Artificial Intelligence}  (2020)

\bibitem{woodward2001causation}
J.~Woodward, Causation and manipulability  (2001)

\bibitem{wright1928tariff}
P.~G. Wright, \emph{Tariff on animal and vegetable oils}. Macmillan Company,
  New York, 1928

\bibitem{wright1921correlation}
S.~Wright, Correlation and causation. \emph{Journal of Agricultural Research}
  \textbf{20} (1921), 557--580

\bibitem{ZhaBar18}
J.~Zhang and E.~Bareinboim, Fairness in decision-making - the causal
  explanation formula. In \emph{Proceedings of the thirty-second {AAAI}
  conference on artificial intelligence}, pp. 2037--2045, 2018

\bibitem{zhang_multi-source_2015}
K.~Zhang, M.~Gong, and B.~Sch{\"o}lkopf, Multi-source domain adaptation: A
  causal view. In \emph{Proceedings of the 29th aaai conference on artificial
  intelligence}, pp. 3150--3157, 2015

\bibitem{Zhang2009}
K.~Zhang and A.~Hyv\"{a}rinen, On the identifiability of the post-nonlinear
  causal model. In \emph{Proceedings of the 25th annual conference on
  {U}ncertainty in {A}rtificial {I}ntelligence ({UAI})}, pp. 647--655, 2009

\bibitem{Zhang2011uai}
K.~Zhang, J.~Peters, D.~Janzing, and B.~Sch{\"o}lkopf, Kernel-based conditional
  independence test and application in causal discovery. In \emph{Proceedings
  of the 27th annual conference on {U}ncertainty in {A}rtificial {I}ntelligence
  ({UAI})}, pp. 804--813, 2011

\bibitem{zhang_domain_2013}
K.~Zhang, B.~Sch{\"o}lkopf, K.~Muandet, and Z.~Wang, Domain adaptation under
  target and conditional shift. In \emph{Proceedings of the 30th international
  conference on machine learning}, pp. 819--827, 2013

\end{thebibliography}
}

\end{document}